%% file: main.tex
\documentclass[12pt]{clear2027}
\jmlrproceedings{}{Preprint}\jmlrvolume{}\jmlrpages{}\jmlryear{}
\makeatletter\renewcommand*{\@titlefoot}{}\makeatother %

\title[Joint Learning for Causal PFNs]{Joint Learning of Factual and Counterfactual worlds for Causal Prior-data Fitted Networks}
\usepackage{times}
\input{math_commands.tex}

\usepackage{makecell}\usepackage{rotating}\usepackage{placeins}
\usepackage{booktabs}
\usepackage{multirow}

\newcommand{\PP}{\mathbb{P}}
\newcommand{\stwo}{\rlap{\raisebox{0.35ex}{$^{**}$}}}

\author[Danisman Zhang Jin Kong]{\parbox{\textwidth}{\centering%
    \textbf{Furkan Danisman$^{1,2,*}$},
    \textbf{Luke Zhang$^{1,2,*}$},
    \textbf{Zhijing Jin$^{1,2,3,4,\dagger}$},
    \textbf{Dehan Kong$^{1,\dagger}$}
    \\[0.4em]
    {\normalsize\mdseries University of Toronto$^{1}$, Jinesis Lab \& Vector Institute$^{2}$}\\
    {\normalsize\mdseries Max Planck Institute for Intelligent Systems$^{3}$, EuroSafeAI$^{4}$}\\[0.4em]
    {\small\mdseries $^{*}$Corresponding authors:}\\ {\small\mdseries \texttt{furkan.danisman@mail.utoronto.ca}, \texttt{lukelz.zhang@mail.utoronto.ca}}\\
    {\small\mdseries $^{\dagger}$Equal co-supervision}}}


\begin{document}

\maketitle

\begin{abstract}
Estimating treatment effects is central to science across many disciplines and requires counterfactual reasoning about alternative worlds. In the real world, however, each unit reveals its outcome only under the factual world. Thus, the factual and counterfactual outcomes can never be jointly learned by any model trained on real data. The problem is therefore tractable only in synthetic settings, where both worlds can be generated. Recent progress on prior-data fitted networks (PFNs) has shown that models pre-trained purely on synthetic data can transfer to real-world prediction, achieving state-of-the-art tabular performance via in-context learning. This paradigm has since carried over to causal effect estimation, predicting interventional outcomes given observational data. Despite this fully synthetic setup, existing causal PFNs carry the real-world constraint into their training objective, confining learning to a single world's outcome per unit. The dependence between the two worlds is thereby lost, and the density of the treatment effect relies by default on independence, which is implausible in practice. We therefore propose joint learning, a training setup in which a causal PFN learns the joint distribution of the factual and counterfactual outcomes of each unit. From the joint distribution, we obtain both the response and the individual treatment effect densities, and at inference, we apply optimal transport across units to obtain a calibrated approximation of the average treatment effect density. Ours is therefore, to our knowledge, the first causal PFN approach to provide all three densities together while accounting for the cross-world dependence. We apply joint learning and optimal transport to existing causal PFNs and evaluate it across semi-synthetic, synthetic, and real benchmarks spanning diverse causal scenarios. The jointly trained models significantly improve upon their baselines in point estimation and uncertainty quantification at both the individual and the average level in most settings, indicating that joint learning is a preferable training setup for causal PFNs.
\end{abstract}

\section{Introduction}

Treatment effect estimation is central to medicine, economics, finance, and public policy \citep{Imbens_Rubin_2015,policy}. The effect can be considered at the unit or at the population level \citep{unit}. At the unit level, it is the difference between a unit's factual and counterfactual outcomes; its estimation is therefore inherently a joint learning problem over them. In practice, however, each unit is observed only under the factual world, so the pair can never be jointly learned from real data \citep{heckman1997making,rubin1974estimating,peters,unit}. The only viable option is therefore learning each outcome separately \citep{shalit2017estimating, alaa2017bayesian}. Yet at best, this recovers the two marginal distributions; the dependence between the outcomes remains unidentified even under randomization \citep{heckman1997making, fan2010sharp}. Assuming the outcomes are independent makes learning practical and solves the joint problem but is implausible \citep{andrews2021insights, vanlancker2024chasing}, and such misspecification carries a real inferential cost \citep{Ding02012}. Joint learning at unit level therefore remains tractable only in synthetic settings, in which both outcomes can be generated. 

At the population level, in contrast to unit level, the average treatment effect depends only on the two marginal distributions, which randomization identifies. Randomized controlled trials are therefore the ideal setup for it, but are often unethical, expensive, slow, or infeasible \citep{hernan2020}. In these cases, one has to rely on observational data. Recovering the marginals from observational data, however, requires strong assumptions, such as unconfoundedness, positivity, or a valid instrument \citep{hernan2020, pearl2009, chernozhukov2018, wager2018}. These assumptions are difficult to justify in practice: confounders are frequently unmeasured, valid instruments seldom exist, and the causal graph is rarely known fully.

To address the assumption problem, a family of causal PFNs has recently been proposed \citep{dopfn, uwyk, causalpfn, ma2026causalfm, thumm2026causal}. A PFN is a transformer that is trained on synthetic data and returns a posterior-predictive distribution \citep{muller2022pfn}. PFNs achieve state-of-the-art performance on tabular prediction on observational data \citep{hollmann2022tabpfn, hollmann2025tabpfn}, showing that a model trained purely on synthetic data can transfer to real-world prediction. Causal PFNs adopt the same architecture but predict interventional rather than observational outcomes \citep{stith2026causalfoundationmodels}. They resolve the assumption problem by encoding the identification and structural assumptions in the training prior, and have outperformed traditional causal machine-learning methods \citep{chernozhukov2018, wager2018, nonpfn1,nonpfn2,nonpfn3} on standard tabular benchmarks \citep{dopfn, causalpfn, uwyk}.

Causal PFNs thus provide a promising direction by addressing the assumption problem at the population level. However, critical gaps remain. First, existing models report only a point estimate of the average treatment effect, and they fail to provide a mechanism to estimate its density. Second, they do not address the joint learning problem at the unit level. Despite the fully synthetic setup, most existing models carry the real-world constraint into their training objective, confining learning to a single world's outcome per unit \citep{dopfn, uwyk, causalpfn}. These models therefore cannot learn the cross-world dependence and default to the implausible independence assumption. Training a model on the treatment effect directly avoids this assumption \citep{ma2026causalfm} but discards the response information. Hence, a decision maker cannot obtain the response, individual treatment effect, and average treatment effect densities together from any existing causal PFN. To address this, we propose joint learning, a training setup in which a causal PFN learns the joint distribution of each unit's factual and counterfactual outcomes. From this joint distribution, the response and individual treatment effect densities follow, and optimal transport across units at inference provides a calibrated average treatment effect density. Ours is thus the first causal PFN approach to provide all three densities while accounting for the cross-world dependence. We apply joint learning with optimal transport to existing causal PFNs and evaluate it across semi-synthetic, synthetic, and real benchmarks spanning diverse causal scenarios. The jointly trained models significantly improve upon their baselines in point estimation and uncertainty quantification in most settings, indicating that joint learning is a preferable training setup for causal PFNs.

\section{Background and related work}

\subsection{Structural causal model and interventions}
\label{sec:scm-outcomes}
We work with an observational dataset $\mathcal{D}_{\text{obs}} = \{(X_i, T_i, Y_i)\}_{i=1}^n$, where each row records a unit's covariates $X_i \in \mathbb{R}^d$, the treatment it received $T_i \in \{0,1\}$, and the outcome observed under that treatment, $Y_i \in \mathbb{R}$.

We assume $\mathcal{D}_{\text{obs}}$ is generated by a structural causal model \citep{pearl2009, peters} (SCM) $\psi = (G, f, p_\varepsilon)$ over $\{X, T, Y\}$: $G$ a directed graph, $f = \{f_V\}_{V \in \{X,T,Y\}}$ structural equations shared across all units, $p_\varepsilon$ the exogenous noise distribution. Each row of $\mathcal{D}_{\text{obs}}$ is one sample $\varepsilon_i \sim p_\varepsilon$ propagated through $f$. An intervention $\text{do}(T=t)$ under SCMs sets the treatment to value $t$ and severs its dependence on its causal parents, while leaving every other structural equation untouched. This defines the potential outcome:
\begin{equation}
    Y_{\text{do}(t)}(x,\varepsilon) = f_Y\big(x, t, \mathrm{pa}_{-T}(\psi;x,\varepsilon), \varepsilon_Y\big), \quad t\in\{0,1\},
\end{equation}
where $\mathrm{pa}_{-T}(\psi; x, \varepsilon)$ denotes the values of $Y$'s graph-parents other than $T$, propagated through $\psi$. The individual treatment effect (ITE) of a unit is the difference of its two potential outcomes,
\begin{equation}
    {\delta := Y_{\text{do}(1)} - Y_{\text{do}(0)}.}
    \label{eq:ite}
\end{equation}
The conditional average treatment effect (CATE) is its mean at covariate value $x$, $\tau(x) := \mathbb{E}[\delta \mid X = x]$, and the average treatment effect (ATE) is the average of the CATE over the covariate population, $\mathbb{E}_X[\tau(X)]$.

\subsection{Joint learning of outcomes}
\label{sec:joint-supervision}

The dependence between a unit's two potential outcomes in Equation~\ref{eq:ite} is captured by the \textit{cross-world correlation} \citep{bodik2026}:
\begin{equation}
    \rho(x) = \mathrm{corr}\big(Y_{\text{do}(1)}, Y_{\text{do}(0)} \mid X = x\big). \label{eq:rho}
\end{equation}
Since in real data each unit is observed under a single treatment, $\rho(x)$ is not identifiable \citep{rubin1974estimating, heckman1997making}. Hence, models trained on real data \citep{alaa2017bayesian,louizos2017causaleffectinferencedeep,deeptwin,yoon2018ganite} cannot learn this correlation directly. This limitation does not apply to models trained on synthetic data, where both potential outcomes of every unit can be computed and $\rho(x)$ is therefore known. Since the two outcomes are defined by the same exogenous variables \citep{pearl2009}, $\rho(x)$ is determined by the mechanisms of the SCM. A model can therefore be trained on data simulated from SCMs, with the pair $\big(y_{\text{do}(0)}, y_{\text{do}(1)}\big)$ of every unit as its target, which makes the joint law $P^\star\big(Y_{\text{do}(0)}, Y_{\text{do}(1)} \mid X = x\big)$ a direct learning target. We refer to training on such a pair as \textit{joint learning}.

\subsection{Prior-data fitted networks for Bayesian inference}
\label{sec:pfn}
Prior-Data Fitted Networks (PFNs) \citep{muller2022pfn} are neural networks trained to perform Bayesian prediction. Under a prior $p(\psi)$ over SCMs, the target is the posterior predictive distribution
\begin{equation}
    p(y \mid x, \mathcal{D}_{\text{obs}}) = \int p(y \mid x, \psi)\, p(\psi \mid \mathcal{D}_{\text{obs}})\, d\psi. \label{eq:ppd}
\end{equation}
A PFN is trained once, offline, on synthetic datasets generated from the prior to approximate this distribution. At inference time, it returns the distribution for a new $\mathcal{D}_{\text{obs}}$ in a single forward pass, without gradient updates.

\subsection{Causal prior-data fitted networks}
\label{sec:cfm}

Standard PFNs such as TabPFN \citep{hollmann2025tabpfn} approximate the distribution in Equation~\ref{eq:ppd} for an observational outcome $y$, whereas causal PFNs approximate it for an interventional outcome $Y_{\text{do}(t)}$. Since this distribution depends on the prior $p(\psi)$, existing causal PFNs differ mainly in what causal structures their priors commit to. Do-PFN \citep{dopfn} samples SCMs from a broad prior with no identifiability restrictions and the model is trained to predict interventional outcomes from $\mathcal{D}_{\text{obs}}$ alone, without knowledge of the graph. Use What You Know (UWYK) \citep{uwyk} uses an unrestricted prior similar to Do-PFN, but supplies causal graph knowledge at inference time as a partially known ancestral matrix $\tilde{\mathbf{T}} \in \{-1,0,1\}^{(d+2) \times (d+2)}$, injected into the feature-wise attention of the transformer, so that a single model spans the range from no graph information to full ancestral knowledge. CausalPFN \citep{causalpfn} builds its prior from real and synthetic tables, taking random columns as the expected potential outcomes $m_t$ and the noise terms $\eta_t$, with $Y_{\mathrm{do}(t)} = m_t + \eta_t$. The model is trained on $m_t$ instead of $Y_{\mathrm{do}(t)}$. This removes the random noise $\eta_t$ from its target, but variation of $m_t$ remains, which comes from characteristics that $x$ does not capture. The treatment is drawn as $T \sim \mathrm{Bernoulli}(\sigma(f(X)))$ for a random propensity function $f$ that depends on $X$ alone, which guarantees ignorability by design. Its estimates are therefore valid only on data that satisfy this assumption. CausalFM \citep{ma2026causalfm} uses separate priors encoding different identifiability constraints. Each model thus satisfies its constraint by design but covers only that setting. It predicts the distribution of the ITE $\delta$ directly, without modeling either $Y_{\mathrm{do}(0)}$ or $Y_{\mathrm{do}(1)}$; because the map $(Y_{\mathrm{do}(0)}, Y_{\mathrm{do}(1)}) \mapsto \delta$ is not injective, neither potential outcome can be recovered from $\delta$ alone.

Despite these different approaches, no causal PFN performs the joint learning of Section~\ref{sec:joint-supervision}. Since Do-PFN, UWYK and CausalPFN are trained on one outcome per unit, the dependence between a unit's two potential outcomes is never learned, and they are treated as independent. CausalFM is trained on their difference $\delta$, which does not determine their joint law. Two limitations follow.

\paragraph{Statistical efficiency.}
Single-outcome models never see both outcomes of the same unit, which limits how efficiently they learn the treatment effect. They see each unit under one treatment only, and learn the other outcome from different units. Their estimate of the treatment effect therefore carries the noise of both outcomes. A model trained on both sees the two outcomes of each unit. When $\rho(x) > 0$, the noise they share cancels in their difference, so this model can learn the treatment effect from fewer units. Section~\ref{sec:performance} discusses this gap.

\paragraph{Distributional information.}

The second limitation concerns the densities that existing causal PFNs report. A decision maker needs the response densities (e.g., whether a treated unit's outcome exceeds a clinical or economic threshold), and the ITE and ATE densities (e.g., whether to intervene at all). To our knowledge, no causal PFN reports an ATE density, as for the response and ITE densities, each model falls short in a different way. CausalFM provides an ITE density, but no response densities. Do-PFN, UWYK and CausalPFN can provide both,\footnote{CausalPFN outputs densities of the expected outcomes, which give the ITE only under shared idiosyncratic noise.} but their ITE density treats the two outcomes as independent. This assumption is likely to be violated in practice \citep{andrews2021insights, vanlancker2024chasing, bodik2026} and may introduce spurious treatment effects.

Our proposed joint learning, combined with an optimal-transport aggregation at inference, resolves these gaps. It provides statistical efficiency, gives both the response and ITE densities without an independence assumption, and adds a calibrated approximation of the ATE density.

\section{Methodology}
\label{sec:joint-head}

\begin{figure}[tbp]
    \centering
    \includegraphics[width=\linewidth]{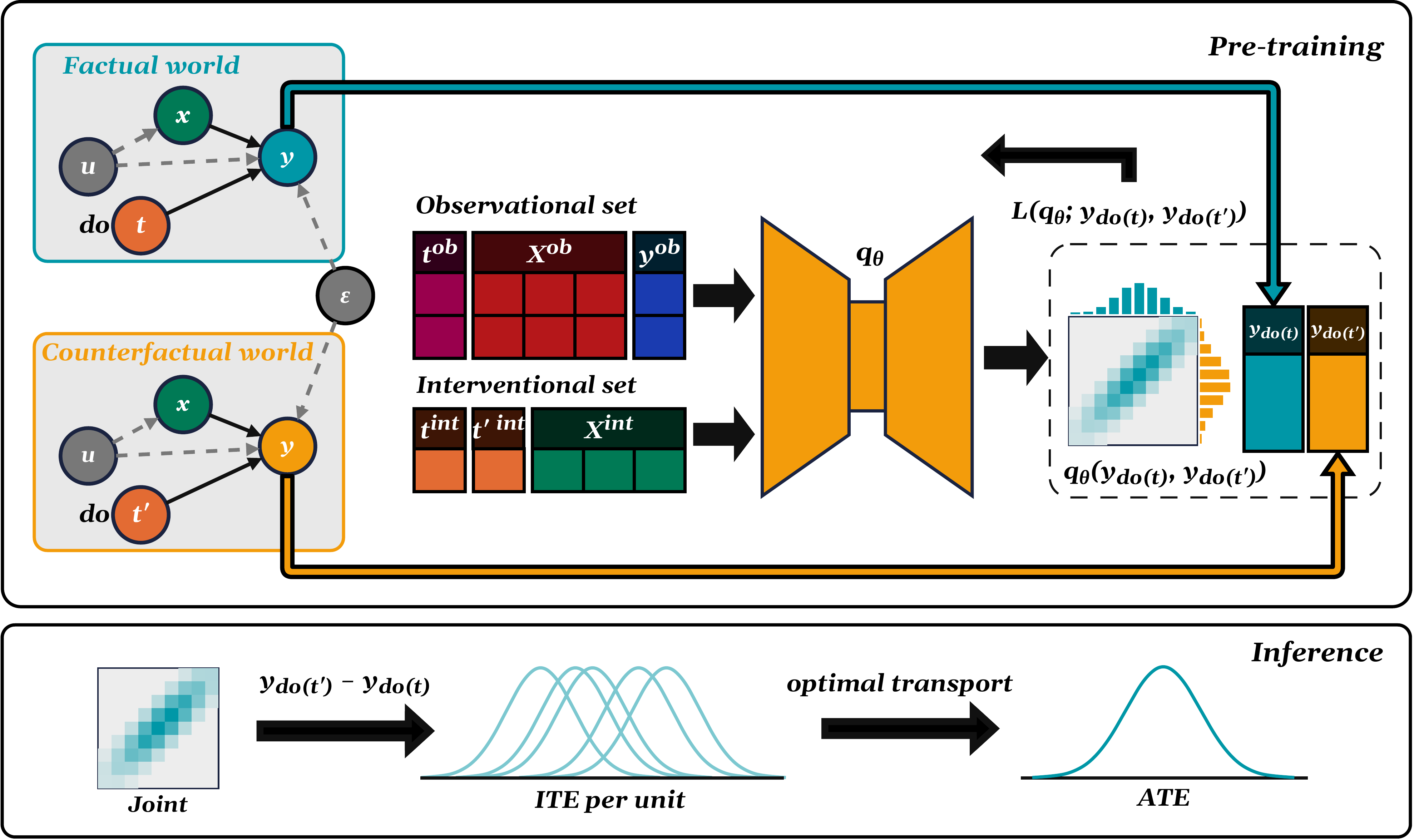}
    \caption{\small Overview of joint learning. \textit{Pre-training:} for each unit of a sampled SCM, the two potential outcomes are generated in a factual world, under $\mathrm{do}(t)$, and a counterfactual world, under $\mathrm{do}(t')$, which share the same exogenous noise $\varepsilon$. The model $q_\theta$ receives an observational set and the covariates of query units, whose outcomes are unknown, predicts the joint distribution of their two potential outcomes, and is trained on the negative log-likelihood of the true pair. \textit{Inference:} the difference of the two outcomes under the predicted joint gives the ITE distribution of each unit, and optimal transport aggregates these into the ATE distribution.\protect\footnotemark}
    \label{fig:overview}
\end{figure}

We approximate the joint posterior predictive distribution of a unit's factual and counterfactual outcomes. For a query with covariates $x$, observational context $\mathcal{D}_{\mathrm{obs}}$, and treatments $t \neq t'$, it is defined as
\begin{equation}
    p(y_{\mathrm{do}(t)}, y_{\mathrm{do}(t')} \mid x, \mathcal{D}_{\mathrm{obs}}) = \int p(y_{\mathrm{do}(t)}, y_{\mathrm{do}(t')} \mid x, \psi)\, p(\psi \mid \mathcal{D}_{\mathrm{obs}})\, d\psi.
    \label{eq:joint-ppd}
\end{equation}
The uncertainty decomposition and consistency result of \citet{dopfn} extend directly to this distribution (Appendix~\ref{app:proofs}). At inference, the ITE density of each unit is computed from the predicted joint, and optimal transport aggregates these into the ATE density. Figure~\ref{fig:overview} gives an overview of the method. Section~\ref{sec:performance} establishes the statistical efficiency of the joint objective, and Section~\ref{sec:info-gain} details how the ITE and ATE densities are obtained.

\footnotetext{The layout of this figure is inspired by the overview figure of \citet{dopfn}.}

\subsection{Statistical efficiency of the joint objective}
\label{sec:performance}

We estimate the CATE $\tau(x)$ by predicting the distribution of the ITE $\delta$ at $x$. Its mean is the point estimate $\hat\tau(x)$, and its quantiles give a posterior predictive interval for $\delta$. As defined in Section~\ref{sec:scm-outcomes}, each potential outcome is a function of the covariates and the exogenous noise, $Y_{\mathrm{do}(t)}(x, \varepsilon)$, and both potential outcomes of a unit share the same $\varepsilon$. Let $\mu_t(x)$ and $\sigma_t^2(x)$ denote the mean and variance of $Y_{\mathrm{do}(t)}(x, \varepsilon)$ over $\varepsilon$ given $X = x$, and $\rho(x)$ the cross-world correlation of Equation~\ref{eq:rho}. Under an SCM $\psi$, the variance of the ITE at $x$ is
 \begin{equation}
     v_\psi(x) = \sigma_1^2(x) + \sigma_0^2(x) - 2\rho(x)\,\sigma_1(x)\,\sigma_0(x),
 \end{equation}
 and by the law of total variance, the posterior predictive variance of $\delta$ decomposes as
 \begin{equation}
     \mathrm{Var}(\delta \mid x, \mathcal{D}_{\mathrm{obs}}) = \mathrm{Var}\big(\tau_\psi(x) \mid \mathcal{D}_{\mathrm{obs}}\big) + \mathbb{E}\big[v_\psi(x) \mid \mathcal{D}_{\mathrm{obs}}\big],
 \label{eq:var-decomp}
 \end{equation}
 
where $\tau_\psi(x)$ is the CATE under $\psi$. The first term, the uncertainty about the CATE, shrinks as more data are observed; the second, the spread of units with the same $x$, does not, and it depends on $\rho(x)$. A CATE posterior contains only the first term; the two are equal only when $\delta$ is deterministic given $x$.

The second term is where the learning objectives mainly differ, since it includes $\rho(x)$. The single-outcome models see each unit under a single treatment, thus discarding the dependence and treating the two outcomes as independent, which implies $\rho(x) = 0$. This creates a mismatch with what we expect in real datasets. Setting $\rho(x)$ to a different value at inference can address the mismatch, but choosing it correctly is not a trivial task. From $\mathcal{D}_{\mathrm{obs}}$ alone, one can only get a conservative bound on $\rho(x)$. Any value is therefore a random guess within this bound and does not improve with more data. Choosing $\rho(x)$ is, however, not the only problem. Even if the true $\rho(x)$ were known, it would not suffice in general, since the two predicted distributions also contain the model's uncertainty about $\mu_0$ and $\mu_1$, whose dependence $\rho(x)$ does not describe (Appendix~\ref{app:malc}). Joint learning instead approximates the second term with the $\rho(x)$ of each SCM, weighted by how well that SCM fits $\mathcal{D}_{\mathrm{obs}}$; as each SCM contributes $\mu_0$ and $\mu_1$ together, the uncertainty about them keeps its dependence as well. As more data is observed, this weight concentrates on the SCMs compatible with $\mathcal{D}_{\mathrm{obs}}$, and the estimate improves. The remaining uncertainty about $\rho(x)$ then reduces to the uncertainty over these SCMs.

The objectives also differ in how efficiently they learn $\tau(x)$ from finite training data. A single-outcome model learns each outcome from different units, so its estimate carries the full noise of both. A jointly trained model learns from both outcomes of the same unit. Their difference $\delta$ carries less noise whenever $\rho(x) > 0$, so it can learn $\tau(x)$ more efficiently. The joint objective also weighs errors differently. Under the joint loss, a predicted $\tau(x)$ that is off by $e$ increases the loss by at least $e^2/\big(2\,\mathrm{Var}(\delta)\big)$, against $e^2/\big(4(\sigma_0^2+\sigma_1^2)\big)$ under the single-outcome loss (Proposition~\ref{prop:effect-cost}). The joint loss thus penalizes these errors more when $\rho(x) > 0$, which can direct more of the learning toward $\tau(x)$. Both arguments assume that the objectives learn the same outcomes. A model trained to predict $m_t$ instead learns $\tau(x)$ without the idiosyncratic noise $\eta_t$, and can therefore be more efficient than the joint. However, this comes at a cost, since without that noise its response densities would be overconfident. The efficiency gaps between these objectives vanish with unlimited training data and model capacity, since under the same prior every objective then recovers the same posterior mean of $\tau(x)$. Yet, the estimates of Equation~\ref{eq:var-decomp} still differ, and only joint learning estimates it consistently. Hence, the joint objective provides the largest information set to learn the ITE and response densities.

\paragraph{Data efficiency.}

As the single-outcome observation is a randomised function of the joint one, any parameter that can be estimated from it can be estimated from the joint at least as well (Lemma~\ref{lem:sufficiency}). Nothing is therefore lost by joint training. In addition to this preservation, joint training also gains information from two sources. Each unit contributes an additional outcome, and observing the two outcomes together makes their dependence learnable. To separate the two sources, we conduct an ablation in which the baseline is trained on the same number of outcomes as the joint model.

\paragraph{Computational cost and resolution.}
For PFN models with a one-dimensional output, the output is a distribution over a grid of $K$ points, and the output layer costs $O(K)$ per query. For the joint model, the output is two-dimensional, and the output layer costs $O(K^2)$ per query. To compare against the published baselines at a similar cost, each joint model uses the square root of its baseline's grid size. The coarser grid reduces the model's representational capacity and places it at a disadvantage relative to its baseline. Improvements under this setup therefore reflect a conservative gain. The resolution loss can be mitigated at inference by a density estimator such as kernel density estimation~\citep{parzen1962estimation}, a Gaussian mixture~\citep{mclachlan2000finite}, or MALC~\citep{danisman2026malc}, at the cost of additional inference time; our results use the grid without it. We further conduct an ablation in which each baseline is re-trained on the grid of its joint model, for a matched comparison. The comparisons thus show the trade-off between cost and performance, that is, whether, under a fixed budget, one should prioritize resolution or joint learning.

\subsection{Distributional inference for ITE and ATE}
\label{sec:info-gain}

Let $x_1, \dots, x_n$ be a sample of query covariates. For a query at $x_i$, the ITE density is the distribution of $\delta$ under the joint of Equation~\ref{eq:joint-ppd}, with $t = 1$ and $t' = 0$,
\begin{equation}
    f_\delta^{(i)}(d) = \int p\big(y_{\mathrm{do}(t')} = y,\, y_{\mathrm{do}(t)} = y + d \mid x_i, \mathcal{D}_{\mathrm{obs}}\big)\, dy,
    \label{eq:ite-integral}
\end{equation}
which is computed from the output of the joint (Appendix~\ref{app:malc}). Since the ATE is the average of the ITEs across units, these densities need to be aggregated into one. A naive pointwise average would not describe the average effect, since differences between the units would inflate its spread. We therefore average their quantile functions instead and scale the result by conformal calibration \citep{vovk2005}. This gives $f_{\mathrm{ATE}}$, a calibrated approximation of the ATE density, as the scaled 2-Wasserstein barycenter of the ITE densities,
\begin{equation}
    Q_{\mathrm{ATE}}(u) = \bar{\tau} + s \Big( \frac{1}{n} \sum_{i=1}^{n} Q_{\delta}^{(i)}(u) - \bar{\tau} \Big), \qquad u \in (0, 1),
    \label{eq:ate-quantile}
\end{equation}
where $Q_{\delta}^{(i)}$ is the quantile function of $f_{\delta}^{(i)}$ and $s$ the scaling parameter, calibrated on the model's prior. Its mean $\bar{\tau}$ is the average of the per-query ITE means, leaving the ATE point estimate unchanged. Appendix~\ref{app:ot} gives the derivation, and Figure~\ref{fig:ot-illustration} illustrates the procedure.

\section{Experimental Setup}
\label{sec:results}

We evaluate our proposed joint learning method across three causal PFN models and four benchmarks, measuring point estimation and uncertainty calibration. This section describes the compared baselines, ablations, benchmarks, and evaluation metrics.

\paragraph{Compared baselines.}
We evaluate the effect of joint learning on three major causal PFN models: Do-PFN \citep{dopfn}, UWYK \citep{uwyk}, and CausalPFN \citep{causalpfn}. For each model, we hold the prior over SCMs and the transformer architecture fixed, generate both potential outcomes of each unit, and change only the learning objective to the joint learning of Section~\ref{sec:joint-supervision}. The jointly trained variant is denoted as JT throughout the results. We exclude another influential causal PFN model, CausalFM \citep{ma2026causalfm}, for two reasons. It trains a separate model per identification setting, which is computationally heavy to replicate across our ablation setups. It also already supervises the ITE $\delta$, so joint learning would be expected to add its response densities rather than change its ITE results.

We compare each JT model against its published checkpoint, except for CausalPFN. It has two training stages, and the first cannot be trained on paired outcomes. Hence, we re-train its baseline from the second (causal) stage and refer to it as CausalPFN-C. We provide the details and the results against the published two-stage checkpoint on Appendix~\ref{app:twostagecomparison}.Additionally, UWYK's authors provide two training options, with a binary or a continuous treatment. The published checkpoint covers only the continuous one, while our benchmarks have a binary treatment, and we therefore re-train the binary version under the same protocol and report it in Appendix~\ref{app:binarytraining}. Both comparisons lead to conclusions similar to those of our main results.

\paragraph{Cross-world correlation of the baselines.}
As discussed in Section~\ref{sec:performance}, the single-outcome baselines treat the two outcomes as independent, which implies $\rho(x) = 0$, although another value of $\rho(x)$ can be set at inference. To test whether this closes the gap to joint learning, we report each baseline under $\rho \in \{-0.5, 0, 0.5, 1\}$ or a finer grid for an averaged comparison, applied through a Gaussian copula, while the JT models use their learned $\rho(x)$.

\paragraph{Ablations.}
We conduct two ablations, following the arguments in Section~\ref{sec:performance}. In the first, each baseline is re-trained at the reduced resolution $K$ of its JT counterpart. This isolates the effect of joint learning from the effect of resolution. In the second, each baseline is given the same amount of data information. This isolates the effect of the dependence from the effect of the additional data. Each baseline thus enters with three variants: its JT version, a single-outcome version at the matched resolution $K$, and a single-outcome version with the matched label budget. Together with the CausalPFN second-stage and UWYK binary re-trainings, this amounts to eleven trained models in total. The ablation results are reported in Appendix~\ref{app:resolution-ablation} and computation details in Appendix~\ref{app:compute}.

\paragraph{Benchmarks.}
We evaluate on four benchmarks with semi-synthetic, synthetic, and real datasets. The semi-synthetic benchmark, RealCause \citep{neal2020realcause}, generates the two potential outcomes independently, reflecting the $\rho \equiv 0$ setting. The two synthetic benchmarks have dependent outcomes. ComplexMech, adopted from UWYK, draws random structural causal models with nonlinear mechanisms. The case study, adopted from Do-PFN, covers six causal scenarios: observed confounding, observed mediation, their combination, unobserved confounding, and frontdoor and backdoor structures. The real benchmark, Twins \citep{louizos2017causaleffectinferencedeep}, consists of twin pairs whose two outcomes are treated as the factual and counterfactual outcomes according to the twins' weight at birth. Due to the nature of twins, dependence between the outcomes therefore also exists in this case. For Twins, we follow \citet{bodik2026retro} and predict the outcome of one twin from the outcome of the other. Appendix~\ref{app:benchmarks} describes how each benchmark is constructed, and Table~\ref{tab:rho-all} reports how the cross-world correlation varies across realizations.

\paragraph{Evaluation metrics.}
From each predicted ITE distribution, we evaluate its mean, the CATE estimate, by $\sqrt{\mathrm{PEHE}}$, and its 95\% posterior predictive interval by the standardised interval score $\mathrm{IS}_{0.05}/\sigma_Y$ \citep{Gneiting01032007}, which combines coverage and length into a single metric by penalising misses in proportion to their distance from the interval. The ATE estimate is evaluated by its relative error $\epsilon_{\mathrm{ATE}}$. Across datasets, all metrics are summarised by the geometric mean, Average$_g$, and differences between models are tested for significance with Hochberg correction. Definitions and rationales are given in Appendix~\ref{app:metrics}.

\section{Results}
\label{sec:outcomes}

We report the results on the semi-synthetic (Section~\ref{sec:semi-synthetic}), synthetic (Section~\ref{sec:synthetic}), and real (Section~\ref{sec:twins-results}) benchmarks. Due to space constraints, for uncertainty quantification, we discuss only the ITE in detail in the main text. Nevertheless, similar results also hold for the ATE, and our JT models attain significantly better interval scores than their baselines in $42$ of the $48$ comparisons across model families and benchmarks. We provide the full results and discussion for it in Appendix~\ref{app:ate-results}.

\subsection{Semi-synthetic}
\label{sec:semi-synthetic}

RealCause is a commonly used benchmark in causal inference, but it generates the factual and counterfactual outcomes independently (Table~\ref{tab:rho-all}). In reality, the two outcomes of a unit are expected to depend on each other \citep{bodik2026,andrews2021insights,vanlancker2024chasing}. We therefore treat RealCause as a robustness test in which the dependence assumption is violated. As causal PFN evaluations on this benchmark commonly focus on point estimation, and due to space constraints, we report point estimates in Table~\ref{tab:rc-point} and defer interval calibration to Figure~\ref{fig:rc-is} (Appendix~\ref{app:rc-is}).

\begin{table*}[t]
\centering
\caption{\small Unit- and population-level point estimation errors on RealCause across five datasets (mean $\pm$ standard error of the mean over realizations). Lower is better. Bold marks the better model within each family. Significance on Average$_g$ at the $0.1$ and $0.05$ level is indicated by $^{*}$ and $^{**}$, respectively. Average$_g$ is the geometric mean over datasets, of $\sqrt{\mathrm{PEHE}}$ standardised by each dataset's outcome scale $\sigma_Y$ and of $\epsilon_{\mathrm{ATE}}$.}
\label{tab:rc-point}
\small
\setlength{\tabcolsep}{1.5pt}
\resizebox{\textwidth}{!}{%
\begin{tabular}{l cc cc cc cc cc | cc}
\toprule
& \multicolumn{2}{c}{IHDP} & \multicolumn{2}{c}{ACIC} & \multicolumn{2}{c}{CPS} & \multicolumn{2}{c}{PSID} & \multicolumn{2}{c}{PSID-B} & \multicolumn{2}{c}{\textbf{Average}$_{\boldsymbol{g}}$} \\
\cmidrule(lr){2-3}\cmidrule(lr){4-5}\cmidrule(lr){6-7}\cmidrule(lr){8-9}\cmidrule(lr){10-11}\cmidrule(lr){12-13}
\textbf{Model} & $\sqrt{\mathrm{PEHE}}\downarrow$ & $\epsilon_{\mathrm{ATE}}\downarrow$ & $\sqrt{\mathrm{PEHE}}\downarrow$ & $\epsilon_{\mathrm{ATE}}\downarrow$ & $\sqrt{\mathrm{PEHE}}\downarrow$ & $\epsilon_{\mathrm{ATE}}\downarrow$ & $\sqrt{\mathrm{PEHE}}\downarrow$ & $\epsilon_{\mathrm{ATE}}\downarrow$ & $\sqrt{\mathrm{PEHE}}\downarrow$ & $\epsilon_{\mathrm{ATE}}\downarrow$ & $\sqrt{\mathrm{PEHE}}\downarrow$ & $\epsilon_{\mathrm{ATE}}\downarrow$\\
\midrule
Do-PFN & 6.07{\tiny$\pm$0.9} & 0.93{\tiny$\pm$0.4} & 4.11{\tiny$\pm$0.5} & 0.66{\tiny$\pm$0.0} & 12014{\tiny$\pm$32} & 0.88{\tiny$\pm$0.0} & 20907{\tiny$\pm$138} & 0.93{\tiny$\pm$0.0} & 22892{\tiny$\pm$161} & 1.09{\tiny$\pm$0.0} & 1.22{\tiny$\pm$0.1} & 0.89{\tiny$\pm$0.1} \\
Do-PFN JT & \textbf{5.72}{\tiny$\pm$0.8} & \textbf{0.60}{\tiny$\pm$0.2} & \textbf{3.65}{\tiny$\pm$0.6} & \textbf{0.53}{\tiny$\pm$0.1} & \textbf{11121}{\tiny$\pm$23} & \textbf{0.53}{\tiny$\pm$0.0} & \textbf{15844}{\tiny$\pm$157} & \textbf{0.10}{\tiny$\pm$0.0} & \textbf{16356}{\tiny$\pm$157} & \textbf{0.18}{\tiny$\pm$0.0} & \textbf{1.03}\stwo{\tiny$\pm$0.1} & \textbf{0.31}\stwo{\tiny$\pm$0.0} \\
\midrule
UWYK No-Anc & 6.29{\tiny$\pm$0.8} & 0.67{\tiny$\pm$0.1} & 3.35{\tiny$\pm$0.5} & 0.45{\tiny$\pm$0.1} & 12562{\tiny$\pm$37} & 0.96{\tiny$\pm$0.0} & 22443{\tiny$\pm$132} & 0.98{\tiny$\pm$0.0} & 21886{\tiny$\pm$136} & 0.94{\tiny$\pm$0.0} & 1.20{\tiny$\pm$0.1} & 0.76{\tiny$\pm$0.0} \\
UWYK No-Anc JT & \textbf{4.51}{\tiny$\pm$0.6} & \textbf{0.38}{\tiny$\pm$0.1} & \textbf{2.74}{\tiny$\pm$0.5} & \textbf{0.13}{\tiny$\pm$0.0} & \textbf{12299}{\tiny$\pm$31} & \textbf{0.88}{\tiny$\pm$0.0} & \textbf{21426}{\tiny$\pm$145} & \textbf{0.85}{\tiny$\pm$0.0} & \textbf{20039}{\tiny$\pm$141} & \textbf{0.68}{\tiny$\pm$0.0} & \textbf{1.05}\stwo{\tiny$\pm$0.1} & \textbf{0.48}\stwo{\tiny$\pm$0.0} \\
\midrule
UWYK Anc & 5.49{\tiny$\pm$0.8} & 0.49{\tiny$\pm$0.1} & 2.77{\tiny$\pm$0.5} & 0.23{\tiny$\pm$0.1} & \textbf{10773}{\tiny$\pm$58} & \textbf{0.61}{\tiny$\pm$0.0} & 23304{\tiny$\pm$156} & 1.06{\tiny$\pm$0.0} & \textbf{19039}{\tiny$\pm$212} & \textbf{0.62}{\tiny$\pm$0.0} & 1.07{\tiny$\pm$0.1} & 0.54{\tiny$\pm$0.0} \\
UWYK Anc JT & \textbf{4.30}{\tiny$\pm$0.6} & \textbf{0.33}{\tiny$\pm$0.1} & \textbf{2.72}{\tiny$\pm$0.5} & \textbf{0.12}{\tiny$\pm$0.0} & 12475{\tiny$\pm$28} & 0.92{\tiny$\pm$0.0} & \textbf{21850}{\tiny$\pm$136} & \textbf{0.90}{\tiny$\pm$0.0} & 21467{\tiny$\pm$141} & 0.86{\tiny$\pm$0.0} & \textbf{1.06}{\tiny$\pm$0.1} & \textbf{0.49}{\tiny$\pm$0.0} \\
\midrule
CausalPFN-C & \textbf{0.82}{\tiny$\pm$0.1} & \textbf{0.05}{\tiny$\pm$0.0} & 1.41{\tiny$\pm$0.2} & 0.09{\tiny$\pm$0.0} & \textbf{10043}{\tiny$\pm$181} & \textbf{0.35}{\tiny$\pm$0.0} & 18281{\tiny$\pm$297} & 0.68{\tiny$\pm$0.0} & 18645{\tiny$\pm$314} & 0.71{\tiny$\pm$0.0} & \textbf{0.60}{\tiny$\pm$0.0} & 0.24{\tiny$\pm$0.0} \\
CausalPFN-C JT & 1.20{\tiny$\pm$0.1} & 0.07{\tiny$\pm$0.0} & \textbf{1.26}{\tiny$\pm$0.2} & \textbf{0.09}{\tiny$\pm$0.0} & 10902{\tiny$\pm$176} & 0.39{\tiny$\pm$0.0} & \textbf{15557}{\tiny$\pm$212} & \textbf{0.25}{\tiny$\pm$0.0} & \textbf{16769}{\tiny$\pm$310} & \textbf{0.33}{\tiny$\pm$0.0} & 0.61{\tiny$\pm$0.0} & \textbf{0.18}{\tiny$\pm$0.0} \\
\bottomrule
\end{tabular}%
}
\end{table*}

\textit{Finding 1: JT models improve point estimation even under independent generation.} Do-PFN JT and UWYK No-Anc JT significantly improve upon their baselines on average in both unit and population level metrics, with a lower error on all ten cells. UWYK Anc JT and CausalPFN-C JT have a lower error on six of the ten cells, but on average neither differs significantly from its baseline.

\subsection{Synthetic}
\label{sec:synthetic}

Both synthetic benchmarks generate the two outcomes under dependence, and their cross-world correlation can be computed for most datasets (Table~\ref{tab:rho-all}). On the case studies the correlation $\rho \approx 1$ wherever it can be computed, hence $\rho = 1$ matches the cross-world correlation of the data-generating process. On ComplexMech, $\rho$ is computable for fewer datasets and varies across them, with means between $0.82$ and $1$ for most graph sizes, making $\rho = 0.5$ and $\rho = 1$ the closest approximations among the values we assume. We focus on the interval score, which reflects the dependence that joint learning targets, and report it in Figures~\ref{fig:cs-is} and~\ref{fig:cm-is}. Due to space constraints, point estimation is reported in Appendix~\ref{app:cs-cm-point}. In brief, joint learning also significantly improves it for Do-PFN and UWYK, reducing $\sqrt{\mathrm{PEHE}}$ on average by $40$--$62\%$ and $11$--$51\%$, respectively. For CausalPFN-C, whose baseline learns the expected outcomes without the idiosyncratic noise (Section~\ref{sec:performance}), the baseline holds a slight advantage, and the JT model is at most $9\%$ worse.

\begin{figure}[tbp]
\centering
\includegraphics[width=\textwidth]{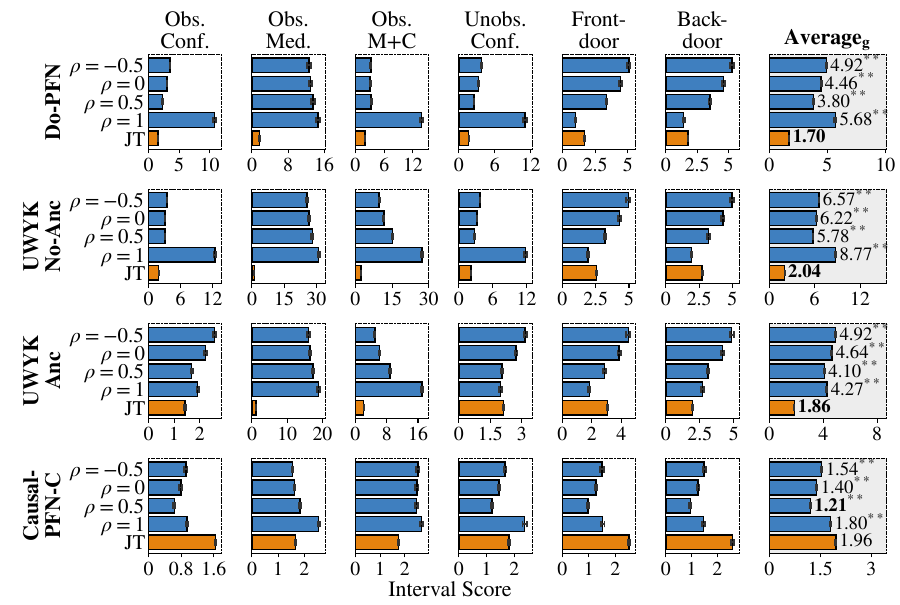}
\caption{\small Interval score $\mathrm{IS}_{0.05}/\sigma_Y$ (lower is better) of the 95\% posterior predictive interval for the \textbf{ITE} on the six causal scenarios of the Do-PFN case studies. Each row is one model family: its single-outcome model under each assumed $\rho$ (blue) and its JT model with the learned dependence (orange). Bars show the mean over realizations $\pm$ the standard error. The last column, Average$_g$, is the geometric mean over datasets $\pm$ its standard error, with the lowest value in bold. Significance against the JT model at the $0.1$ and $0.05$ level is indicated by $^{*}$ and $^{**}$, respectively. Each panel has its own scale.}
\label{fig:cs-is}
\end{figure}

\begin{figure}[tbp]
\centering
\includegraphics[width=\textwidth]{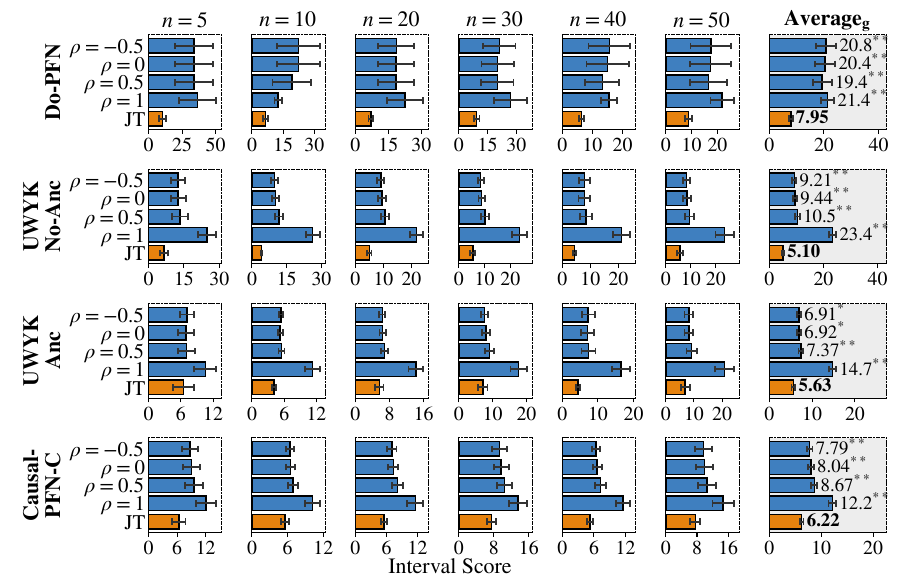}
\caption{\small Interval score $\mathrm{IS}_{0.05}/\sigma_Y$ on ComplexMech, per graph size $n$, as in Figure~\ref{fig:cs-is}. }
\label{fig:cm-is}
\end{figure}

\textit{Finding 2: When the dependence is present, joint learning significantly improves the posterior predictive intervals at every tested $\rho$; on the case studies, the improvement for CausalPFN-C occurs only at matched resolution.} On the case studies, the Do-PFN and UWYK JT models significantly improve the interval score of their baselines under every tested $\rho$, by $55$--$77\%$. This holds even when the baselines are given the true cross-world correlation, $\rho = 1$. For CausalPFN-C, the baseline attains the better interval score on the case studies, since the reduced resolution costs the JT model the most, but once the resolution is matched, the JT model is significantly better under every tested $\rho$, by $6$--$44\%$ (Appendix~\ref{app:resolution-ablation}). On ComplexMech, this large resolution cost of CausalPFN-C is outweighed by the gains of joint learning. Hence, every JT model significantly improves the interval score of its baseline by $18$--$78\%$ under every tested $\rho$.

\subsection{Real}
\label{sec:twins-results}

On Twins, we follow \citet{bodik2026retro}: for each pair, one twin is observed and the other, its counterfactual, is predicted from it. The observed twin is informative about the other because the two share unobserved characteristics, which induce dependence between them. A single-outcome model can use the observed twin only through an assumed cross-world correlation, which the observed data do not identify. We therefore evaluate it over the range of $\rho$ compatible with the observed data. We evaluate the models on the dequantized outcomes with $d = 1$ and $d = 10$ covariates and report the mean squared error only, since the binary nature of the outcome leaves the interval score uninformative even after dequantization.

\textit{Finding 3: On real paired outcomes, joint learning on average significantly outperforms a baseline that selects its cross-world correlation uniformly from the range compatible with the data.} A JT model predicts the counterfactual twin by conditioning its joint on the observed twin, with the dependence taken from the SCMs that fit $\mathcal{D}_{\mathrm{obs}}$ (Section~\ref{sec:performance}). The baseline must instead assume $\rho$, and since $\mathcal{D}_{\mathrm{obs}}$ carries no information about it, any value within the compatible range is a random guess. Averaged over these guesses, every JT model significantly reduces the mean squared error of its baseline at both covariate sets, by $22$--$48\%$ (Table~\ref{tab:twins-mse}). The baseline reaches the JT model only when its guess falls mostly near the reported range of $0.5$--$0.7$ \citep{bodik2026}, which requires knowing the outcomes that are being predicted.

\begin{table*}[tbp]
\centering
\caption{\small Counterfactual prediction on dequantized Twins: mean squared error of the counterfactual twin given the factual twin and $d$ covariates, averaged over three jitter widths. The baseline is averaged over $\rho$ drawn uniformly from the range compatible with the data. Mean $\pm$ standard error, lower is better. Bold marks the better model of each family. Significance at the $0.1$ and $0.05$ level is indicated by $^{*}$ and $^{**}$, respectively.}
\label{tab:twins-mse}
\small
\setlength{\tabcolsep}{3pt}
\resizebox{\textwidth}{!}{%
\begin{tabular}{l cc cc cc cc}
\toprule
 & \multicolumn{2}{c}{Do-PFN} & \multicolumn{2}{c}{UWYK No-Anc} & \multicolumn{2}{c}{UWYK Anc} & \multicolumn{2}{c}{CausalPFN-C} \\
\cmidrule(lr){2-3}\cmidrule(lr){4-5}\cmidrule(lr){6-7}\cmidrule(lr){8-9}
 & Baseline & JT & Baseline & JT & Baseline & JT & Baseline & JT \\
\midrule
$\mathrm{MSE}_{d=1}\downarrow$ & 0.206\,{\scriptsize$\pm$0.00} & \textbf{0.141}$^{**}$\,{\scriptsize$\pm$0.00} & 0.202\,{\scriptsize$\pm$0.00} & \textbf{0.117}$^{**}$\,{\scriptsize$\pm$0.00} & 0.212\,{\scriptsize$\pm$0.00} & \textbf{0.118}$^{**}$\,{\scriptsize$\pm$0.00} & 0.210\,{\scriptsize$\pm$0.00} & \textbf{0.164}$^{**}$\,{\scriptsize$\pm$0.00} \\
$\mathrm{MSE}_{d=10}\downarrow$ & 0.203\,{\scriptsize$\pm$0.00} & \textbf{0.131}$^{**}$\,{\scriptsize$\pm$0.00} & 0.202\,{\scriptsize$\pm$0.00} & \textbf{0.116}$^{**}$\,{\scriptsize$\pm$0.00} & 0.222\,{\scriptsize$\pm$0.00} & \textbf{0.115}$^{**}$\,{\scriptsize$\pm$0.00} & 0.213\,{\scriptsize$\pm$0.00} & \textbf{0.160}$^{**}$\,{\scriptsize$\pm$0.00} \\
\bottomrule
\end{tabular}%
}
\end{table*}

\section{Discussion and Conclusion}

Although causal prior-data fitted networks are a step forward, at the current stage they do not solve causal inference. We discuss what this means for their evaluation and training.

Causal inference carries strict identifiability limits. The Markov equivalence class and the function class of the underlying SCM cannot be recovered from observational data \citep{spirtes2000, peters}. Two PFN models of equal capacity can favour different members of the same equivalence class, and point estimation alone cannot separate them. Evaluation should therefore prioritise uncertainty estimation, with coverage and density estimation given a higher standing as standard metrics in the field.

Beyond how we evaluate these models, what we evaluate them on should also be scrutinized. Real datasets rarely offer an objectively accepted ground truth for causal inference, so evaluation mostly falls to semi-synthetic and synthetic benchmarks. There, what one believes about the connection of the factual and counterfactual worlds shapes the design of the data generation process. This belief can instead be informed by the evidence in the literature, which indicates that the two worlds are rarely independent and are typically positively correlated \citep{bodik2026,andrews2021insights,vanlancker2024chasing}. The Twins \citep{louizos2017causaleffectinferencedeep} dataset is one of the motivating applications for this. Benchmarks may therefore benefit from spanning levels of this dependence rather than assuming independence, bringing the setup closer to reality.

This connection between the two worlds should also be carried to the training side. Jointly learning both factual and counterfactual worlds for every unit captures their dependence. In our results, this significantly improves point estimation and posterior predictive intervals in most settings, for some models even with bins up to $32$ times wider. Hence, under a limited budget, joint learning should be favoured, provided that the learning regime and the standardization of the outcomes keep the resolution cost moderate enough for the model to capture the dependence. If the budget is not limited, raising the resolution removes this cost and only strengthens the case for joint learning (Appendix~\ref{app:resolution-ablation}). As our results suggest, providing the same amount of data is not equivalent: seeing the same number of worlds does not substitute for learning them together.

Taken together, these considerations provide three main recommendations for where the field should move. Evaluation should measure the uncertainty that identifiability leaves behind, benchmarks should carry the connection between the worlds that reality carries, and training should learn the two worlds jointly. As evaluation and benchmarks move in this direction, the gains of joint learning would only become clearer. Although the field has much progress ahead, we conclude that, with its learning approach closer to reality, joint learning overall is a preferable training setup for causal prior-data fitted networks.

\subsection*{Acknowledgments}
This material is based in part upon work supported
by the German Federal Ministry of Education and Research (BMBF): Tübingen AI Center, FKZ: 01IS18039B; by the Machine Learning Cluster of Excellence, EXC number 2064/1 – Project number 390727645; by NSERC Discovery Grant RGPIN-2025-06491; by an NSERC Canada Graduate Research Scholarship – Master's (CGRS M); by the University of Toronto’s Acceleration Consortium, which receives funding from the Canada First Research Excellence Fund (CFREF); Resources used in preparing this research project were provided, in part, by the Digital Research Alliance of Canada; the Province of Ontario; the Swiss AI Compute grants; the Government of Canada through CIFAR; and companies sponsoring the Vector Institute. We thank Rahul Shrestha, Yongjin Yang, Erivan Inan, and Angelo Huang for their helpful comments. We further acknowledge the work of AdminBot for administrative support that streamlined our research workflow.

\FloatBarrier   %
\bibliography{references}

\section*{AI use statement}

In this work, we used generative AI tools (Claude Opus 5.5) to assist in writing and formatting, editing software code, summarizing existing literature, and targeted information retrieval in papers verified by the authors.

We did not use generative AI tools to generate datasets, develop theoretical models or conceptual frameworks, formulate mathematical claims or critical ingredients for proofs, assist in writing proofs, propose or refine research hypotheses, or design or provide feedback on the research methodology or experiments. The rest of the required disclosure tasks are not applicable to this work. Synthetic data used in training and evaluation were generated using the benchmark mechanisms described in Sections~\ref{sec:results}, \ref{sec:outcomes} and Appendix~\ref{app:benchmarks} rather than by generative AI tools.

We have reviewed all AI-assisted work. AI-generated code was verified and tested for correctness before any interpretation of results. We take responsibility for the final content of this work, including text, claims or artifacts produced with the aid of generative AI.

\section*{Reproducibility statement}

We provide the methodological and experimental details needed to reproduce the work throughout the paper and appendices. Section~\ref{sec:joint-head} and Appendix~\ref{app:head-spec} specify the joint learning architecture and training objective, while Appendix~\ref{app:proofs} provides the proofs of the theoretical results. Appendices~\ref{app:malc} and~\ref{app:ot} describe the ITE- and ATE-density inference procedures, including optimal-transport aggregation. Section~\ref{sec:results} and Appendix~\ref{app:benchmarks} describe the experimental setup and benchmark construction, including modifications to existing benchmarks, and Appendix~\ref{app:metrics} defines the evaluation protocol and metrics. Appendix~\ref{app:additional} provide the additional results and ablations, including the matched-resolution, matched-label, correlation, and checkpoint comparisons. Finally, Appendix~\ref{app:compute} reports the computational environment, software versions, hardware, model sizes, training steps, throughput, and compute cost. The code is available on \href{https://github.com/FurkanDanisman/CFM-for-Decision-Makers}{GitHub}.

\appendix

\section{Overview}

The appendix is organised as follows. Appendix~\ref{app:head-spec} details the joint output distribution and its training loss. Appendix~\ref{app:proofs} contains the theoretical results, namely the proof of Lemma~\ref{lem:sufficiency}, the extension of the results of \citet{dopfn} to the joint distribution, and the comparison of the joint and single-outcome losses. Appendix~\ref{app:malc} describes how the ITE density is obtained for each model type, and Appendix~\ref{app:ot} how the ATE density is obtained by optimal transport and calibrated on the model's prior. Appendix~\ref{app:metrics} defines the evaluation protocol and metrics, and Appendix~\ref{app:benchmarks} the construction of the four benchmarks. Appendix~\ref{app:additional} collects the additional results, including the interval score on RealCause, the point estimation on the case studies and ComplexMech, the counterfactual prediction on Twins, the ATE density on the benchmarks, the matched-resolution and matched-label ablations, the robustness to the cross-world correlation, and the comparisons against the published CausalPFN and re-trained UWYK checkpoints. Appendix~\ref{app:compute} lists the computational details.

\section{Joint Output Distribution}
\label{app:head-spec}

The joint distribution predicted by a jointly trained model (Equation~\ref{eq:joint-ppd}) must return a
finite density at every point of $\mathbb{R}^{2}$ so that the log-loss
is well-posed on unbounded outcomes, while remaining
expressive enough on the bounded region $\mathcal{I} = [a, b]^{2}$ \footnote{Outcomes are rescaled so that the observed range of $Y_T$ maps to $[a, b]$. Each baseline and its JT variant receive identically rescaled outcomes, so differences between them do not come from the scaling. The rescaling follows each model's own evaluation pipeline, computed on the outcomes of both treatment groups together: Do-PFN's own pipeline for Do-PFN, a min--max rescaling for UWYK, and a standardization for CausalPFN-C.}

\paragraph{Grid and indexing.}
Partition $\mathcal{I}$ into a uniform $K \times K$ grid of square bins
of width $h \coloneqq (b - a)/K$. Bins are indexed by
$(j_{0}, j_{1}) \in \{0, \ldots, K-1\}^{2}$; bin $B_{j_{0}, j_{1}}$ is
the $h \times h$ square with lower-left corner
$\bigl(a + j_{0}h,\; a + j_{1}h\bigr)$. For any
$y \in [a, b]$ and coordinate $k \in \{0, 1\}$, the containing bin index
along that coordinate is
\begin{equation}
  j_{k}(y) \;\coloneqq\; \lfloor (y - a) / h \rfloor.
  \label{eq:bin-index}
\end{equation}

\paragraph{Per-query outputs.}
For each query the transformer emits $K^{2} + 9 + 4$ real numbers,
grouped into three tensors mapped to normalised parameters by fixed
link functions:
\begin{itemize}
  \item \emph{Interior logits} $\ell \in \mathbb{R}^{K^{2}}$ softmaxed
    into a $K \times K$ probability table over the bins,
    \begin{equation}
      p_{j_{0}, j_{1}}(X, \mathcal{D}_{\mathrm{obs}})
      \;=\;
      \frac{\exp(\ell_{j_{0}, j_{1}})}
           {\sum_{i_{0}, i_{1}} \exp(\ell_{i_{0}, i_{1}})};
      \label{eq:pmat-softmax}
    \end{equation}
  \item \emph{Region logits} $u \in \mathbb{R}^{9}$ softmaxed into
    non-negative mixture weights across the nine regions defined below,
    \begin{equation}
      w_{r}(X, \mathcal{D}_{\mathrm{obs}})
      \;=\; \frac{\exp(u_{r})}{\sum_{s = 1}^{9} \exp(u_{s})},
      \qquad r \in \{1, \ldots, 9\};
      \label{eq:region-weights}
    \end{equation}
  \item \emph{Raw tail scales}
    $\bigl(\sigma_{L}^{(0), \mathrm{raw}},\; \sigma_{R}^{(0), \mathrm{raw}},\;
           \sigma_{L}^{(1), \mathrm{raw}},\; \sigma_{R}^{(1), \mathrm{raw}}\bigr)$,
    one for each coordinate--direction pair, passed through softplus
    with a small floor $\varepsilon > 0$ and scaled by $h$,
    \begin{equation}
      \sigma_{d}^{(k)}(X, \mathcal{D}_{\mathrm{obs}})
      \;=\; h \cdot
      \bigl(\mathrm{softplus}\bigl(\sigma_{d}^{(k), \mathrm{raw}}\bigr)
            + \varepsilon\bigr),
      \qquad k \in \{0, 1\},\; d \in \{L, R\}.
      \label{eq:tail-scales}
    \end{equation}
\end{itemize}
The $h$ prefactor in~\eqref{eq:tail-scales} makes the tail scales
compatible with the interior resolution: doubling $K$ halves both the
bin width and the initial tail width, so the density is invariant under
resolution changes at initialisation.

\paragraph{Nine regions.}
The plane $\mathbb{R}^{2}$ is partitioned into nine regions in three groups: (1)~the interior square $\mathcal{I}$, where both $y_{0}, y_{1} \in [a, b]$; (2)~four edge strips, where exactly one of $y_{0}, y_{1}$ lies outside $[a, b]$; and (3)~four corner quadrants, where both lie outside. Every query point by default falls in one region, so at inference the
mixture weight $w_{r}$ selects a single component; only its density
contributes to $f$ and to $\log f$.

\paragraph{Interior region ($y_{0}, y_{1} \in [a, b]$).}
The histogram is converted to a piecewise-constant density by dividing
by the bin area,
\begin{equation}
  f_{\mathrm{inner}}(y_{0}, y_{1} \mid X, \mathcal{D}_{\mathrm{obs}})
  \;=\;
  \frac{p_{j_{0}(y_{0}),\, j_{1}(y_{1})}}{h^{2}}.
  \label{eq:inner-density}
\end{equation}
The $h^{2}$ factor is the Jacobian from bin probability to density
on $[a, b]^{2}$.

\paragraph{Edge regions ($y_{i} \in [a, b]$, $y_{1-i} \notin [a, b]$).}
Consider the representative subcase $y_{0} \in [a, b]$ and $y_{1} < a$;
the other three edges are symmetric. We factor the edge density as
$p(y_{1})\, p(y_{0} \mid y_{1})$, model the marginal $p(y_{1})$ by a
half-Gaussian anchored at the boundary, and approximate the conditional
$p(y_{0} \mid y_{1})$ by the inner region at the boundary:
\begin{equation}
  f_{\mathrm{edge}}(y_{0}, y_{1} \mid X, \mathcal{D}_{\mathrm{obs}})
  \;=\;
  p_{L}^{(1)}(y_{1})\, g(y_{0} \mid y_{1} = a),
  \label{eq:mix-density}
\end{equation}
with
\begin{equation}
  p_{L}^{(1)}(y_{1})
  \;=\;
  \frac{2}{\sigma_{L}^{(1)}}\,
  \phi\!\left(\frac{y_{1} - a}{\sigma_{L}^{(1)}}\right),
  \qquad
  g(y_{0} \mid y_{1} = a)
  \;=\;
  \frac{p_{j_{0}(y_{0}),\, 0}}
       {\bigl(\sum_{i_{0}} p_{i_{0},\, 0}\bigr)\, h},
  \label{eq:boundary-conditional}
\end{equation}
where $\phi$ is the standard normal density. 

Holding $p(y_{0} \mid y_{1})$ fixed at $y_{1} = a$ is exact when $Y_{0} \perp Y_{1}$ and otherwise
introduces a bias that grows with the strength of dependence and the
distance past the boundary; for a bivariate normal with correlation
$\rho$ and tail scale $\sigma$ the excess NLL per edge observation is
approximately $\rho^{2}\sigma^{2} / \pi(1 - \rho^{2})$. The expression grows without bound as $|\rho| \to 1$. The approximation, however, enters only the edge regions, which hold the observations with exactly one outcome outside the grid. 

\paragraph{Corner regions ($y_{0}, y_{1} \notin [a, b]$).}
Let $(s_{0}, s_{1}) \in \{L, R\}^{2}$ select the quadrant with corner
$(c_{0}, c_{1})$, with $c_k = a$ if $s_k = L$ and $c_k = b$ if $s_k = R$. The corner density is a bivariate
normal centred at the corner, with correlation $\rho$ (defined below)
and standardised coordinates $\tilde{y}_{k} = (y_{k} - c_{k}) / \sigma_{s_{k}}^{(k)}$:
\begin{equation}
  f_{\mathrm{corner}}(y_{0}, y_{1} \mid X, \mathcal{D}_{\mathrm{obs}})
  \;=\;
  \frac{1}{Z(s_{0}, s_{1}, \rho)}\,
  \phi_{2}\!\bigl(\tilde{y}_{0}, \tilde{y}_{1};\, \rho\bigr).
  \label{eq:tail-density}
\end{equation}
The truncation constant $Z$ is the probability mass the bivariate normal
places in the quadrant selected by $(s_{0}, s_{1})$; by Sheppard's
identity,
\begin{equation}
  Z(s_{0}, s_{1}, \rho)
  \;=\;
  \begin{cases}
    \tfrac{1}{4} + \tfrac{\arcsin \rho}{2\pi}
      & \text{if } s_{0} = s_{1}, \\[4pt]
    \tfrac{1}{4} - \tfrac{\arcsin \rho}{2\pi}
      & \text{if } s_{0} \neq s_{1}.
  \end{cases}
  \label{eq:sheppard}
\end{equation}

\paragraph{Correlation inherited from the interior.}
The correlation $\rho$ in \eqref{eq:tail-density}--\eqref{eq:sheppard}
is not a separate trained parameter by design choice. It is defined as the Pearson correlation of the discrete distribution $\{p_{j_{0}, j_{1}}\}$ evaluated at bin midpoints
$m_{k}(j_{k}) \coloneqq a + h\bigl(j_{k} + \tfrac{1}{2}\bigr)$:
\begin{equation}
\begin{aligned}
\bar{y}_{k}
  \;&=\; \sum_{j_{0}, j_{1}} p_{j_{0}, j_{1}}\, m_{k}(j_{k}), \\
v_{k}
  \;&=\; \sum_{j_{0}, j_{1}} p_{j_{0}, j_{1}}
         \bigl(m_{k}(j_{k}) - \bar{y}_{k}\bigr)^{2}, \\
c
  \;&=\; \sum_{j_{0}, j_{1}} p_{j_{0}, j_{1}}
         \bigl(m_{0}(j_{0}) - \bar{y}_{0}\bigr)
         \bigl(m_{1}(j_{1}) - \bar{y}_{1}\bigr), \\
\rho
  \;&=\; \frac{c}{\sqrt{v_{0}\, v_{1}} + \varepsilon}.
\end{aligned}
\label{eq:rho-derived}
\end{equation}
Tying $\rho$ to the interior grid removes one degree of freedom, keeps
the corner coupling consistent with what the interior region already
represents, and prevents the tail and interior components from
disagreeing about the dependence between $Y_{\mathrm{do}(0)}$ and
$Y_{\mathrm{do}(1)}$.

\paragraph{Full mixture and training loss.}
The joint density is
\begin{equation}
  f(y_{0}, y_{1} \mid X, \mathcal{D}_{\mathrm{obs}})
  \;=\;
  w_{1}\, f_{\mathrm{inner}}
  \;+\; \sum_{r = 2}^{5} w_{r}\, f^{(r)}_{\mathrm{edge}}
  \;+\; \sum_{r = 6}^{9} w_{r}\, f^{(r)}_{\mathrm{corner}},
  \label{eq:full-density}
\end{equation}
and the negative log-density loss is evaluated at the region of the
observed label $(y_{0}^{\star}, y_{1}^{\star})$; the region indicator
selects a single component, so only its weight and parameters receive
gradient at that point. Because the region assignment is deterministic
in the label, the mixture weights $w$ are trained purely from which
region observations fall into, while the interior grid, tail scales,
and derived correlation are trained within each region.

\section{Theoretical Results}
\label{app:proofs}

\subsection{Proof of Lemma~\ref{lem:sufficiency}}

\begin{lemma}
\label{lem:sufficiency}
Let $Z_i^{\mathrm{R}} \coloneqq \big(X_i,\, Y_{\mathrm{do}(0)}(x_i, \varepsilon_i),\, Y_{\mathrm{do}(1)}(x_i, \varepsilon_i)\big)$ denote the per-unit observation under the joint objective and $Z_i^{\mathrm{U}} \coloneqq \big(X_i,\, T_i,\, Y_{\mathrm{do}(T_i)}(x_i, \varepsilon_i)\big)$ the per-unit observation under the single-outcome objective, with $T_i$ drawn from any distribution on $\{0, 1\}$ independent of $\varepsilon_i$. Then $Z^{\mathrm{U}}$ is a randomised function of $Z^{\mathrm{R}}$: sampling $T$ from the same distribution and setting $Z^{\mathrm{U}} = \big(X, T, Y_{\mathrm{do}(T)}\big)$ recovers the law of the single-outcome observation from that of the paired one. Consequently, any parameter of the joint law $P^\star\big(Y_{\mathrm{do}(0)}, Y_{\mathrm{do}(1)} \mid X\big)$ estimable from $Z^{\mathrm{U}}$ is estimable from $Z^{\mathrm{R}}$ at least as well.
\end{lemma}

\begin{proof}
Let $\lambda$ denote the law of $T$ on $\{0, 1\}$, chosen independently
of $\varepsilon_{i}$. Define the measurable map
$\Phi(x, y_{0}, y_{1};\, t) \coloneqq (x, t, y_{t})$; then for any
$Z^{\mathrm{R}}$ with joint law $\PP^{\star}$,
\begin{equation}
  Z^{\mathrm{U}} \;\overset{d}{=}\; \Phi\!\bigl(Z^{\mathrm{R}};\, T\bigr),
  \qquad T \sim \lambda \perp Z^{\mathrm{R}},
\end{equation}
recovers the law of the marginal observation from the joint one. The
map is measurable and marginalises over $T$ without reference to
$\PP^{\star}$, so the experiment observing $Z^{\mathrm{U}}$ is a garbling of the one observing $Z^{\mathrm{R}}$. By Blackwell's comparison of experiments, any procedure based on $Z^{\mathrm{U}}$ can be matched by one based on $Z^{\mathrm{R}}$, and the Fisher information about any parameter of $\PP^{\star}$ is no larger under $Z^{\mathrm{U}}$; hence any parameter estimable from $Z^{\mathrm{U}}$ is estimable from $Z^{\mathrm{R}}$ at least as well.
\end{proof}

\subsection{{Extension of the results of \citet{dopfn}}}
\label{app:dopfn-extension}
\citet{dopfn} factorise the posterior predictive distribution of an interventional outcome as $p\big(y \mid x, \mathrm{do}(t), \mathcal{D}_{\mathrm{obs}}\big) = \iint p\big(y \mid x, \mathrm{do}(t), \psi\big)\, p(\psi \mid r)\, p(r \mid \mathcal{D}_{\mathrm{obs}})\, dr\, d\psi$, where $r$ is the representative of the observational equivalence class of $\psi$ (their Equation~4), and show in their Proposition~3 that, for observational data drawn i.i.d.\ from an SCM $\psi_0$, this distribution converges weakly to $p\big(y \mid x, \mathrm{do}(t), [\psi_0]\big)$ as the number of observations grows. Neither result uses the form of the target beyond two properties: it takes values in a complete separable metric space, and, given $\psi$, its distribution does not depend on $\mathcal{D}_{\mathrm{obs}}$. Both hold for the pair $\big(Y_{\mathrm{do}(0)}, Y_{\mathrm{do}(1)}\big)$, which takes values in $\mathbb{R}^2$ and, given $\psi$, is generated by the two interventions from the same exogenous noise, independently of $\mathcal{D}_{\mathrm{obs}}$. Replacing $y$ by $(y_0, y_1)$ in their factorisation gives the same three sources of uncertainty for Equation~\ref{eq:joint-ppd}: the noise within an SCM, the unidentifiability within an equivalence class, and the epistemic uncertainty about the class. Their proof of Proposition~3 then applies with Borel sets $B \subseteq \mathbb{R}^2$, where the two tower steps and Doob's martingale convergence theorem give the weak convergence of $p(y_0, y_1 \mid x, \mathcal{D}_{\mathrm{obs}})$ to $p(y_0, y_1 \mid x, [\psi_0])$.

\subsection{Comparison of the joint and single-outcome losses}

\begin{proposition}
\label{prop:effect-cost}
Let $p$ be the distribution of $\big(Y_{\mathrm{do}(0)}, Y_{\mathrm{do}(1)}\big)$ given $x$, with finite second moments, marginals $p_0$ and $p_1$, variances $\sigma_0^2$ and $\sigma_1^2$, correlation $\rho$, and $\delta = Y_{\mathrm{do}(1)} - Y_{\mathrm{do}(0)}$. For a prediction whose mean of $\delta$ is off by $e$, let $C_{\mathrm{JT}}(e)$, $C_{\mathrm{ML}}(e)$ and $C_{\mathrm{Base}}(e)$ be the smallest expected loss increase per unit under the joint loss, $\mathrm{KL}(p \,\|\, q)$, under the matched-label loss, $\mathrm{KL}(p_0 \,\|\, q_0) + \mathrm{KL}(p_1 \,\|\, q_1)$, and under the single-outcome loss, which scores one outcome per unit, each with probability $1/2$, $\tfrac{1}{2}\mathrm{KL}(p_0 \,\|\, q_0) + \tfrac{1}{2}\mathrm{KL}(p_1 \,\|\, q_1)$. To second order in $e$,
\begin{equation*}
C_{\mathrm{JT}}(e) = \frac{e^2}{2\,\mathrm{Var}_p(\delta)}, \qquad C_{\mathrm{ML}}(e) = \frac{e^2}{2(\sigma_0^2 + \sigma_1^2)}, \qquad C_{\mathrm{Base}}(e) = \frac{e^2}{4(\sigma_0^2 + \sigma_1^2)}.
\end{equation*}
Since $\mathrm{Var}_p(\delta) = \sigma_0^2 + \sigma_1^2 - 2\rho\,\sigma_0\sigma_1$, $C_{\mathrm{JT}}(e) > C_{\mathrm{ML}}(e)$ whenever $\rho > 0$, and since $\mathrm{Var}_p(\delta) \le (\sigma_0 + \sigma_1)^2 \le 2(\sigma_0^2 + \sigma_1^2)$, $C_{\mathrm{JT}}(e) \ge C_{\mathrm{Base}}(e)$ for every $\rho$.
\end{proposition}

\begin{proof}
To second order, $\mathrm{KL}(p \,\|\, q) = \tfrac{1}{2}\int (q - p)^2/p$. Minimising this subject to $\int (q - p) = 0$ and $\int \delta\,(q - p) = e$ gives $q - p = c\,p\,(\delta - \mathbb{E}_p\delta)$ with $c = e/\mathrm{Var}_p(\delta)$, and the value $e^2/\big(2\,\mathrm{Var}_p(\delta)\big)$. For a single outcome, the same argument gives the cost $e_t^2/(2\sigma_t^2)$ of a shift $e_t$ in its mean. Minimising $e_0^2/(2\sigma_0^2) + e_1^2/(2\sigma_1^2)$ subject to $e_1 - e_0 = e$ gives $e^2/\big(2(\sigma_0^2+\sigma_1^2)\big)$ for the matched-label loss, and the single-outcome loss is half of it.
\end{proof}

\section{ITE Density Estimation}
\label{app:malc}

This section details how the ITE density $f_{\delta}$ is obtained for the single-outcome and the jointly trained models.

\paragraph{Single-outcome models.}
A single-outcome model outputs the two marginal densities $\hat{p}(y \mid x, do(0))$ and $\hat{p}(y \mid x, do(1))$ separately, with CDFs $\hat{F}_0$ and $\hat{F}_1$. Since it does not learn their dependence, a value of $\rho$ has to be assumed. Under a Gaussian copula with parameter $\rho$, the ITE density is
\begin{equation}
f_{\delta}(d) = \int \hat{p}(y + d \mid x, do(1)) \, \hat{p}(y \mid x, do(0)) \, c_\rho\big(\hat{F}_1(y + d), \hat{F}_0(y)\big) \, dy,
\end{equation}
where, with $a = \Phi^{-1}(u)$ and $b = \Phi^{-1}(v)$,
\begin{equation}
c_\rho(u, v) = \frac{1}{\sqrt{1 - \rho^2}} \exp\Big(-\frac{\rho^2 (a^2 + b^2) - 2\rho a b}{2(1 - \rho^2)}\Big).
\end{equation}
At $\rho = 0$, $c_\rho = 1$ and the ITE density reduces to the convolution of the marginals under independence, the default of these models, computed on the grid as the discrete cross-correlation of the two bin vectors. At $\rho = 1$, it is the comonotone limit.

\paragraph{Why the true $\rho$ is not sufficient.} The copula is applied to the predicted distributions of the two outcomes, whose spread contains both the noise and the model's uncertainty about $\mu_0$ and $\mu_1$. The true correlation $\rho(x)$ of Equation~\ref{eq:rho} describes only the noise, so even when it is known, it does not in general describe how the two predicted outcomes depend on each other. A simple example helps us illustrate this. Let $Y_{\mathrm{do}(t)} = \mu_t + \varepsilon$ with shared noise $\varepsilon \sim \mathcal{N}(0, 1)$, so that the true correlation is $\rho(x) = 1$, and let the model's uncertainty about $\mu_1$ and $\mu_0$ be independent, with variance $1$ each. Since the shared noise cancels in the ITE, the correct variance of the ITE is $\mathrm{Var}(\mu_1 - \mu_0) = 2$. Coupling the two predicted distributions at the true value $\rho = 1$, however, gives a variance of $0$, and assuming independence gives $4$. Giving a single-outcome model the true $\rho(x)$ therefore does not recover the correct ITE distribution. The dependence it would need varies with the model's uncertainty about each $\mathcal{D}_{\mathrm{obs}}$, and a single-outcome PFN outputs only the two separate predicted distributions, so this dependence can neither be specified in advance nor recovered from its output. The joint model does not face this problem, since it learns the two outcomes together and therefore learns their dependence as well.

\paragraph{Jointly trained models.}
A jointly trained model outputs $\hat{p}(y_{\mathrm{do}(0)}, y_{\mathrm{do}(1)} \mid x, \mathcal{D}_{\mathrm{obs}})$ directly, so the ITE density requires no assumption on the dependence. It is the pushforward of the joint under the difference map, as in Equation~\ref{eq:ite-integral}, and it is computed from the two parts of its output. On the interior $K \times K$ grid, of bin width $h$ with centres $(y_0^i, y_1^j)$, the model places probability $p_{ij}$, summing to one over the interior, on cell $(i, j)$ and spreads it uniformly over the cell. Under the difference, each cell becomes a triangle of half-width $h$ centred at $y_1^j - y_0^i$. The remaining mass lies in the eight tail regions of Equation~\ref{eq:full-density}, and each tail component $f^{(r)}$ becomes the density $f^{(r)}_\delta(d) = \int f^{(r)}(y,\, y + d)\, dy$. The ITE density is therefore the mixture
\begin{equation}
    f_\delta(d) \;=\; w_1 \sum_{ij} p_{ij}\, \tfrac{1}{h}\, \Lambda\!\Big(\tfrac{d - (y_1^j - y_0^i)}{h}\Big) \;+\; \sum_{r=2}^{9} w_r\, f^{(r)}_\delta(d), \qquad \Lambda(u) = \max(0,\, 1 - |u|),
    \label{eq:ite-density}
\end{equation}
with the region weights $w_r$ of Equation~\ref{eq:full-density}. The learned dependence enters through the joint probabilities $p_{ij}$.

\section{ATE Density via Optimal Transport}
\label{app:ot}

\begin{figure}[t]
    \centering
    \includegraphics[width=\textwidth]{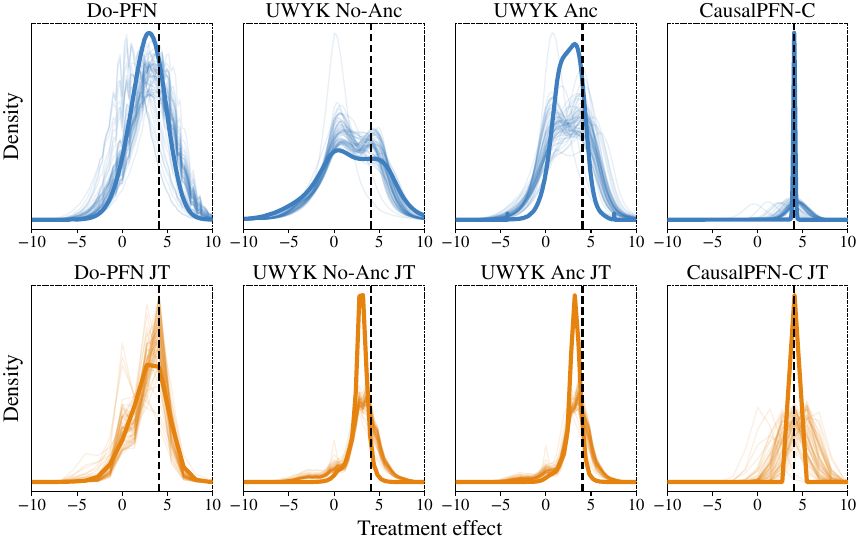}
    \caption{\small \textbf{ATE} density via optimal transport on one \textbf{IHDP} realization. Faint curves are per-query ITE densities, the bold curve the ATE density, and the dashed line the true ATE. Top row shows the single-outcome models, bottom row their jointly trained (JT) variants.}
    \label{fig:ot-illustration}
\end{figure}

Section~\ref{sec:info-gain} defines the ATE density
$f_{\mathrm{ATE}}$ as the scaled, uniformly-weighted 2-Wasserstein barycenter
of the per-query ITE densities $\{f_{\delta}^{(i)}\}_{i=1}^{n}$, through its quantile function
in~\eqref{eq:ate-quantile}. Since each $f_\delta^{(i)}$ is continuous (Equation~\ref{eq:ite-density}), its CDF inverts cleanly and the barycenter follows directly. In this segment we report the derivation, the
grid-level algorithm, the calibration of the scale, and finally an illustration of the process (Figure~\ref{fig:ot-illustration}).

\paragraph{Notation.}
Let $F_{i}$ and $Q_{i} \coloneqq F_{i}^{-1}$ be the CDF and quantile
function of the per-query ITE density $f_{{\delta}}^{(i)}$, and let
$\alpha \in \Delta^{n-1}$ be nonnegative weights summing to $1$. On
$\mathbb{R}$ the weighted 2-Wasserstein barycenter of
$\{f_{{\delta}}^{(i)}\}_{i=1}^{n}$ admits the closed-form quantile
representation \citep{w2}
\begin{equation}
  Q^{\star}(u)
  \;=\;
  \sum_{i=1}^{n} \alpha_{i}\, Q_{i}(u),
  \qquad
  u \in (0, 1),
  \label{eq:agueh-carlier}
\end{equation}
and the barycenter in~\eqref{eq:ate-quantile} is its uniform-weight case $\alpha_{i} = 1/n$, before scaling.

\paragraph{Grid-level implementation.}
Because $Q^{\star}$ is defined explicitly by~\eqref{eq:agueh-carlier},
no iterative optimisation is needed. Given the per-query densities
sampled on a common grid $x_{1} < \cdots < x_{M}$ with spacing
$\Delta x$, the barycenter density is produced in three quantile
passes and one density-recovery pass:
\begin{enumerate}[label=\arabic*., leftmargin=1.8em]
  \item Discretise the probability axis by $\tau_{j} \coloneqq
    (j - \tfrac{1}{2}) / M_{\tau}$ for $j = 1, \dots, M_{\tau}$
    (we use $M_{\tau} = 4001$).
  \item For each query $i \in \{1, \dots, n\}$, compute the CDF by
    trapezoidal integration, renormalise, and invert to obtain the
    quantile function at every $\tau_{j}$:
    \[
      F_{i}(x_{k}) \;\gets\;
        \sum_{\ell \leq k}
        \tfrac{1}{2}\bigl(\hat{p}_{i}(x_{\ell - 1})
        + \hat{p}_{i}(x_{\ell})\bigr)\, \Delta x,
      \quad
      F_{i} \;\gets\; F_{i} / F_{i}(x_{M}),
      \quad
      Q_{i}(\tau_{j}) \;\gets\; F_{i}^{-1}(\tau_{j}).
    \]
  \item Average the quantile functions pointwise per
    \eqref{eq:agueh-carlier}: $Q^{\star}(\tau_{j}) = \sum_{i}
    \alpha_{i}\, Q_{i}(\tau_{j})$.
  \item Recover the barycenter density on the original grid: invert
    $Q^{\star}$ back to a CDF $F^{\star}$, take a central-difference
    derivative, and renormalise to unit integral,
    \[
      \hat{f}_{\mathrm{ATE}}(x_{k}) \;\gets\;
        \frac{F^{\star}(x_{k+1}) - F^{\star}(x_{k-1})}{2\, \Delta x},
      \qquad
      \hat{f}_{\mathrm{ATE}} \;\gets\; \hat{f}_{\mathrm{ATE}} \Big/
        \Bigl(\Delta x \sum_{k} \hat{f}_{\mathrm{ATE}}(x_{k})\Bigr).
    \]
\end{enumerate}
Every step is $\mathcal{O}(n \cdot M)$ vector arithmetic; producing
$\hat{f}_{\mathrm{ATE}}$ therefore adds little cost to computing the per-query
ITE densities it aggregates.

\paragraph{Calibration of the scale.}
The barycenter fixes the location and shape of $f_{\mathrm{ATE}}$, and the factor $s$ in~\eqref{eq:ate-quantile} sets its scale. We set $s$ once per model by split conformal calibration \citep{vovk2005}. For each model, we draw $N = 500$ datasets from its training prior, where the true ATE is known, and compute the barycenter of each. For dataset $j$, let $r_j$ be the smallest factor for which the central $95\%$ interval of the rescaled barycenter contains the true ATE. The scale $s$ is the $\lceil 0.95(N+1) \rceil$-th smallest of $r_1, \dots, r_N$, which guarantees a coverage of at least $95\%$ on new datasets from the same prior. Rescaling around $\bar{\tau}$ leaves the ATE point estimate unchanged. For the single-outcome models, whose ITE densities depend on $\rho$, $s$ is set separately for each $\rho$, and for UWYK separately for the No-Anc and Anc variants. Appendix~\ref{app:inprior} reports the in-prior coverage of the resulting intervals.

\paragraph{Scope and future directions.} The scaled barycenter is a first step toward an ATE density for causal PFNs, and it approximates, rather than equals, the posterior of the ATE. Averaging quantile functions couples the units comonotonically, whereas the exact posterior depends on the joint posterior of the ITEs across units, which share the uncertainty about the SCM. The calibration of $s$ matches the spread of the density to the ATE errors on the model's prior, as the held-out coverage in Table~\ref{tab:inprior-ate} confirms, and the resulting densities let the JT models attain better ATE interval scores than their baselines in $42$ of the $48$ comparisons (Appendix~\ref{app:ate-results}). The calibration, however, does not fix the shape of the density. We therefore see this construction as a practical starting point on which future work can build, for example by modeling the dependence between units or by learning the ATE density directly.

\paragraph{Illustration.}
Figure~\ref{fig:ot-illustration} shows the aggregation on one IHDP realization. The faint curves are the per-query ITE densities, and the bold curve is the ATE density. The barycenter averages the locations of the per-query densities while preserving their shape. The Do-PFN and UWYK JT models concentrate the per-query densities, and their ATE density tightens around the true value. CausalPFN-C JT looks wider than its baseline, mainly because its bin width is $32$ times larger.

\section{Evaluation Metrics}
\label{app:metrics}

\paragraph{Evaluation protocol.}
For each benchmark setting we draw $B$ SCMs $\psi_1,\dots,\psi_B$ independently. From each $\psi_b$ we generate a single observational set $\mathcal{D}^{(b)}_{\mathrm{obs}}$, independently across SCMs, and $Q$ query covariates $x_{b1},\dots,x_{bQ}$, each with a target effect $\tau_b(x_{bq})$ and a population target $\mathrm{ATE}_b$, defined below, and the realized ITE $\delta_{bq}$ of each query unit. Each model is conditioned on $\mathcal{D}^{(b)}_{\mathrm{obs}}$ and returns, for every query, a CATE estimate $\hat\tau_b(x_{bq})$ and an ITE density, and, per SCM, an ATE estimate $\widehat{\mathrm{ATE}}_b$. 

\paragraph{Point estimation.}
 Per SCM we compute
\begin{equation} 
\sqrt{\mathrm{PEHE}}_b = \Bigl(\tfrac{1}{Q}\textstyle\sum_{q=1}^{Q}
   \bigl(\hat\tau_b(x_{bq})-\tau_b(x_{bq})\bigr)^2\Bigr)^{1/2},
   \qquad
   \epsilon_{\mathrm{ATE},b} = \frac{\bigl|\widehat{\mathrm{ATE}}_b-\mathrm{ATE}_b\bigr|}{\max\bigl(\bigl|\mathrm{ATE}_b\bigr|,0.1\bigr)},
\end{equation}
and report their means over the $B$ SCMs; the floor of $0.1$ keeps $\epsilon_{\mathrm{ATE}}$ finite when the true ATE is close to zero, as in the case studies with a zero mean shift. The target $\tau_b(x_{bq})$ is the analytical CATE $\mu_1(x_{bq}) - \mu_0(x_{bq})$ where the data-generating process provides one (IHDP and ACIC); elsewhere it is the effect of the query unit itself. It is the realized ITE $\delta_{bq}$ on CPS, PSID and ComplexMech, and on the case studies. $\mathrm{ATE}_b$ is the corresponding mean. Evaluating against a unit's own effect $\tilde\tau_{bq}$ rather than the CATE $\tau_b(x_{bq})$ does not affect the comparison. Write $\eta_{bq} = \tilde\tau_{bq} - \tau_b(x_{bq})$, so that $\mathbb{E}[\eta_{bq} \mid x_{bq}] = 0$. Since the query unit's outcomes are not seen by the model, $\eta_{bq}$ is independent of $\hat\tau_b(x_{bq})$ given $x_{bq}$, and the cross term vanishes:
\begin{equation}
    \mathbb{E}\big[(\hat\tau_b(x_{bq}) - \tilde\tau_{bq})^2\big] = \mathbb{E}\big[(\hat\tau_b(x_{bq}) - \tau_b(x_{bq}))^2\big] + \mathbb{E}\big[\eta_{bq}^2\big].
\end{equation}
The second term does not depend on the model, so the ranking is unchanged in expectation, and the error against the CATE is smaller than the reported one by $\mathbb{E}[\eta_{bq}^2]$.

\paragraph{Posterior predictive intervals.}
For each query, the $95\%$ posterior predictive interval $\mathrm{PI}_b(x_{bq}) = [l_{bq}, u_{bq}]$ is the central interval between the $2.5\%$ and $97.5\%$ quantiles of the predicted ITE density, and its length is $u_{bq} - l_{bq}$.

\paragraph{Interval score.}
We evaluate these intervals by the interval score \citep{Gneiting01032007} at $\alpha = 0.05$, which combines coverage and length into a single metric,
\begin{equation}
    \mathrm{IS}_{bq} = (u_{bq} - l_{bq}) + \tfrac{2}{\alpha}\,(l_{bq} - \delta_{bq})\,\mathbf{1}\{\delta_{bq} < l_{bq}\} + \tfrac{2}{\alpha}\,(\delta_{bq} - u_{bq})\,\mathbf{1}\{\delta_{bq} > u_{bq}\}.
\end{equation}
It equals the interval length when the realized ITE is covered, and adds a penalty proportional to the distance of a miss otherwise. Within each SCM we average it over queries and normalise it by the outcome scale, yielding one value per observational set,
\begin{equation}
    S_b = \frac{1}{Q \sigma_Y}\sum_{q=1}^{Q} \mathrm{IS}_{bq}.
\end{equation}
The reported score is the mean across SCMs, $\bar S = \frac{1}{B}\sum_{b=1}^{B} S_b$. Since the SCMs and their observational sets are drawn independently, $S_1, \dots, S_B$ are i.i.d., so $\bar S$ is an unbiased estimate of the expected score over the benchmark's SCM distribution, and its standard error is $s_S/\sqrt{B}$, with $s_S$ the sample standard deviation of the $S_b$. Queries within an SCM share one posterior and are therefore dependent; aggregating them into $S_b$ before computing uncertainty accounts for this. The benchmark SCMs are not drawn from the models' training priors, so these intervals carry no coverage guarantee, which holds only on average over SCMs drawn from the prior. The benchmark results therefore show how the models transfer to test data; the in-prior check, where the guarantee holds, is also reported.

\paragraph{Evaluation of CausalPFN.}
The CausalPFN baseline is trained on $m_t$ rather than on the outcomes. This does not change what it predicts in our setting: in our design both outcomes of a unit receive the same idiosyncratic noise draw, which cancels in their difference, so the ITE equals $m_1 - m_0$ and the target of the model is the ITE itself. We therefore treat its prediction as an ITE distribution like every other model. The alternative is not viable either. The true CATE is available only on IHDP and ACIC, so evaluating CausalPFN against it would exclude the model from the uncertainty metrics on most datasets, and would leave the models without a common target.

\paragraph{Averages across datasets.}
For a metric $m$ where lower is better, with values $m_1, \dots, m_K$ on the $K$ datasets or settings of a comparison, we report the geometric mean
\begin{equation}
    \mathrm{Average}_g = \Bigl(\textstyle\prod_{k=1}^{K} m_k\Bigr)^{1/K},
\end{equation}
with its standard error obtained by the delta method. We report $\sqrt{\mathrm{PEHE}}$, the standard metric in the causal inference literature, in its usual units, which differ across datasets. An arithmetic mean of these values would be dominated by the datasets with the largest scale, whereas the ratio of two geometric means does not depend on the scale of any dataset. The percentage improvements in the text are therefore computed from Average$_g$, and do not depend on how the metrics are normalised. The printed Average$_g$ of $\sqrt{\mathrm{PEHE}}$ is standardised by $\sigma_Y$, which leaves these percentages unchanged. The same mean is used for $\epsilon_{\mathrm{ATE}}$ and for $\mathrm{IS}_{0.05}/\sigma_Y$, so that all comparisons are made on the same basis.

\paragraph{Significance.}
For two compared models, let $A_1$ and $A_2$ be their Average$_g$ over the $K$ datasets of a benchmark, computed from the per-dataset means $m_{jk}$ with standard errors $s_{jk}$. By the delta method, the standard error of each Average$_g$ is
\begin{equation}
    \mathrm{SE}(A_j) = \frac{A_j}{K}\Big(\textstyle\sum_{k=1}^{K} \big(s_{jk}/m_{jk}\big)^2\Big)^{1/2}.
\end{equation}
We compare the two by the $z$-statistic and its two-sided $p$-value,
\begin{equation}
    z = \frac{A_1 - A_2}{\sqrt{\mathrm{SE}(A_1)^2 + \mathrm{SE}(A_2)^2}}, \qquad p = 2\big(1 - \Phi(|z|)\big),
\end{equation}
with $\Phi$ the standard normal distribution function. On Twins, the same test is applied to the mean squared error at each covariate set. The baseline's error is first averaged over the values of $\rho$ in the range compatible with the data, then over the three jitter widths, as is the JT model's error over the widths. The standard error of such an average is bounded by the average of the standard errors, which holds whatever the dependence between the averaged terms, and makes the test conservative. The comparisons of one model family within a figure or table, across the assumed values of $\rho$ and, for the ablations, across both ablations, form one family of $M$ tests. Their $p$-values are corrected with the Hochberg procedure. The $p$-values are ordered, $p_{(1)} \le \dots \le p_{(M)}$, and at level $\alpha$ every comparison with $p_{(i)} \le p_{(k)}$ is significant, where $k$ is the largest index with $p_{(k)} \le \alpha/(M - k + 1)$. We apply this at $\alpha = 0.1$ and $\alpha = 0.05$, and mark the significant differences with $^{*}$ and $^{**}$, respectively. The procedure assumes positive dependence among the tests, which holds here, since all comparisons of a family share the same JT model and are evaluated on the same SCMs. The test also treats the two models as independent, although both are evaluated on the same SCMs. As their errors are expected to be positively correlated, this overstates the standard error of the difference, so the test is conservative.

\paragraph{In-prior calibration.}\label{app:inprior}

We report the calibration of our trained models on data drawn from each model's own training prior, where the ground truth is known, with a new SCM drawn from the prior for every dataset. A posterior predictive interval averages over SCMs according to their posterior, so when the posterior is exact, its coverage equals the nominal level on average over SCMs drawn from the prior. This applies to the predicted outcomes of every model and to the ITE of the JT models; the ITE interval of a single-outcome model also depends on how its two predicted outcomes are combined.

Each dataset has 1000 observational units and 500 query units. Whether a query's realized ITE falls in its interval is averaged within a dataset, as queries share a context, and the standard error is taken over datasets. Table~\ref{tab:inprior-new} reports the coverage of each outcome and of the ITE. We report the ITE with shared noise in the two potential outcomes for every model, as the single-outcome models never see the two outcomes of a unit together.

\begin{table}[t]\centering\small\setlength{\tabcolsep}{6pt}
\caption{\small In-prior calibration, new SCM each time. Cells give the coverage of the 95\% interval $\pm$ SE over SCMs, where each SCM gives one dataset. $Y_{\mathrm{do}(1)}$, $Y_{\mathrm{do}(0)}$: the interval of each predicted outcome, from the marginals of the joint for the JT models. UWYK-B: UWYK re-trained with binary treatments (Appendix~\ref{app:binarytraining}). $^{\ast}$UWYK's published checkpoint, trained with continuous treatments and evaluated on its own continuous-treatment prior; its columns are the two outcomes $Y_{\mathrm{do}(t_1)}$, $Y_{\mathrm{do}(t_0)}$ at two treatment levels $t_0 < t_1$ drawn from that prior, and their contrast $Y_{\mathrm{do}(t_1)} - Y_{\mathrm{do}(t_0)}$.}
\label{tab:inprior-new}
\begin{tabular}{@{}l ccc@{}}
\toprule
Model & $Y_{\mathrm{do}(1)}$ & $Y_{\mathrm{do}(0)}$ & ITE \\
\midrule
Do-PFN & 0.954{\tiny$\pm$0.00} & 0.929{\tiny$\pm$0.00} & 0.953{\tiny$\pm$0.01} \\
Do-PFN JT & 0.964{\tiny$\pm$0.00} & 0.949{\tiny$\pm$0.01} & 0.944{\tiny$\pm$0.01} \\
\midrule
UWYK-B & 0.842{\tiny$\pm$0.01} & 0.856{\tiny$\pm$0.01} & 0.837{\tiny$\pm$0.01} \\
UWYK$^{\ast}$ & 0.872{\tiny$\pm$0.01} & 0.879{\tiny$\pm$0.01} & 0.916{\tiny$\pm$0.01} \\
UWYK JT & 0.972{\tiny$\pm$0.00} & 0.971{\tiny$\pm$0.00} & 0.963{\tiny$\pm$0.00} \\
\midrule
CausalPFN-C & 0.943{\tiny$\pm$0.00} & 0.943{\tiny$\pm$0.00} & 0.964{\tiny$\pm$0.00} \\
CausalPFN-C JT & 0.937{\tiny$\pm$0.00} & 0.946{\tiny$\pm$0.00} & 0.946{\tiny$\pm$0.00} \\
\midrule
SCMs ($R$) & 1000 & 1000 & 1000 \\
Queries per SCM ($Q$) & 500 & 500 & 500 \\
Observational units per SCM & 1000 & 1000 & 1000 \\
\bottomrule
\end{tabular}
\end{table}

\begin{figure}[t]
  \centering
  \includegraphics[width=\textwidth]{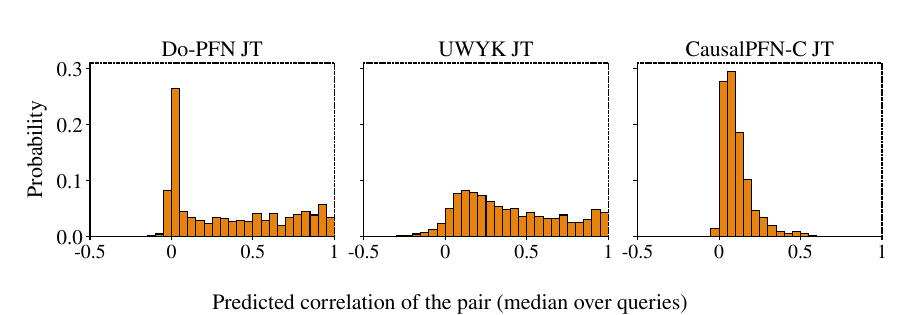}
  \caption{\small Correlation of the predicted pair $\big(Y_{\mathrm{do}(0)}, Y_{\mathrm{do}(1)}\big)$ output by the JT models in their own training prior. Each model is evaluated on 1000 datasets drawn from its training prior; for every query, $\rho$ is the correlation of the predicted joint distribution of $(Y_{\mathrm{do}(0)}, Y_{\mathrm{do}(1)})$, and each dataset contributes the median over its queries. Bars give the share of datasets; titles give the mean $\pm$ sd over datasets.}
  \label{fig:rho-pred-inprior}
\end{figure}

Most models are close to nominal on both the outcomes and the ITE: Do-PFN, Do-PFN JT, UWYK JT, and both CausalPFN-C models lie between $0.93$ and $0.97$. UWYK-B under-covers even in its own prior, with $0.84$--$0.86$ on the outcomes and $0.84$ on the ITE, and so does UWYK$^{\ast}$, with $0.87$--$0.88$ on the outcomes and $0.92$ on the ITE. Figure~\ref{fig:rho-pred-inprior} shows the correlation of the predicted pair $\big(Y_{\mathrm{do}(0)}, Y_{\mathrm{do}(1)}\big)$ that the JT models output on datasets drawn from their training priors: between the $5\%$ and $95\%$ quantiles over datasets, it ranges from about $0$ to above $0.9$ for Do-PFN JT and UWYK JT, and from $0.01$ to $0.30$ for CausalPFN-C JT. The correlation is lower for CausalPFN-C JT for two reasons. First, CausalPFN's prior gives the two outcomes a weaker dependence to learn. In this prior, an outcome is the sum of an expected outcome $m_t$ and a noise term $\eta_t$. In our JT prior, the noise $\eta_t$ is the same for both outcomes of a unit, so it makes them move together. The expected outcomes $m_0$ and $m_1$, however, also vary between units with the same covariates (Section~\ref{sec:cfm}), and this variation is not the same for the two outcomes. In about four out of five datasets drawn from the prior, it accounts for more than half of an outcome's variance given $x$, so only the remaining part comes from the shared noise. The two outcomes of a unit are therefore only moderately correlated in this prior, and the dependence the model can learn is correspondingly weaker. Second, the grid is coarse for this model. On standardized outcomes, each bin is $0.625$ wide, so about half of the training data falls within two bins (Appendix~\ref{app:resolution-ablation}). When both outcomes fall in the same bin, the model cannot tell how they move together, which pulls its predicted correlation down further. Both causes stem from choices made for single-outcome training, which we keep unchanged so that each JT model differs from its baseline only in its learning objective. A prior in which the two worlds share more of their variation, or a grid adapted to joint learning, would let the model learn the cross-world dependence more fully. Designing priors and output distributions for joint learning is therefore a natural direction for future work.

For the ATE density, Table~\ref{tab:inprior-ate} reports the coverage of its 95\% interval on held-out datasets. The $500$ datasets of each model's prior are split into five folds, and the scale $s$ (Appendix~\ref{app:ot}) is set on four of them and evaluated on the fifth. Every model attains close to the nominal level under every $\rho$, between $94.6\%$ and $95.4\%$.

\begin{table}[t]
\centering
\small
\caption{\small In-prior calibration of the \textbf{ATE} density. Cells give the coverage (\%) of its 95\% interval on held-out datasets $\pm$ SE over the $500$ datasets of each model's prior, with the scale $s$ set on the other four of five folds. Single-outcome models under each $\rho$, and JT models with the learned dependence.}
\label{tab:inprior-ate}
\begin{tabular}{l ccccc}
\toprule
& $\rho=-0.5$ & $\rho=0$ & $\rho=0.5$ & $\rho=1$ & JT \\
\midrule
Do-PFN & 95.0\,{\scriptsize$\pm$1.0} & 94.8\,{\scriptsize$\pm$1.0} & 94.8\,{\scriptsize$\pm$1.0} & 95.2\,{\scriptsize$\pm$1.0} & 95.2\,{\scriptsize$\pm$1.0} \\
UWYK No-Anc & 95.4\,{\scriptsize$\pm$0.9} & 95.4\,{\scriptsize$\pm$0.9} & 94.8\,{\scriptsize$\pm$1.0} & 95.0\,{\scriptsize$\pm$1.0} & 94.8\,{\scriptsize$\pm$1.0} \\
UWYK Anc & 95.4\,{\scriptsize$\pm$0.9} & 95.0\,{\scriptsize$\pm$1.0} & 95.2\,{\scriptsize$\pm$1.0} & 94.8\,{\scriptsize$\pm$1.0} & 94.8\,{\scriptsize$\pm$1.0} \\
CausalPFN-C & 95.2\,{\scriptsize$\pm$1.0} & 95.2\,{\scriptsize$\pm$1.0} & 95.0\,{\scriptsize$\pm$1.0} & 94.6\,{\scriptsize$\pm$1.0} & 94.8\,{\scriptsize$\pm$1.0} \\
\bottomrule
\end{tabular}
\end{table}

\section{Benchmark Construction}
\label{app:benchmarks}

\paragraph{RealCause.}

We use the RealCause datasets \citep{neal2020realcause} directly without any modification. Its generative models fit the outcome distribution under each treatment separately and sample the factual and counterfactual outcomes as independent draws. Table~\ref{tab:rho-all} confirms this. PSID (bal.) uses the PSID samples with an observational context that keeps every treated unit and $500$ randomly drawn controls, so its query units and realized ITEs are those of PSID. For UWYK Anc, we supply the ancestral matrix in which the treatment is an ancestor of the outcome and every covariate is an ancestor of both, with all other entries left unknown. This matches the structure of RealCause's generator, which models the treatment given the covariates and the outcome given the covariates and the treatment.

\begin{table*}[t]
  \centering
  \small
  \setlength{\tabcolsep}{5pt}
  \caption{Composition of the six \textbf{case studies} as we vary the observed covariate count $d$. The number of hidden nodes is $\max(1,\lfloor d/10\rfloor)$. They are not considered covariates, so $|X|=d$ in every cell. A node counts as a confounder ($C$) if it is a parent of both $T$ and $Y$, and as a mediator ($M$) if it lies on a $T\rightarrow\cdot\rightarrow Y$ path; subscripts denote observed, unobserved, and hidden nodes.}
  \label{tab:case-study-composition}
  \resizebox{\textwidth}{!}{%
  \begin{tabular}{c c cc c cc cc cc}
    \toprule
    & \makecell{Observed\\Confounder}
    & \multicolumn{2}{c}{\makecell{Backdoor\\Criterion}}
    & \makecell{Observed\\Mediator}
    & \multicolumn{2}{c}{\makecell{Observed Mediator\\+ Confounder}}
    & \multicolumn{2}{c}{\makecell{Unobserved\\Confounder}}
    & \multicolumn{2}{c}{\makecell{Frontdoor\\Criterion}} \\
    \cmidrule(lr){2-2}\cmidrule(r){3-4}\cmidrule(lr){5-5}\cmidrule(lr){6-7}\cmidrule(lr){8-9}\cmidrule(lr){10-11}
    $d$
    & $C_{\text{obs}}$
    & $C_{\text{obs}}$ & $M_{\text{unobs}}$
    & $M_{\text{obs}}$
    & $C_{\text{obs}}$ & $M_{\text{obs}}$
    & $C_{\text{obs}}$ & $C_{\text{unobs}}$
    & $C_{\text{obs}}$ & $M_{\text{unobs}}$ \\
    \midrule
    5  & 5  & 5  & 1 & 5  & 3  & 2  & 5  & 1 & 5  & 1 \\
    10 & 10 & 10 & 1 & 10 & 5  & 5  & 10 & 1 & 10 & 1 \\
    20 & 20 & 20 & 2 & 20 & 10 & 10 & 20 & 2 & 20 & 2 \\
    30 & 30 & 30 & 3 & 30 & 15 & 15 & 30 & 3 & 30 & 3 \\
    40 & 40 & 40 & 4 & 40 & 20 & 20 & 40 & 4 & 40 & 4 \\
    50 & 50 & 50 & 5 & 50 & 25 & 25 & 50 & 5 & 50 & 5 \\
    \bottomrule
  \end{tabular}%
  }
\end{table*}

\paragraph{Case study.}
We take the six causal scenarios of Do-PFN \citep{dopfn}, keeping both the causal graphs and the structural equations. Two types of variations are added. First, a constant is added to the treated outcome equation, shifting the mean ITE to $-2$, $0$, and $+2$; a zero-mean benchmark rewards a model that predicts no effect. Hence, this variation removes that favouritism. Reported numbers pool the three shifts as one dataset. Second, the covariate dimension is varied over $d \in \{5,10,20,30,40, 50\}$ by adding covariates; Table~\ref{tab:case-study-composition} reports the resulting composition per scenario. Both potential outcomes of a unit are generated by propagating the same exogenous draw under $\mathrm{do}(T{=}0)$ and $\mathrm{do}(T{=}1)$, so the noise entering $Y$ is shared by both outcomes and cancels in the ITE; the ITE then varies at fixed $x$ only through hidden variables, noticeably under Unobserved Confounder and negligibly under Backdoor and Frontdoor (Table~\ref{tab:rho-all}).

\paragraph{ComplexMech.}
ComplexMech is adopted from UWYK \citep{uwyk} and samples a fresh SCM with nonlinear mechanisms for every realization. Its graph size $n$ is the number of nodes of the SCM. We modify it by binarizing the treatment. A uniformly selected treatment node is binarized by thresholding at the median of its own sample, which keeps the two treatment groups balanced; the threshold is frozen and reused for the interventional passes, and the observational context is re-drawn after binarization. Realizations without a directed path $T \rightarrow Y$ are discarded, and, as in the case studies, the two interventional passes share the exogenous draw. The original benchmark evaluates interventional distributions and draws no matched pair of potential outcomes, so the ITE is undefined there; sharing the draw defines it as the SCM counterfactual \citep{pearl2009}.

The generator mixes two noise types. In the additive case, $Y = \sigma(f(\mathrm{pa})) + \varepsilon$, the shared noise of $Y$ cancels when the two outcomes are subtracted; in the non-additive case it enters before the activation and does not cancel. Table~\ref{tab:rho-all} gives $\rho(x)$ for the realizations in which it is computable exactly. Because the noise type and the mechanisms are drawn per realization, the correlation differs across realizations.

\begin{table*}[t]
\centering
\caption{\small Cross-world correlation per benchmark. $\rho(x)$ (Eq.~\ref{eq:rho}) is computed exactly from the data-generating process when no covariate is a consequence of the treatment or of the outcome, the assumption $\varepsilon \perp X$ of \citet{bodik2026}: $x$ is fixed and the remaining noise, shared by both outcomes, is redrawn. Check (A): no covariate is a mediator ($T \rightarrow X_k \rightarrow Y$). Check (B): no covariate is a descendant of $Y$. For \textbf{ComplexMech}, whose graph is drawn per dataset, the checks give the number of datasets that pass. $\rho$ is computed on the datasets passing both: per dataset the median of $\rho(x)$ over its queries, then mean $\pm$ sd over datasets. Datasets in which no noise reaches $Y$ have a deterministic ITE given $x$ and are counted as $\rho = 1$. For \textbf{Twins}, the two outcomes belong to different individuals, there is no data-generating process to compute $\rho(x)$ from, and checks (A) and (B) do not apply. Following the statistical computation of \citet{bodik2026}, we report the correlation between the twins' outcomes within the three most frequent value combinations of each pair of the first three covariates, as mean $\pm$ sd over the groups in which it is defined. It is the same on the binary and on the dequantized outcomes (Table~\ref{tab:twins-rho-stat}).}
\label{tab:rho-all}
\small
\setlength{\tabcolsep}{2.5pt}
\resizebox{\textwidth}{!}{%
\begin{tabular}{ll cc c c}
\toprule
Benchmark & Dataset & (A) no mediator & (B) no $Y$-descendant & Computable & $\rho$ (mean $\pm$ sd) \\
\midrule
RealCause & IHDP & $\checkmark$ & $\checkmark$ & all & $0.002 \pm 0.037$ \\
 & ACIC & $\checkmark$ & $\checkmark$ & all & $0.003 \pm 0.013$ \\
 & CPS & $\checkmark$ & $\checkmark$ & all & $-0.001 \pm 0.009$ \\
 & PSID & $\checkmark$ & $\checkmark$ & all & $0.003 \pm 0.020$ \\
\midrule
Case study & Observed Confounder & $\checkmark$ & $\checkmark$ & 1800 / 1800 & $1.000 \pm 0.000$ \\
 & Backdoor & $\checkmark$ & $\checkmark$ & 1800 / 1800 & $1.000 \pm 0.001$ \\
 & Frontdoor & $\checkmark$ & $\checkmark$ & 1800 / 1800 & $1.000 \pm 0.001$ \\
 & Unobserved Confounder & $\checkmark$ & $\checkmark$ & 1800 / 1800 & $0.996 \pm 0.013$ \\
 & Observed Mediator & $\times$ & $\checkmark$ & 0 / 1800 & -- \\
 & Observed Mediator + Confounder & $\times$ & $\checkmark$ & 0 / 1800 & -- \\
\midrule
ComplexMech & $n = 5$ & 53 / 100 & 69 / 100 & 37 / 100 & $0.821 \pm 0.408$ \\
 & $n = 10$ & 32 / 100 & 45 / 100 & 15 / 100 & $0.867 \pm 0.281$ \\
 & $n = 20$ & 12 / 100 & 22 / 100 & 1 / 100 & $0.215$ \\
 & $n = 30$ & 9 / 100 & 14 / 100 & 3 / 100 & $0.958 \pm 0.072$ \\
 & $n = 40$ & 12 / 100 & 12 / 100 & 2 / 100 & $1.000 \pm 0.000$ \\
 & $n = 50$ & 11 / 100 & 14 / 100 & 1 / 100 & $0.946$ \\
\midrule
Twins & {$d = 1$, $d = 10$} & {n/a} & {n/a} & {empirical} & {$0.59 \pm 0.15$} \\
\bottomrule
\end{tabular}}
\end{table*}

\paragraph{Twins.}
\label{app:twins}
Twins \citep{louizos2017causaleffectinferencedeep} contains U.S. twin births from 1989 to 1991, restricted to same-sex pairs with both birth weights below 2\,kg: $11{,}984$ pairs. The treatment considered here is being the heavier twin ($T = 1$) and the outcome is one-year mortality, so under this design both potential outcomes of a pair are observed. We use the dataset as prepared by \citet{bodik2026retro}, in which one twin of each pair is the factual one, and the other is the counterfactual. We also follow their evaluation: given a pair's covariates and the outcome of its factual twin, the outcome of the counterfactual twin is predicted, and the prediction is scored by the mean squared error of the predicted mean against the counterfactual twin's outcome. For UWYK Anc, we supply the ancestral matrix in which the treatment is an ancestor of the outcome and every covariate is an ancestor of both, with all other entries left unknown, as on RealCause. This treats every covariate as a confounder, which might not necessarily reflect the causal structure of Twins. Hence, any performance degradation can be explained with that. It is important to note that we do not know the true causal structure in the dataset, and no ancestral information variant does not suffer from this potential cost.

\textit{Dequantization.} The models predict continuous outcomes, whereas one-year mortality is binary. We therefore add a uniform jitter on $[-w, w]$ to the outcome of each twin, for $w \in \{0.1, 0.25, 0.5\}$, and report results averaged over the three widths. Since $w \le 0.5$, thresholding at $1/2$ recovers the binary outcome. Both the factual outcomes given to the models and the counterfactual outcomes they predict are dequantized. The jitters of the two twins of a pair are coupled through a Gaussian copula, whose correlation is set such that the correlation between the twins' outcomes, given the covariates, equals that of the binary data. Table~\ref{tab:twins-rho-stat} shows that the correlation of Table~\ref{tab:rho-all} is unchanged by the dequantization.

\begin{table}[tbp]
\centering
\caption{\small Correlation between the twins' outcomes, computed following \citet{bodik2026}, within the three most frequent value combinations of each pair of the first three covariates $X_1, X_2, X_3$, which are binary. Binary: on the original $0/1$ mortality outcomes. $w$: half-width of the uniform jitter added to each outcome, with the jitter coupled given the first $d = 10$ or $d = 1$ covariates. A dash marks a group without deaths, in which the correlation is undefined. The last row is the mean $\pm$ sd over the groups in which it is defined.}
\label{tab:twins-rho-stat}
\small
\setlength{\tabcolsep}{3pt}
\begin{tabular}{l c ccc ccc}
\toprule
 & & \multicolumn{3}{c}{$d = 10$} & \multicolumn{3}{c}{$d = 1$} \\
\cmidrule(lr){3-5}\cmidrule(lr){6-8}
Group & Binary & $w = 0.1$ & $0.25$ & $0.5$ & $w = 0.1$ & $0.25$ & $0.5$ \\
\midrule
$X_1 = 0, X_2 = 0$ & 0.653 & 0.652 & 0.654 & 0.644 & 0.653 & 0.648 & 0.651 \\
$X_1 = 0, X_2 = 1$ & 0.646 & 0.646 & 0.636 & 0.647 & 0.644 & 0.654 & 0.648 \\
$X_1 = 1, X_2 = 0$ & 0.241 & 0.277 & 0.368 & 0.481 & 0.291 & 0.277 & 0.385 \\
$X_1 = 0, X_3 = 0$ & 0.587 & 0.588 & 0.594 & 0.611 & 0.586 & 0.592 & 0.613 \\
$X_1 = 0, X_3 = 1$ & 0.690 & 0.689 & 0.683 & 0.671 & 0.691 & 0.691 & 0.677 \\
$X_1 = 1, X_3 = 0$ & -- & 0.015 & 0.462 & 0.364 & 0.135 & 0.289 & 0.350 \\
$X_2 = 0, X_3 = 0$ & 0.573 & 0.575 & 0.586 & 0.600 & 0.575 & 0.573 & 0.598 \\
$X_2 = 0, X_3 = 1$ & 0.703 & 0.701 & 0.700 & 0.678 & 0.704 & 0.701 & 0.691 \\
$X_2 = 1, X_3 = 0$ & 0.607 & 0.607 & 0.605 & 0.625 & 0.602 & 0.622 & 0.638 \\
\midrule
Mean $\pm$ sd & 0.59\,{\scriptsize$\pm$0.15} & 0.59\,{\scriptsize$\pm$0.14} & 0.60\,{\scriptsize$\pm$0.10} & 0.62\,{\scriptsize$\pm$0.06} & 0.59\,{\scriptsize$\pm$0.13} & 0.59\,{\scriptsize$\pm$0.14} & 0.61\,{\scriptsize$\pm$0.10} \\
\bottomrule
\end{tabular}
\end{table}

\textit{Range of $\rho$.} A single-outcome model sees only one twin of each pair. The data therefore cannot tell it how the outcomes of the two twins are correlated. However, we can still limit how large or how small this correlation can be. From the observed twins, we estimate the outcome distribution of each twin given the covariates. The two outcomes are most positively correlated when the twins are ordered in the same way under both distributions, and most negatively correlated when they are ordered in opposite ways. These two cases bound the range of $\rho$ compatible with the data. We compute this range for each pair and take the median over the pairs, which gives $[-0.4, 0.8]$, $[-0.7, 0.9]$ and $[-0.9, 1]$ for $w = 0.1$, $0.25$ and $0.5$, at both covariate sets. The baseline is evaluated over this range in steps of $0.1$.

\textit{Interval score.} On Twins, we report the mean squared error only. For a binary outcome, a $95\%$ interval can contain only survival, death, or both, and it contains both whenever each has a probability above $2.5\%$. A model that predicts a $3\%$ chance of death and one that predicts $70\%$ therefore give the same interval and the same interval score, although their predictions differ greatly. The interval score is thus insensitive to how well the probability of the counterfactual twin's death, given the factual twin, is predicted. Dequantization by large scales might change this, however under the current range it only spreads each outcome over a narrow band around $0$ or $1$. The interval must therefore still cover both bands whichever probability is predicted, and the interval score cannot show whether a model has used the factual twin. The mean squared error, in contrast, scores the predicted probability itself and therefore rewards using it. Hence, on Twins, the interval score remains uninformative even after dequantization, while the mean squared error is more informative about the models' performance, and we report it alone.

\section{Additional Results}
\label{app:additional}

Throughout this appendix, a model is described as better or worse than another only when the difference is significant (Appendix~\ref{app:metrics}). Otherwise, the two are said not to differ significantly.

\subsection{Interval score on RealCause}
\label{app:rc-is}

\begin{figure}[tbp]
\centering
\includegraphics[width=\textwidth]{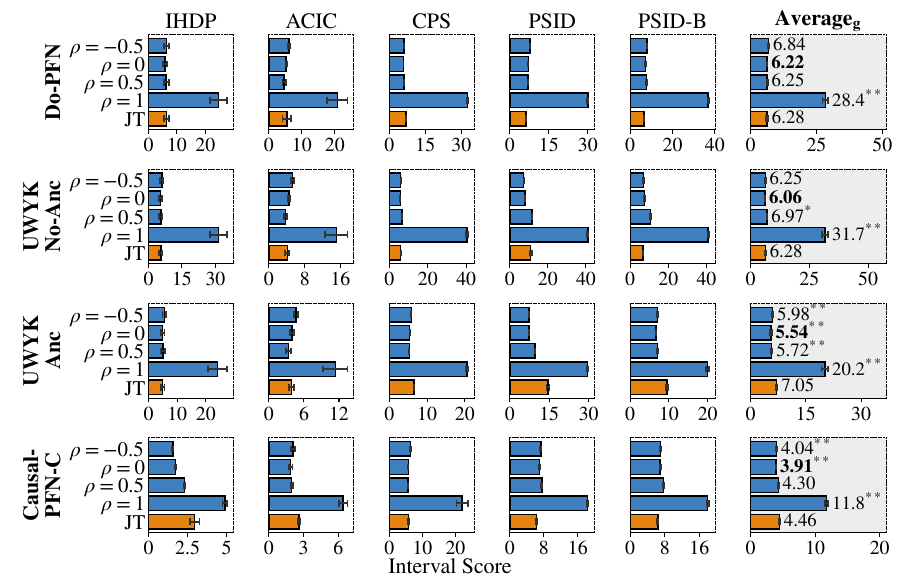}
\caption{\small Interval score $\mathrm{IS}_{0.05}/\sigma_Y$ (lower is better) of the 95\% posterior predictive interval for the \textbf{ITE} on \textbf{RealCause} across five datasets (PSID-B: PSID balanced). Each row is one model family: its single-outcome model under each assumed $\rho$ (blue) and its JT model with the learned dependence (orange). Bars show the mean over realizations $\pm$ the standard error. The last column, Average$_g$, is the geometric mean over datasets $\pm$ its standard error, with the lowest value in bold. Significance against the JT model at the $0.1$ and $0.05$ level is indicated by $^{*}$ and $^{**}$, respectively. Each panel has its own scale.}
\label{fig:rc-is}
\end{figure}

{\textit{Finding 4: When the two outcomes are independent, whether joint learning helps depends on the correlation the baseline assumes.} RealCause generates the two outcomes independently. A baseline that assumes independence is therefore correct, while the JT models, having learned from their prior that the outcomes are dependent, are misspecified. Against this correct baseline, two of the four JT models have a $14$--$27\%$ higher interval score, and the other two do not differ significantly. The further the baseline's assumed correlation moves away from independence, the more misspecified the baseline becomes, and the balance shifts towards joint learning. Under a moderate assumed correlation, $\rho = -0.5$ or $0.5$, the results are mixed, and when the baseline assumes perfect dependence, every JT model improves the interval score by $62$--$80\%$.}

\subsection{Point estimation on the case studies and ComplexMech}
\label{app:cs-cm-point}

\begin{figure}[tbp]
  \centering
  \includegraphics[width=\textwidth]{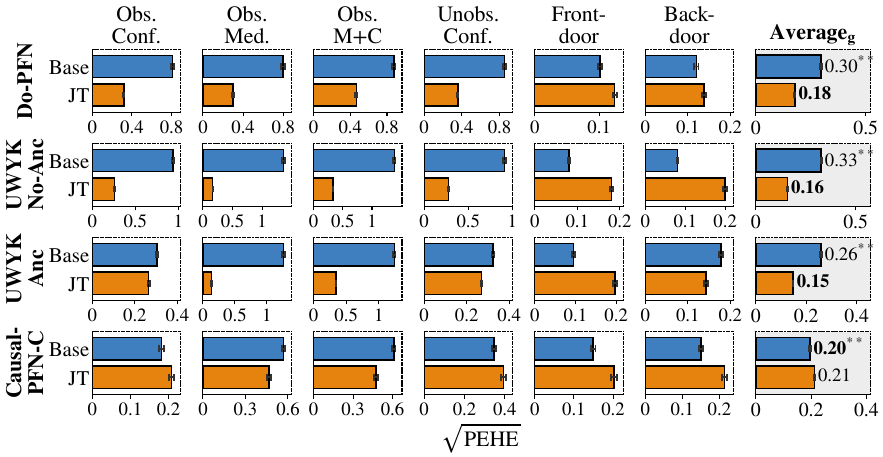}

    \vspace{6pt}
    \includegraphics[width=\textwidth]{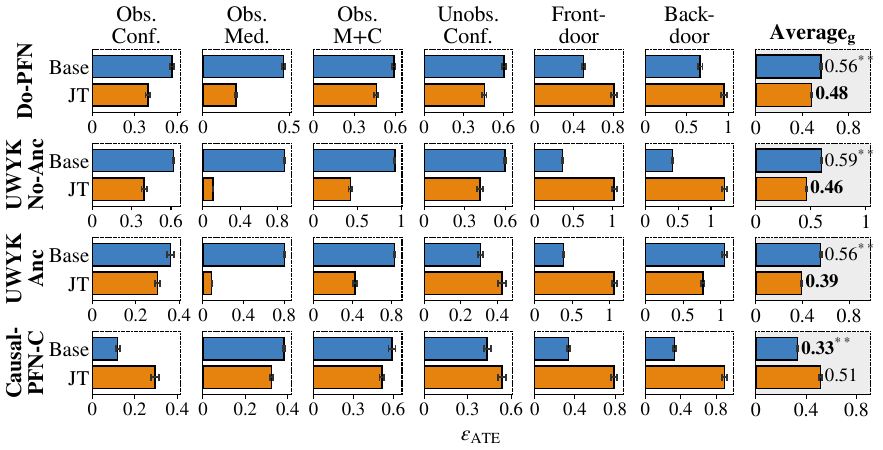}
  \caption{\small $\sqrt{\mathrm{PEHE}}$ (top) and relative ATE error $\epsilon_{\mathrm{ATE}}$ (bottom) on the six \textbf{case studies}; mean $\pm$ standard error, lower is better. Each row is one model family: its baseline (Base, blue) and its JT model (orange); each panel has its own scale. The last column, Average$_g$, is the geometric mean over datasets $\pm$ its standard error, with $\sqrt{\mathrm{PEHE}}$ standardised by each dataset's outcome scale $\sigma_Y$, and the lowest value in bold. Significance against the JT model at the $0.1$ and $0.05$ level is indicated by $^{*}$ and $^{**}$, respectively.}
  \label{fig:pehe-cs-cm}\label{fig:eps-cs-cm}
\end{figure}

\begin{figure}[tbp]
  \centering
  \includegraphics[width=\textwidth]{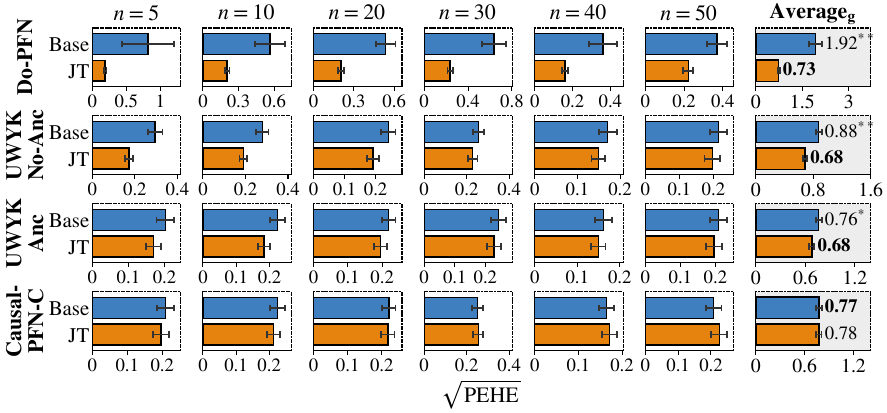}

    \vspace{6pt}
    \includegraphics[width=\textwidth]{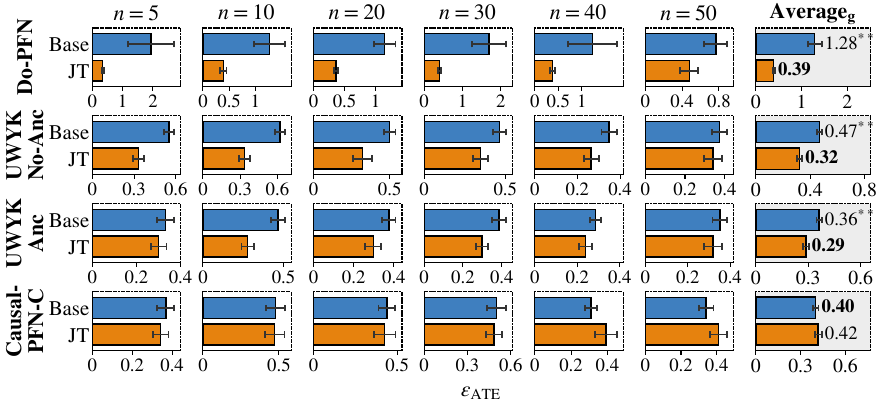}
  \caption{\small $\sqrt{\mathrm{PEHE}}$ (top) and relative ATE error $\epsilon_{\mathrm{ATE}}$ (bottom) on \textbf{ComplexMech} per graph size; mean $\pm$ standard error, lower is better. Each row is one model family: its baseline (Base, blue) and its JT model (orange); each panel has its own scale. The last column, Average$_g$, is the geometric mean over datasets $\pm$ its standard error, with $\sqrt{\mathrm{PEHE}}$ standardised by each dataset's outcome scale $\sigma_Y$, and the lowest value in bold. Significance against the JT model at the $0.1$ and $0.05$ level is indicated by $^{*}$ and $^{**}$, respectively.}
  \label{fig:pehe-cm}\label{fig:eps-cm}
\end{figure}

\textit{Finding 5: When the dependence is present, joint learning significantly improves point estimation for Do-PFN and UWYK, while for CausalPFN the baseline keeps an advantage on the case studies.} On average, joint learning significantly reduces the error of the estimated treatment effects by ${40\%}$ on the case studies and ${62\%}$ on ComplexMech for Do-PFN, and by ${11}${--}${51}\%$ for the UWYK variants. For CausalPFN, whose baseline already removes the idiosyncratic noise $\eta_t$, the JT model does not differ significantly from its baseline on ComplexMech, and its error is significantly higher on the case studies, by $8\%$. The ATE error follows the same pattern for Do-PFN and UWYK, with significant reductions of $15$--$29\%$ on the case studies and $21$--$70\%$ on ComplexMech. For CausalPFN, the ATE error of the JT model is significantly higher on the case studies and does not differ significantly on ComplexMech, so its gain from joint learning is limited to the posterior predictive intervals.

\subsection{Counterfactual prediction on Twins}
\label{app:twins-results}

\begin{figure}[tbp]
\centering
\includegraphics[width=\textwidth]{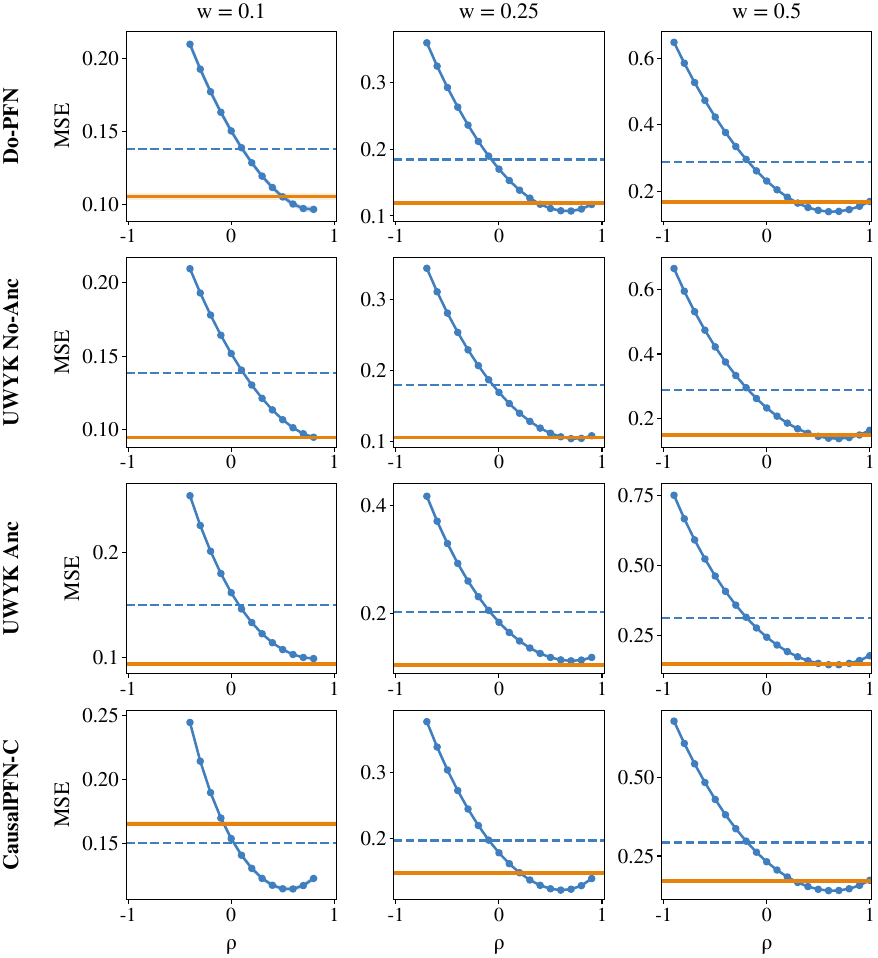}
\caption{\small Mean squared error on Twins at $d = 10$, per jitter width $w$ (columns) and model family (rows): the baseline under each $\rho$ in the range compatible with the data (blue, $\pm$ standard error), its average over the range (dashed), and the JT model (orange, $\pm$ standard error).}
\label{fig:twins-perw-d10}
\end{figure}

\begin{figure}[tbp]
\centering
\includegraphics[width=\textwidth]{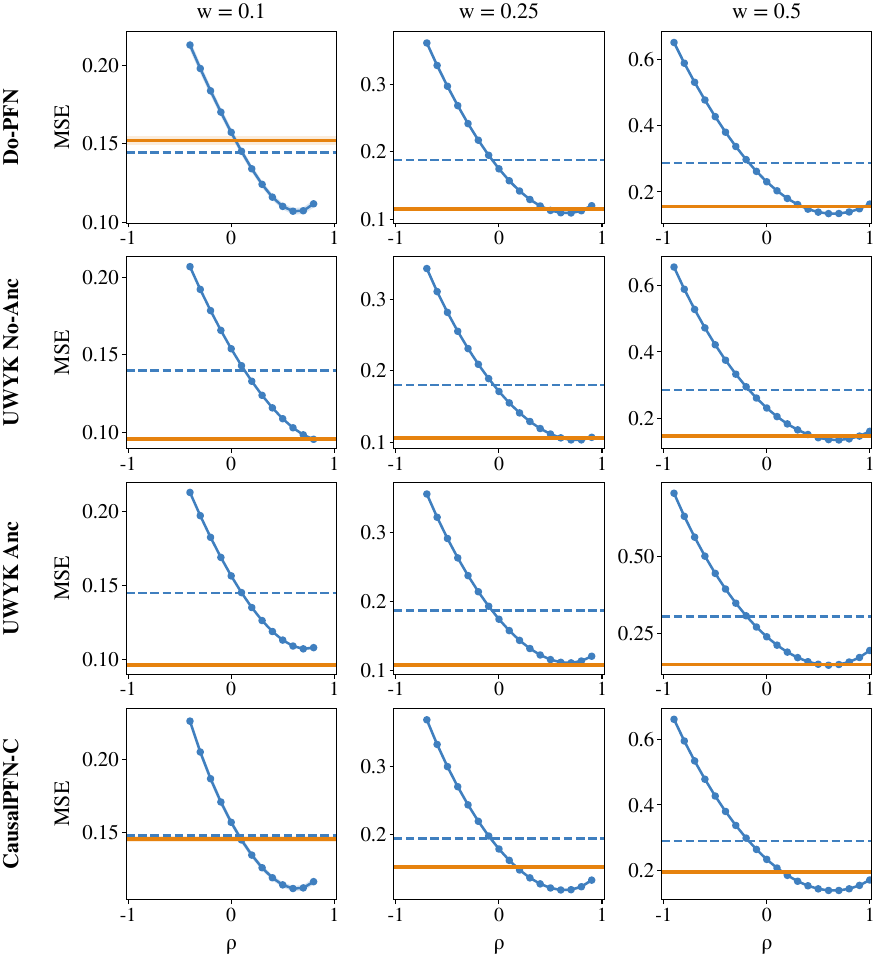}
\caption{\small Mean squared error on Twins at $d = 1$, as in Figure~\ref{fig:twins-perw-d10}.}
\label{fig:twins-perw-d1}
\end{figure}

Figures~\ref{fig:twins-perw-d10} and~\ref{fig:twins-perw-d1} show, for each jitter width, the mean squared error of each baseline under every $\rho$ in the range compatible with the data, together with its average over this range and the error of the JT model. The error of the baseline falls as its assumed $\rho$ gets closer to the reported range of $0.5$--$0.7$ \citep{bodik2026}.

The jitter width enters the comparison in two ways. It adds variance to the outcomes, which both models predict and which therefore raises the error of both. It also widens the range of $\rho$ compatible with the data, since the added spread is not tied together by the observed twins, and a wider range adds strongly negative values to the baseline's average. To ensure that this widening does not drive the result, we repeat the comparison with a single range for all widths, the narrowest one, $[-0.4, 0.8]$, which every width admits. Every JT model remains significantly better than its baseline at both covariate sets, by $7$--$37\%$.

\subsection{ATE density on the benchmarks}
\label{app:ate-results}

\begin{figure}[tbp]
\centering
\includegraphics[width=\textwidth]{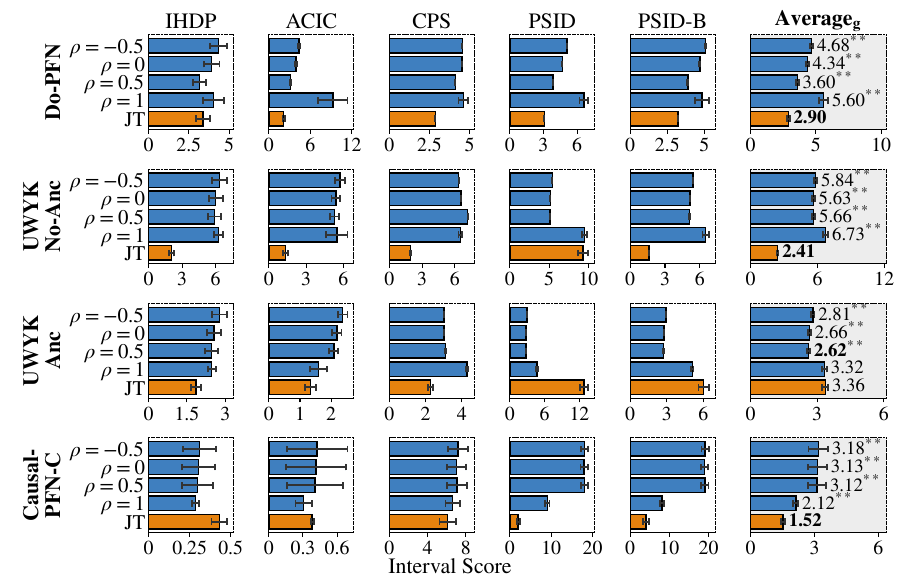}
\caption{\small Interval score $\mathrm{IS}_{0.05}/\sigma_Y$ (lower is better) of the 95\% interval of the \textbf{ATE} density on \textbf{RealCause} across five datasets (PSID-B: PSID balanced). Each row is one model family: its single-outcome model under each $\rho$ (blue) and its JT model with the learned dependence (orange). Bars show the mean over realizations $\pm$ the standard error. The last column, Average$_g$, is the geometric mean over datasets $\pm$ its standard error, with the lowest value in bold. Significance against the JT model at the $0.1$ and $0.05$ level is indicated by $^{*}$ and $^{**}$, respectively. Each panel has its own scale.}
\label{fig:ate-is-rc}
\end{figure}

\begin{figure}[tbp]
\centering
\includegraphics[width=\textwidth]{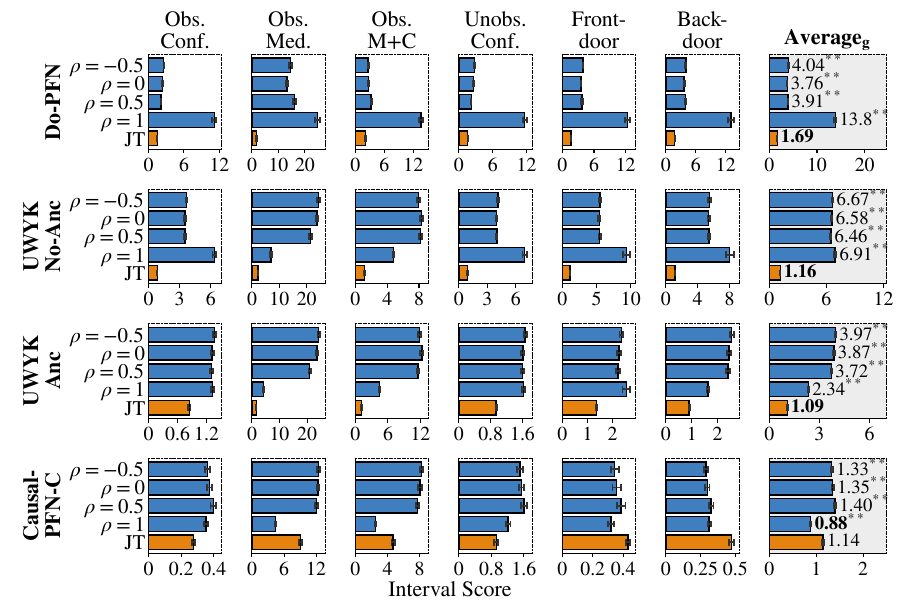}
\caption{\small Interval score $\mathrm{IS}_{0.05}/\sigma_Y$ (lower is better) of the 95\% interval of the \textbf{ATE} density on the six causal scenarios of the Do-PFN case studies. Each row is one model family: its single-outcome model under each $\rho$ (blue) and its JT model with the learned dependence (orange). Bars show the mean over realizations $\pm$ the standard error. The last column, Average$_g$, is the geometric mean over datasets $\pm$ its standard error, with the lowest value in bold. Significance against the JT model at the $0.1$ and $0.05$ level is indicated by $^{*}$ and $^{**}$, respectively. Each panel has its own scale.}
\label{fig:ate-is-cs}
\end{figure}

\begin{figure}[tbp]
\centering
\includegraphics[width=\textwidth]{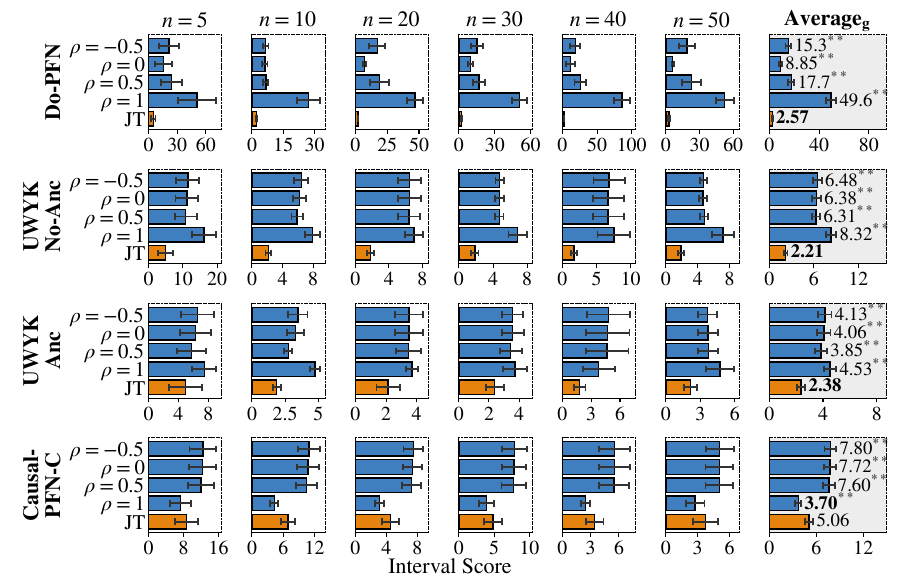}
\caption{\small Interval score $\mathrm{IS}_{0.05}/\sigma_Y$ (lower is better) of the 95\% interval of the \textbf{ATE} density on \textbf{ComplexMech}, per graph size $n$. Each row is one model family: its single-outcome model under each $\rho$ (blue) and its JT model with the learned dependence (orange). Bars show the mean over realizations $\pm$ the standard error. The last column, Average$_g$, is the geometric mean over datasets $\pm$ its standard error, with the lowest value in bold. Significance against the JT model at the $0.1$ and $0.05$ level is indicated by $^{*}$ and $^{**}$, respectively. Each panel has its own scale.}
\label{fig:ate-is-cm}
\end{figure}

Figures~\ref{fig:ate-is-rc}--\ref{fig:ate-is-cm} report the interval score of the 95\% interval of the ATE density (Section~\ref{sec:info-gain}) on RealCause, the case studies and ComplexMech, and Appendix~\ref{app:resolution-ablation} against the matched-resolution and matched-label baselines. Across the four model families, the four values of $\rho$ and the three benchmarks, the JT models attain a significantly lower interval score in $42$ of the $48$ comparisons. The exceptions are UWYK Anc on RealCause, where the outcomes are generated independently, so the baseline is correctly specified under $\rho \le 0.5$, and CausalPFN-C under $\rho = 1$ on the case studies and ComplexMech, where CausalPFN-C JT still attains a significantly lower interval score than its matched-resolution baseline.

\subsection{Matched-resolution and matched-label ablations}
\label{app:resolution-ablation}

In Section~\ref{sec:performance} we discuss that joint learning is more efficient due to two reasons: learning the dependence of outcomes and simply learning more by additional labels. The two ablations separate them. In the matched-resolution version, each baseline is re-trained at its JT counterpart's grid resolution ($K{=}10$ for Do-PFN, $K{=}32$ for UWYK and CausalPFN), removing the resolution advantage of the baseline. In the matched-label version, each baseline is trained on both generated outcomes of every unit, receiving the same data information as the JT model without learning the pair together. {Figures~\ref{fig:coupling-abl-rc}, \ref{fig:coupling-abl-cs}, and~\ref{fig:coupling-abl-cm}} report the calibration with posterior predictive intervals for the ITE and Figures~\ref{fig:ate-abl-rc}--\ref{fig:ate-abl-cm} for the ATE, and {Figures~\ref{fig:ablation-pehe}--\ref{fig:ablation-eps-cm}} the point estimation.

{\textit{Finding 6: Resolution limits CausalPFN-C, and learning the two worlds separately does not substitute for learning them together.} 

On RealCause, where the outcomes are independent, the comparison again depends on the correlation the baselines assume, as in Finding~4. Against their matched-resolution baselines, Do-PFN JT and CausalPFN-C JT are better under three of the four assumed values and do not differ significantly under the fourth. Against the other ablations the results are mixed, and CausalPFN-C JT is worse than its matched-label baseline unless that baseline assumes perfect dependence. 

On the synthetic benchmarks, where the outcomes are dependent, every JT model attains a lower interval score than its matched-resolution baseline under every assumed $\rho$: {$3.2$--$8.3$} times lower for Do-PFN, {$1.3$--$2.7$} for UWYK and {$1.1$--$1.8$} for CausalPFN-C on the case studies, and {$3.2$--$6.4$} and $1.2$--$3.6$ times lower for Do-PFN and UWYK on ComplexMech, where CausalPFN-C JT is comparable to its matched-resolution baseline or ahead of it. Giving the baselines the same data as the JT models does not close the gap either. Against their matched-label baselines, the JT models of Do-PFN and UWYK attain a lower interval score under every assumed $\rho$, {$2.2$--$3.3$} and {$2.9$--$5.8$} times lower on the case studies, and on ComplexMech, UWYK JT is $1.5$--$5.0$ times lower, while Do-PFN JT is lower only under $\rho = 1$, by $2.0$ times. CausalPFN-C JT is $1.3$--$1.9$ times lower on ComplexMech, while on the case studies its matched-label baseline, which retains the fine grid, scores lower. Hence, for both the ITE and the ATE, although the additional data helps, seeing the same number of worlds does not substitute for learning them together.

On Twins, averaged over the range of $\rho$ compatible with the data, every JT model attains a significantly lower mean squared error than both its matched-resolution and its matched-label baseline, at both covariate sets, by $20$--$61\%$ (Figures~\ref{fig:twins-abl-d10} and~\ref{fig:twins-abl-d1}).

It is important to highlight the results for CausalPFN-C particularly. The matched-resolution comparison is the most informative for this model, since its baseline benefits most from the fine grid. During training, CausalPFN outputs the response on a fixed grid over $[-10, 10]$ with a bin width of $\approx 0.02$ on standardized outcomes. For our jointly trained variant the width grows $32$-fold to $0.625$, so about half of the training data falls within two bins; for Do-PFN and UWYK, the share of data within two bin widths is dramatically smaller. Once this advantage is removed at matched resolution, CausalPFN-C JT attains a lower interval score under every assumed $\rho$ on the case studies ($1.1$--$1.8$ times) and under three of the four on RealCause, and a comparable or lower one on ComplexMech. Joint learning therefore improves CausalPFN-C as well; the gain was concealed only by comparing models at different resolutions. Resolution, however, is not a fundamental limitation of joint learning: it can be raised at a higher computational cost or, where that is not feasible, the grid can be adapted to joint learning, for example by narrowing the interior region or changing the standardization of the outcomes. We do not apply such changes here, as our aim is to measure what switching an existing model to joint learning changes, with everything else held fixed.

\begin{figure}[tbp]
\centering
\includegraphics[width=\textwidth]{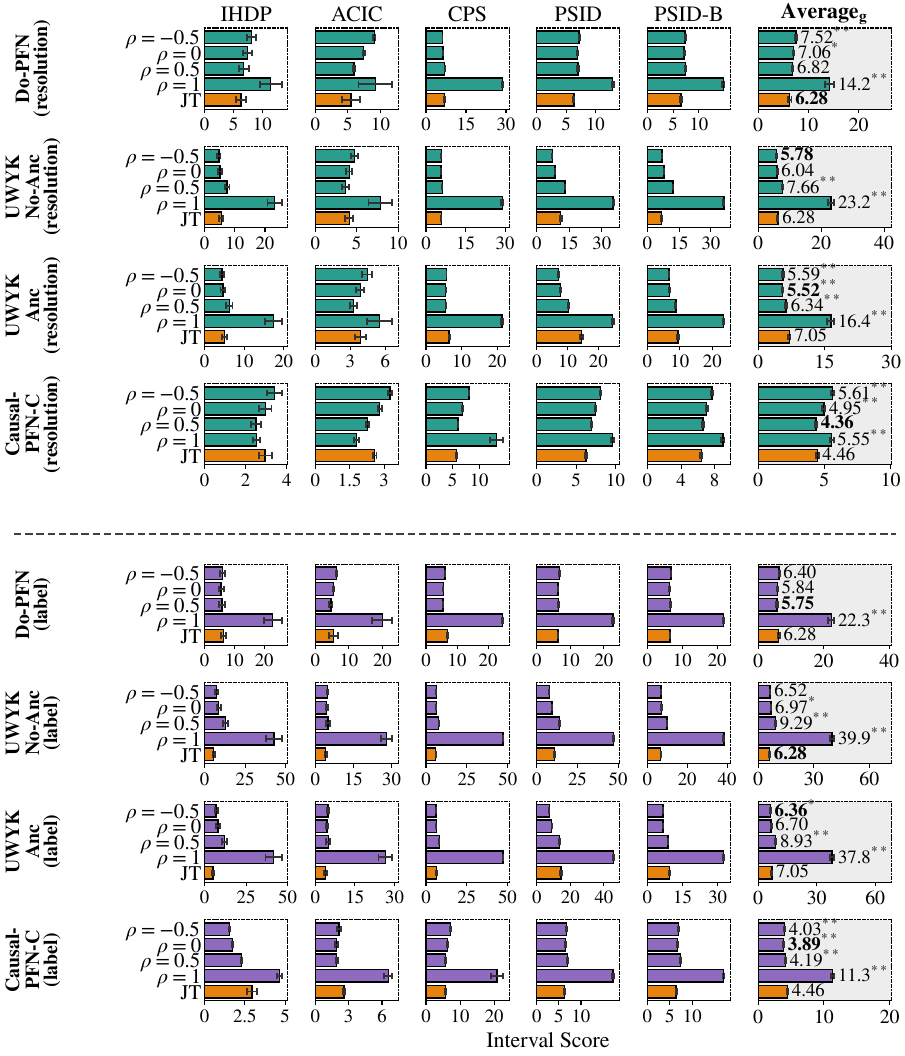}
\caption{\small Coupling sensitivity of the ablation baselines on \textbf{RealCause}: interval score $\mathrm{IS}_{0.05}/\sigma_Y$ (lower is better) of the 95\% posterior predictive interval for the \textbf{ITE}. The rows above the dashed line are the matched-resolution ablations of Appendix~\ref{app:resolution-ablation} (green) and those below are the matched-label ablations (purple), each under every assumed $\rho$, with its family's JT model and the learned dependence (orange). Bars show the mean $\pm$ standard error, and each panel has its own scale. The last column, Average$_g$, is the geometric mean over datasets $\pm$ its standard error, with the lowest value in bold. Significance against the JT model at the $0.1$ and $0.05$ level is indicated by $^{*}$ and $^{**}$, respectively.}
\label{fig:coupling-abl-rc}
\end{figure}

\begin{figure}[tbp]
\centering
\includegraphics[width=\textwidth]{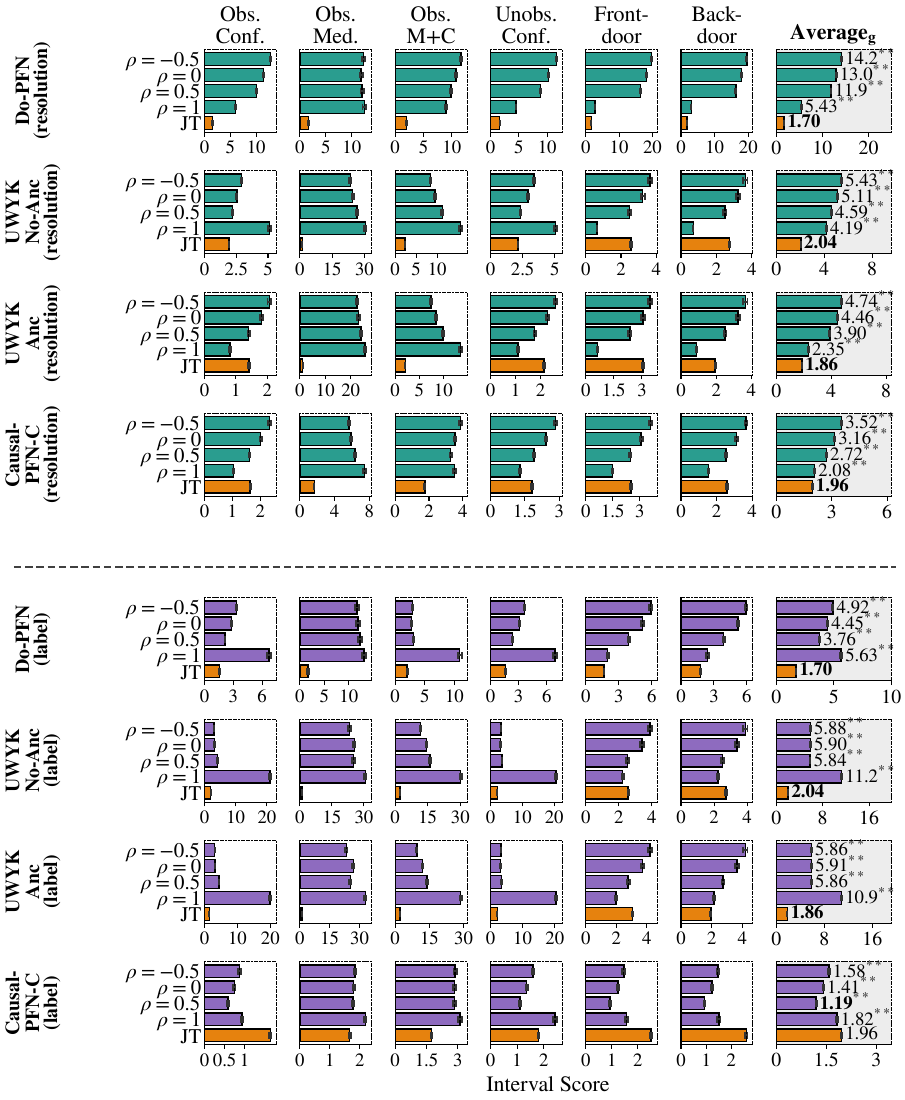}
\caption{\small Coupling sensitivity of the ablation baselines on the six \textbf{case studies}: interval score $\mathrm{IS}_{0.05}/\sigma_Y$ (lower is better) of the 95\% posterior predictive interval for the \textbf{ITE}. The rows above the dashed line are the matched-resolution ablations of Appendix~\ref{app:resolution-ablation} (green) and those below are the matched-label ablations (purple), each under every assumed $\rho$, with its family's JT model and the learned dependence (orange). Bars show the mean $\pm$ standard error, and each panel has its own scale. The last column, Average$_g$, is the geometric mean over datasets $\pm$ its standard error, with the lowest value in bold. Significance against the JT model at the $0.1$ and $0.05$ level is indicated by $^{*}$ and $^{**}$, respectively.}
\label{fig:coupling-abl-cs}
\end{figure}

\begin{figure}[tbp]
\centering
\includegraphics[width=\textwidth]{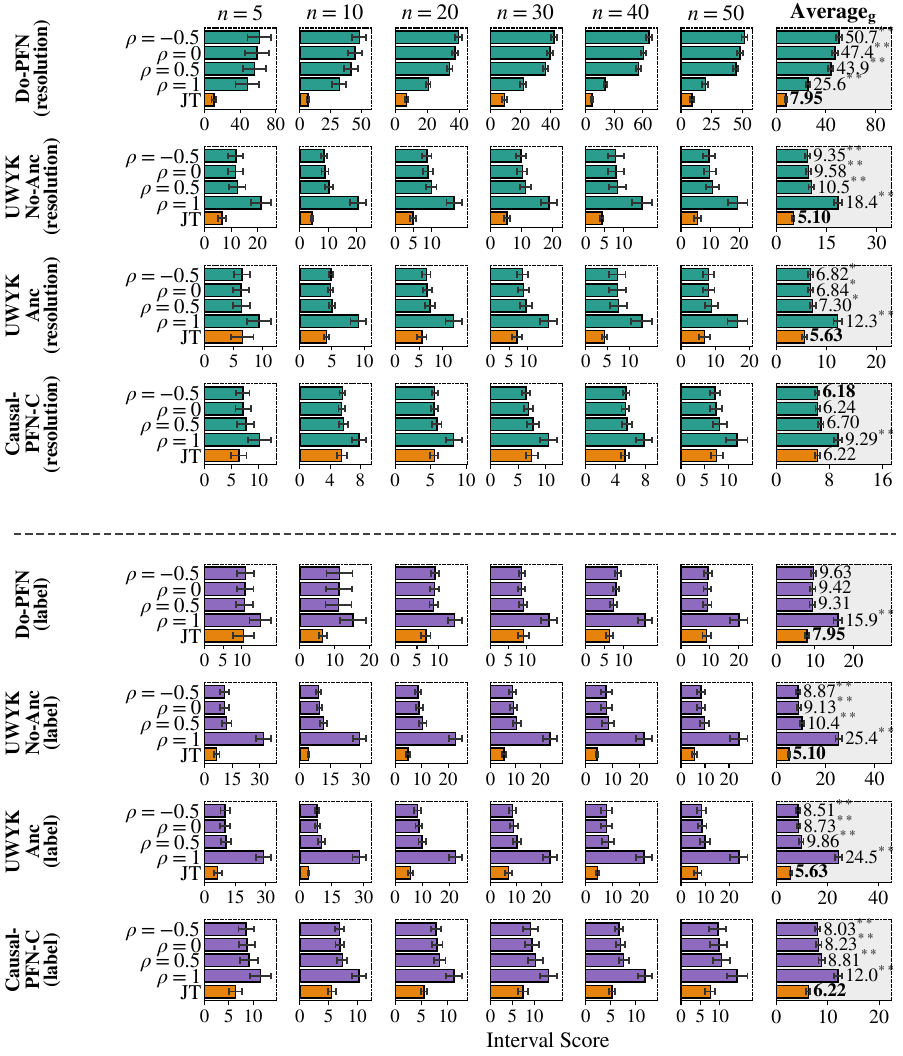}
\caption{\small Coupling sensitivity of the ablation baselines on \textbf{ComplexMech} per graph size $n$: interval score $\mathrm{IS}_{0.05}/\sigma_Y$ (lower is better) of the 95\% posterior predictive interval for the \textbf{ITE}. The rows above the dashed line are the matched-resolution ablations of Appendix~\ref{app:resolution-ablation} (green) and those below are the matched-label ablations (purple), each under every assumed $\rho$, with its family's JT model and the learned dependence (orange). Bars show the mean $\pm$ standard error, and each panel has its own scale. The last column, Average$_g$, is the geometric mean over datasets $\pm$ its standard error, with the lowest value in bold. Significance against the JT model at the $0.1$ and $0.05$ level is indicated by $^{*}$ and $^{**}$, respectively.}
\label{fig:coupling-abl-cm}
\end{figure}

\begin{figure}[tbp]
\centering
\includegraphics[width=\textwidth]{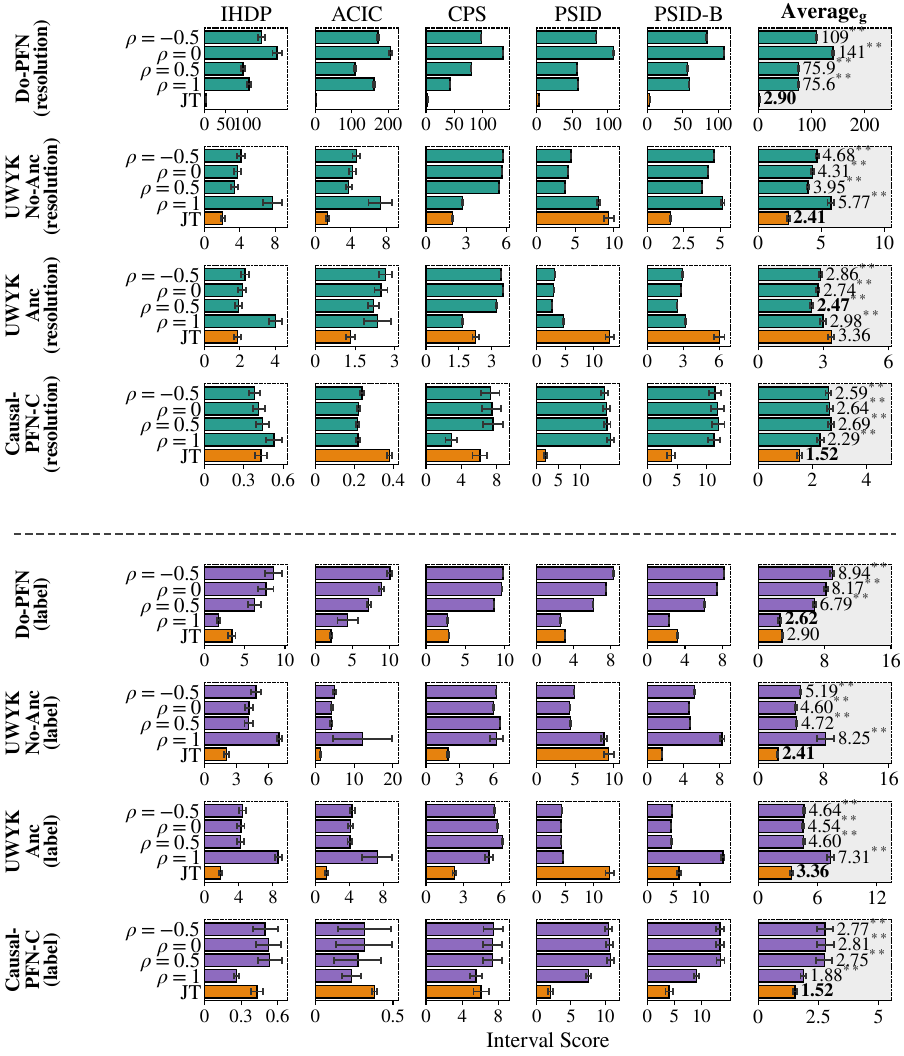}
\caption{\small Interval score $\mathrm{IS}_{0.05}/\sigma_Y$ (lower is better) of the 95\% interval of the \textbf{ATE} density on \textbf{RealCause}, against the ablation baselines. The rows above the dashed line are the matched-resolution ablations (green) and those below are the matched-label ablations (purple), each under every $\rho$, with its family's JT model and the learned dependence (orange). Bars show the mean over realizations $\pm$ the standard error. The last column, Average$_g$, is the geometric mean over datasets $\pm$ its standard error, with the lowest value in bold. Significance against the JT model at the $0.1$ and $0.05$ level is indicated by $^{*}$ and $^{**}$, respectively. Each panel has its own scale.}
\label{fig:ate-abl-rc}
\end{figure}

\begin{figure}[tbp]
\centering
\includegraphics[width=\textwidth]{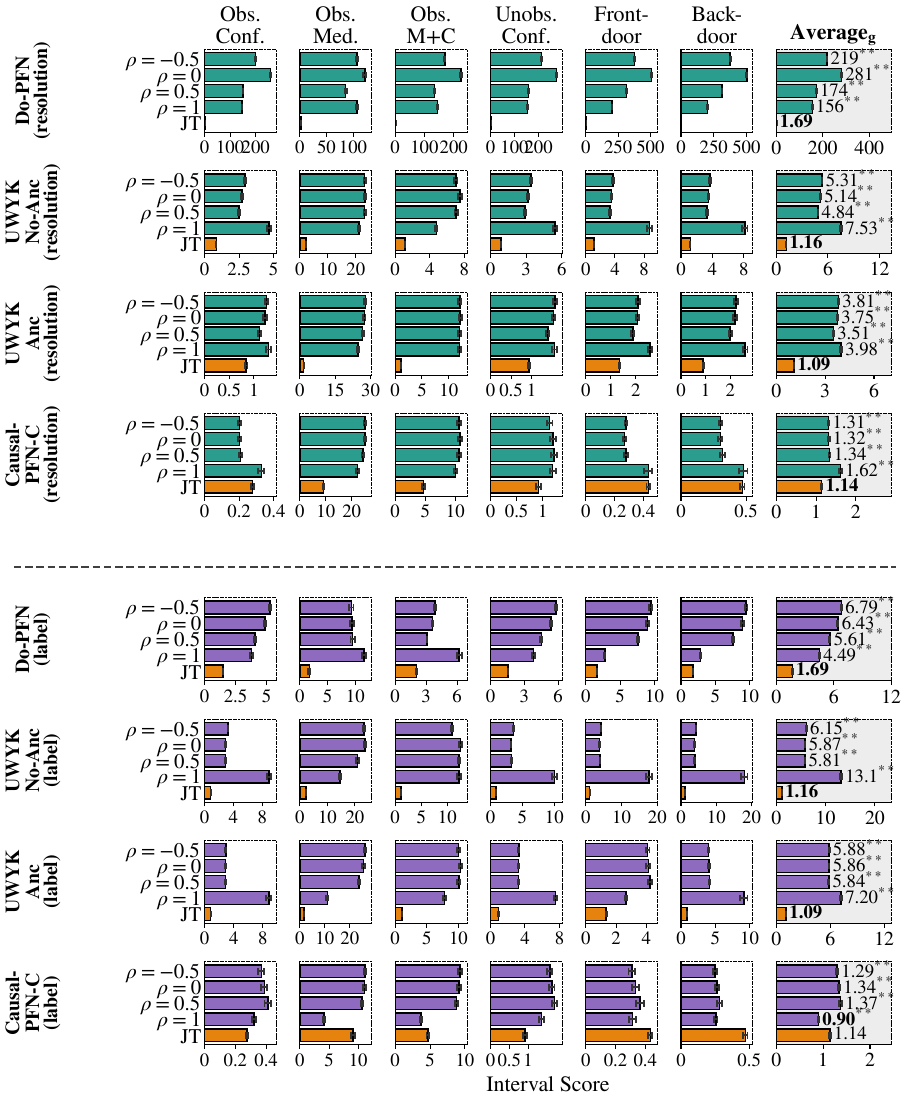}
\caption{\small Interval score $\mathrm{IS}_{0.05}/\sigma_Y$ (lower is better) of the 95\% interval of the \textbf{ATE} density on the six \textbf{case studies}, against the ablation baselines. The rows above the dashed line are the matched-resolution ablations (green) and those below are the matched-label ablations (purple), each under every $\rho$, with its family's JT model and the learned dependence (orange). Bars show the mean over realizations $\pm$ the standard error. The last column, Average$_g$, is the geometric mean over datasets $\pm$ its standard error, with the lowest value in bold. Significance against the JT model at the $0.1$ and $0.05$ level is indicated by $^{*}$ and $^{**}$, respectively. Each panel has its own scale.}
\label{fig:ate-abl-cs}
\end{figure}

\begin{figure}[tbp]
\centering
\includegraphics[width=\textwidth]{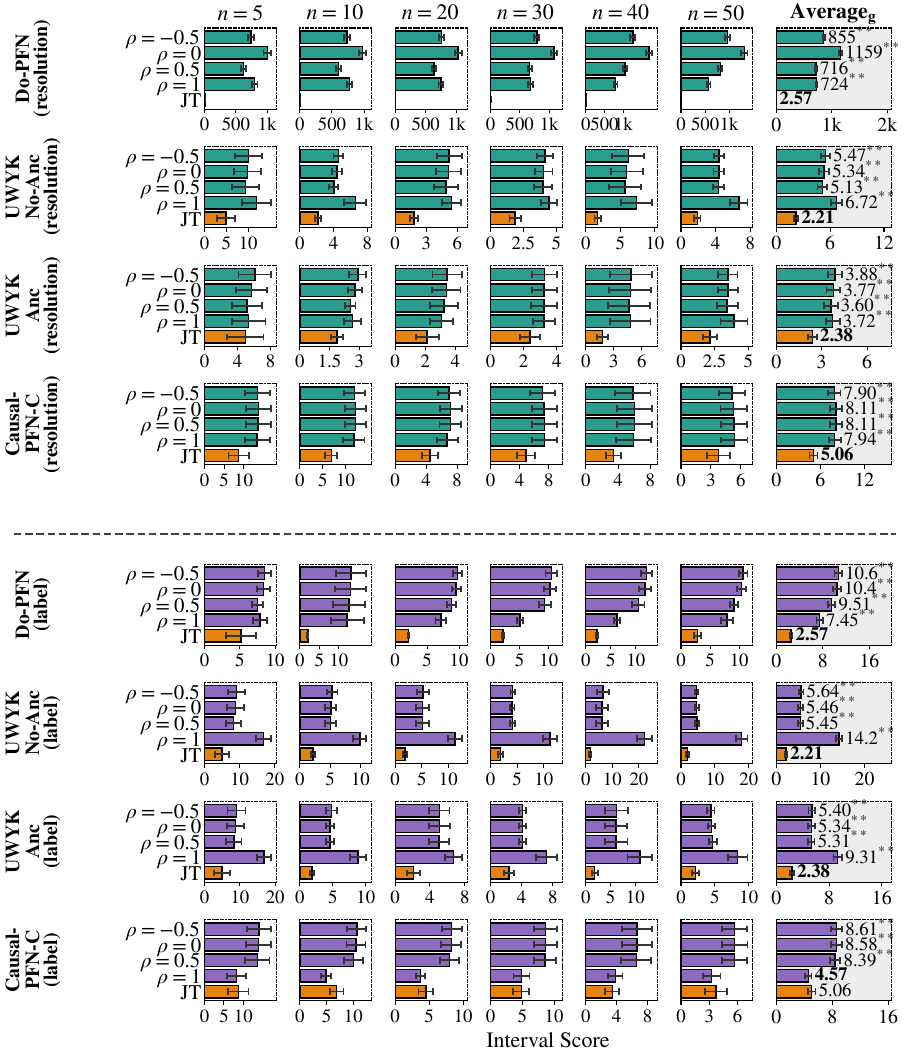}
\caption{\small Interval score $\mathrm{IS}_{0.05}/\sigma_Y$ (lower is better) of the 95\% interval of the \textbf{ATE} density on \textbf{ComplexMech} per graph size $n$, against the ablation baselines. The rows above the dashed line are the matched-resolution ablations (green) and those below are the matched-label ablations (purple), each under every $\rho$, with its family's JT model and the learned dependence (orange). Bars show the mean over realizations $\pm$ the standard error. The last column, Average$_g$, is the geometric mean over datasets $\pm$ its standard error, with the lowest value in bold. Significance against the JT model at the $0.1$ and $0.05$ level is indicated by $^{*}$ and $^{**}$, respectively. Each panel has its own scale.}
\label{fig:ate-abl-cm}
\end{figure}

Figures~\ref{fig:ablation-pehe}--\ref{fig:ablation-eps-cm} report the point estimation of the ablations. Against both ablations, the JT models of Do-PFN and UWYK keep lower unit- and population-level errors on the synthetic benchmarks. On the case studies, these errors are lower by $36$--$72\%$ and $15$--$72\%$, respectively. On ComplexMech, they are lower by $20$--$74\%$ and $17$--$84\%$, respectively, with one exception at the unit level. On RealCause, the population-level error is lower, by $16$--$48\%$. For CausalPFN-C the picture is mixed. On the case studies, its population-level error is $63\%$ and $26\%$ higher than that of the matched-label and matched-resolution baselines, respectively. Its unit-level error is $5\%$ higher than that of the matched-label baseline and $13\%$ lower than that of the matched-resolution baseline. On RealCause and ComplexMech, it does not differ significantly from either ablation.

\begin{figure}[tbp]
\centering
\includegraphics[width=\textwidth]{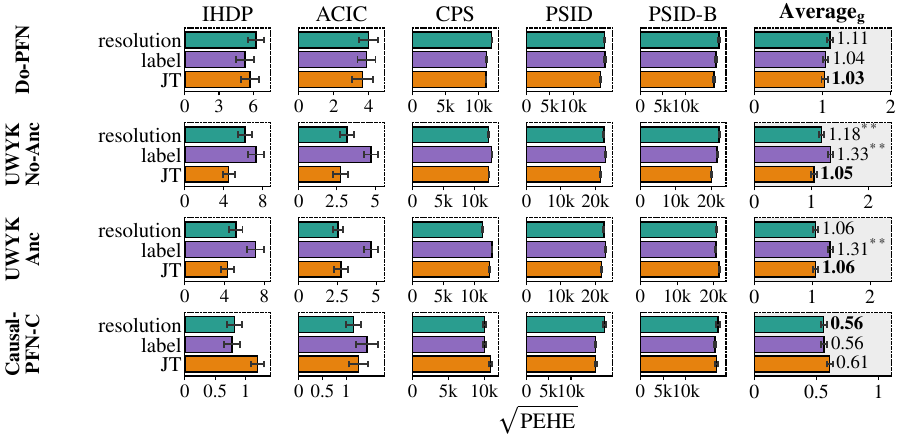}

    \vspace{6pt}
    \includegraphics[width=\textwidth]{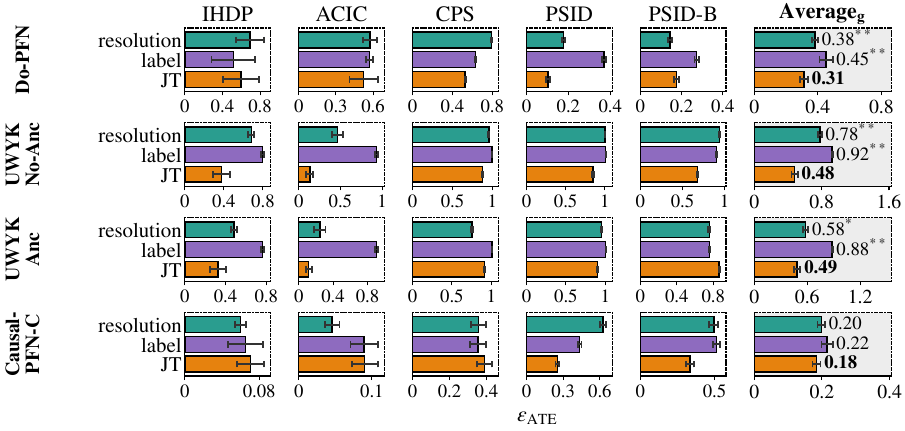}
\caption{\small $\sqrt{\mathrm{PEHE}}$ (top) and relative ATE error $\epsilon_{\mathrm{ATE}}$ (bottom) of the matched-resolution (green) and matched-label (purple) ablations and the JT models (orange) on \textbf{RealCause}; mean $\pm$ standard error, lower is better; each panel has its own scale. The last column, Average$_g$, is the geometric mean over datasets $\pm$ its standard error, with $\sqrt{\mathrm{PEHE}}$ standardised by each dataset's outcome scale $\sigma_Y$, and the lowest value in bold. Significance against the JT model at the $0.1$ and $0.05$ level is indicated by $^{*}$ and $^{**}$, respectively.}
\label{fig:ablation-pehe}\label{fig:ablation-eps}
\end{figure}
\begin{figure}[tbp]
\centering
\includegraphics[width=\textwidth]{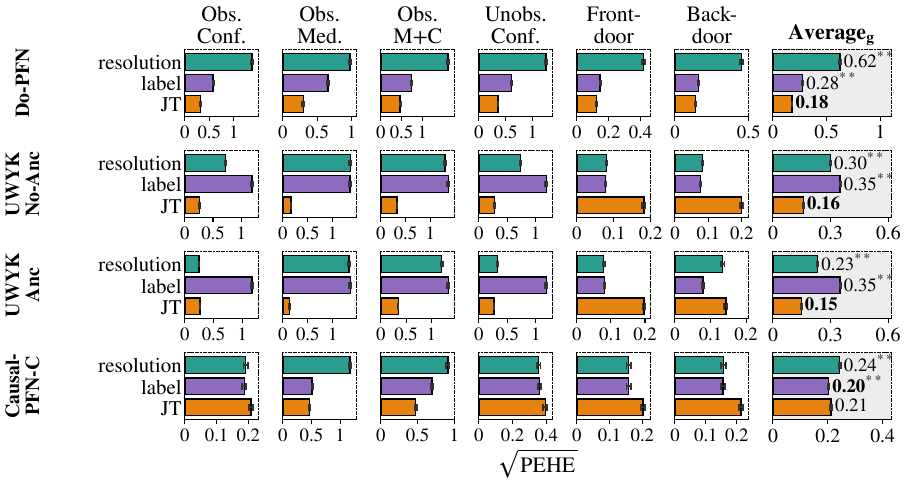}

    \vspace{6pt}
    \includegraphics[width=\textwidth]{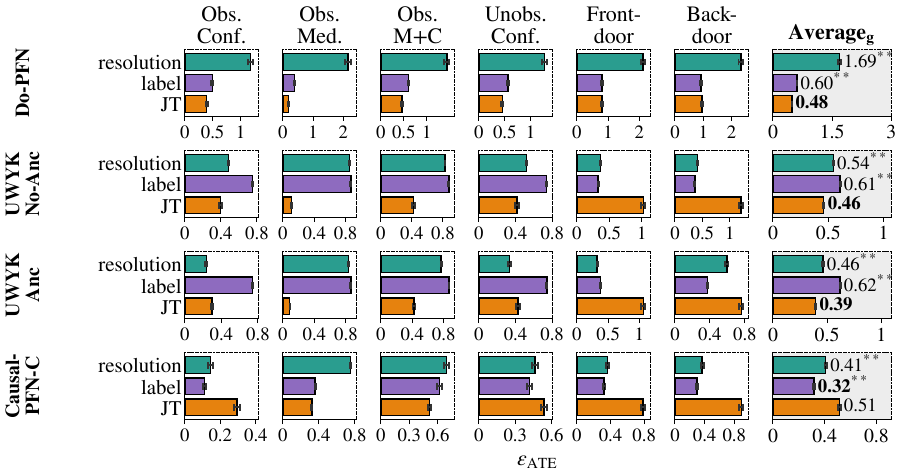}
\caption{\small $\sqrt{\mathrm{PEHE}}$ (top) and relative ATE error $\epsilon_{\mathrm{ATE}}$ (bottom) of the matched-resolution (green) and matched-label (purple) ablations and the JT models (orange) on the six \textbf{case studies}; mean $\pm$ standard error, lower is better; each panel has its own scale. The last column, Average$_g$, is the geometric mean over datasets $\pm$ its standard error, with $\sqrt{\mathrm{PEHE}}$ standardised by each dataset's outcome scale $\sigma_Y$, and the lowest value in bold. Significance against the JT model at the $0.1$ and $0.05$ level is indicated by $^{*}$ and $^{**}$, respectively.}
\label{fig:ablation-pehe-cs}\label{fig:ablation-eps-cs}
\end{figure}
\begin{figure}[tbp]
\centering
\includegraphics[width=\textwidth]{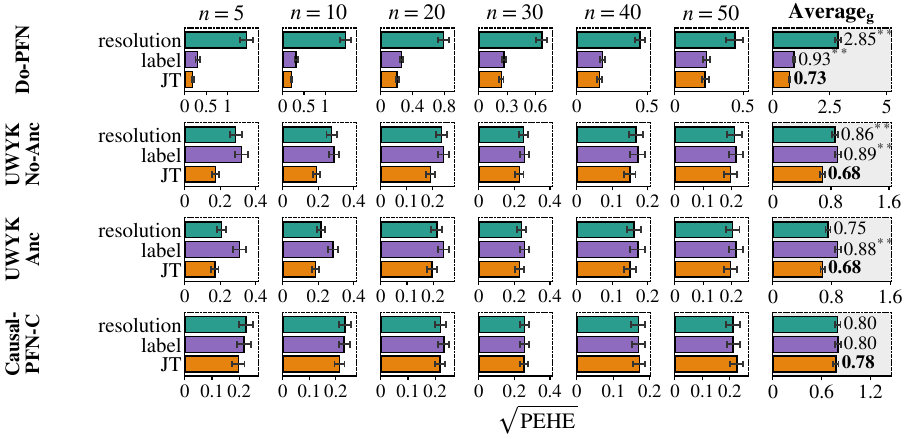}

    \vspace{6pt}
    \includegraphics[width=\textwidth]{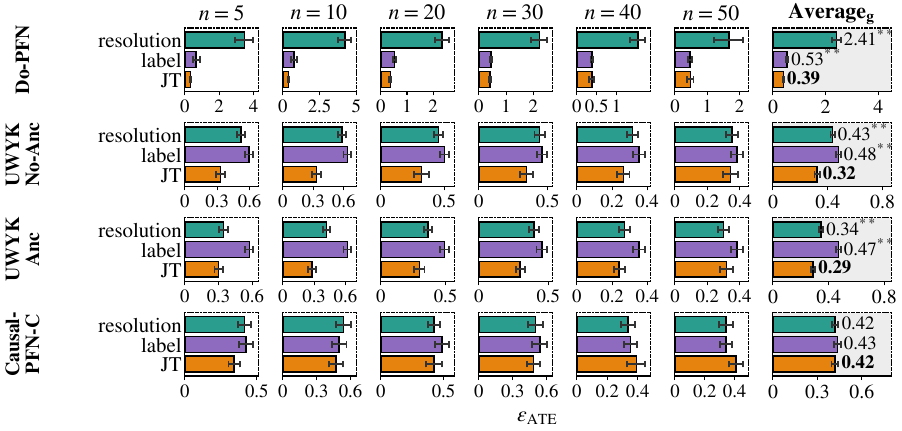}
\caption{\small $\sqrt{\mathrm{PEHE}}$ (top) and relative ATE error $\epsilon_{\mathrm{ATE}}$ (bottom) of the matched-resolution (green) and matched-label (purple) ablations and the JT models (orange) on \textbf{ComplexMech} per graph size $n$; mean $\pm$ standard error, lower is better; each panel has its own scale. The last column, Average$_g$, is the geometric mean over datasets $\pm$ its standard error, with $\sqrt{\mathrm{PEHE}}$ standardised by each dataset's outcome scale $\sigma_Y$, and the lowest value in bold. Significance against the JT model at the $0.1$ and $0.05$ level is indicated by $^{*}$ and $^{**}$, respectively.}
\label{fig:ablation-pehe-cm}\label{fig:ablation-eps-cm}
\end{figure}

\begin{figure}[tbp]
\centering
\includegraphics[width=\textwidth]{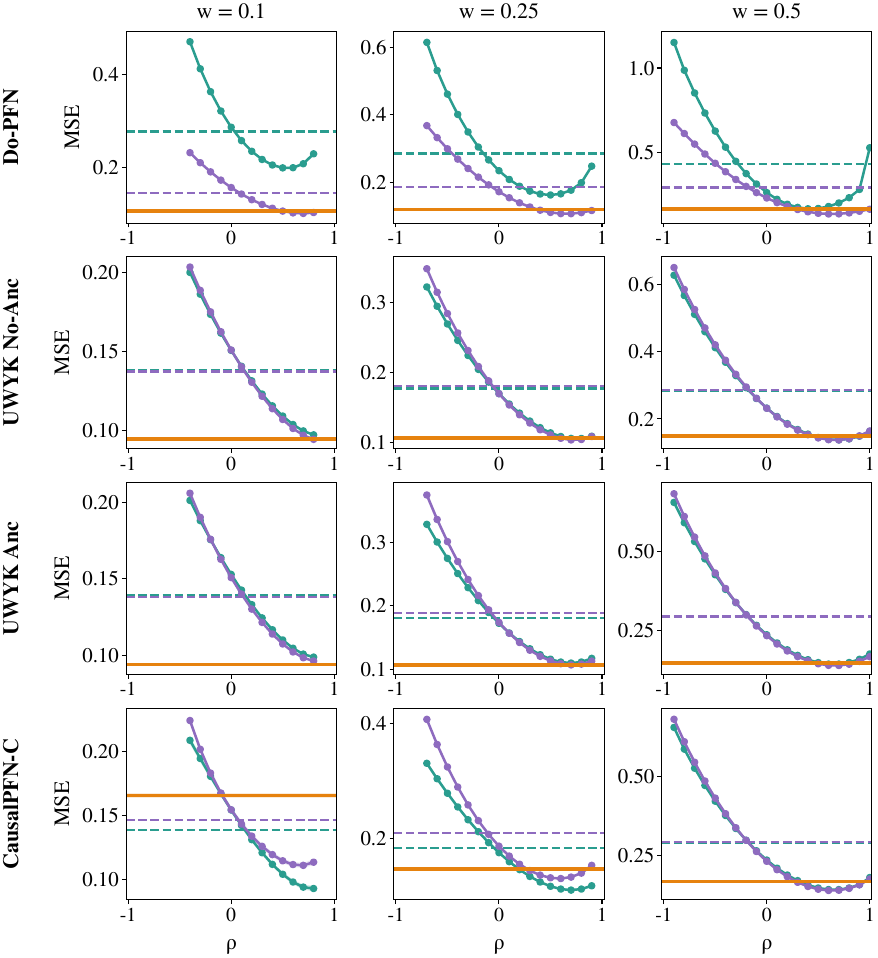}
\caption{\small Mean squared error on Twins at $d = 10$ against the ablation baselines, as in Figure~\ref{fig:twins-perw-d10}: the matched-resolution (green) and the matched-label (purple) baseline under each $\rho$ in the range compatible with the data, their averages over the range (dashed), and the JT model (orange).}
\label{fig:twins-abl-d10}
\end{figure}

\begin{figure}[tbp]
\centering
\includegraphics[width=\textwidth]{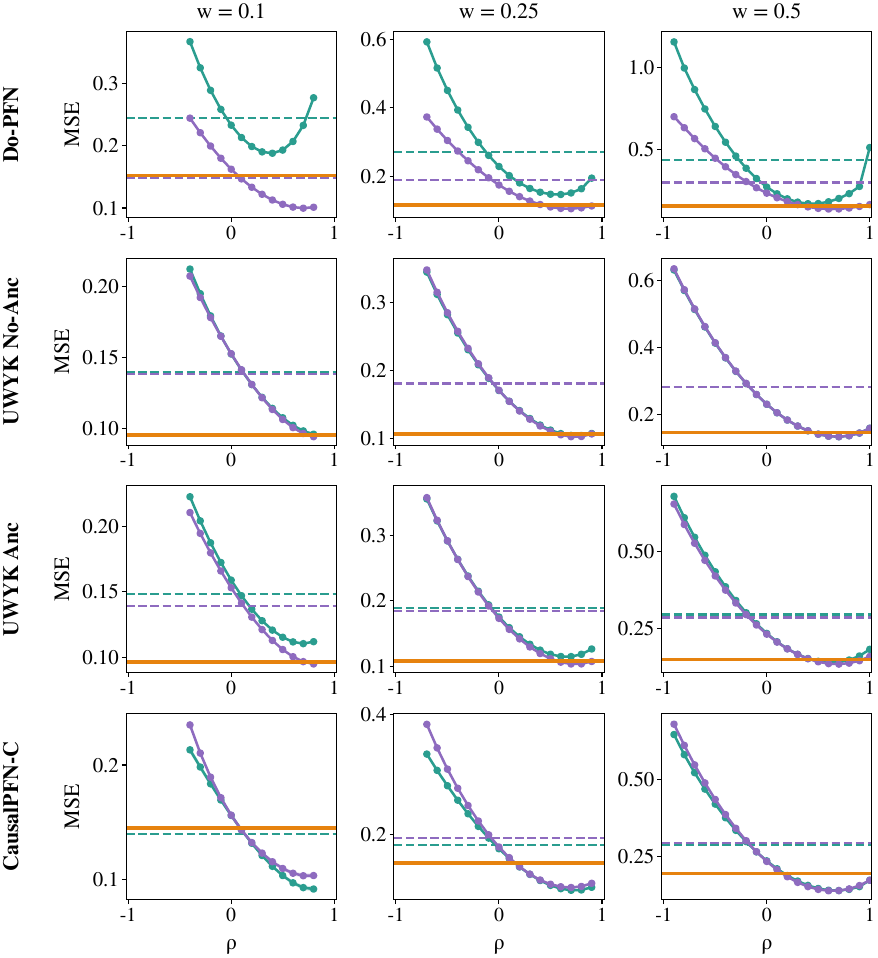}
\caption{\small Mean squared error on Twins at $d = 1$ against the ablation baselines, as in Figure~\ref{fig:twins-abl-d10}.}
\label{fig:twins-abl-d1}
\end{figure}

\subsection{Robustness to the cross-world correlation $\rho(x)$}
\label{app:varyrho}

In our results, we evaluate the models under various correlation structures: a strongly correlated one on the case studies, one that varies widely across datasets on ComplexMech, and a moderate one on Twins. In this section, we test them further in a controlled setting, where the correlation can be computed directly. The four case studies with pre-treatment covariates, IHDP, ACIC and ComplexMech with five nodes allow this, so we generate them with the outcome noise coupled across the two outcomes at a chosen level and group the datasets by their $\rho$ (Table~\ref{tab:rho-ctrl-wins}). The single-outcome baselines are reported under each assumed $\rho \in \{-0.5, 0, 0.5, 1\}$, so in every bin of the true $\rho$, at least one assumed value lies within the bin.

\begin{table}[tbp]\centering\small\setlength{\tabcolsep}{5pt}
\caption{\small Percentage of benchmarks on which the JT model attains a lower interval score $\mathrm{IS}_{0.05}$ of the 95\% posterior predictive interval for the \textbf{ITE} than its single-outcome baseline, on seven benchmarks with a controlled cross-world correlation: the four \textbf{case studies} with pre-treatment covariates, \textbf{IHDP}, \textbf{ACIC} and \textbf{ComplexMech} ($n=5$). Each column groups the datasets by their true $\rho$, and each row gives the $\rho$ used by the baseline at inference. The Average row and column average these percentages, and the last row gives the number of benchmarks. CausalPFN-C (resolution) compares CausalPFN-C JT with its matched-resolution baseline.}
\label{tab:rho-ctrl-wins}
\setlength{\aboverulesep}{0pt}\setlength{\belowrulesep}{0pt}\renewcommand{\arraystretch}{1.3}
\begin{tabular}{@{}cl|ccc|c@{}}
\toprule
& & \multicolumn{3}{c|}{True $\rho$} & \\
\cmidrule{3-5}
Model & \multicolumn{1}{c|}{At inference} & $[-0.5, 0)$ & $[0, 0.5)$ & $[0.5, 1]$ & Average \\
\midrule
\multirow{5}{*}{Do-PFN} & $\rho=-0.5$ & 71\% & 100\% & 100\% & 90\% \\
 & $\rho=0$ & 57\% & 71\% & 71\% & 67\% \\
 & $\rho=0.5$ & 71\% & 71\% & 71\% & 71\% \\
 & $\rho=1$ & 100\% & 71\% & 71\% & 81\% \\
\cmidrule{2-6}
& Average & 75\% & 79\% & 79\% & 77\% \\
\midrule
\multirow{5}{*}{UWYK No-Anc} & $\rho=-0.5$ & 86\% & 100\% & 100\% & 95\% \\
 & $\rho=0$ & 71\% & 86\% & 86\% & 81\% \\
 & $\rho=0.5$ & 86\% & 71\% & 71\% & 76\% \\
 & $\rho=1$ & 100\% & 86\% & 71\% & 86\% \\
\cmidrule{2-6}
& Average & 86\% & 86\% & 82\% & 85\% \\
\midrule
\multirow{5}{*}{UWYK Anc} & $\rho=-0.5$ & 86\% & 86\% & 100\% & 90\% \\
 & $\rho=0$ & 71\% & 71\% & 86\% & 76\% \\
 & $\rho=0.5$ & 43\% & 43\% & 29\% & 38\% \\
 & $\rho=1$ & 100\% & 86\% & 71\% & 86\% \\
\cmidrule{2-6}
& Average & 75\% & 71\% & 71\% & 73\% \\
\midrule
\multirow{5}{*}{CausalPFN-C} & $\rho=-0.5$ & 14\% & 14\% & 14\% & 14\% \\
 & $\rho=0$ & 14\% & 14\% & 14\% & 14\% \\
 & $\rho=0.5$ & 14\% & 14\% & 14\% & 14\% \\
 & $\rho=1$ & 86\% & 57\% & 43\% & 62\% \\
\cmidrule{2-6}
& Average & 32\% & 25\% & 21\% & 26\% \\
\midrule
\multirow{5}{*}{\begin{tabular}{@{}c@{}}CausalPFN-C\\(resolution)\end{tabular}} & $\rho=-0.5$ & 86\% & 100\% & 100\% & 95\% \\
 & $\rho=0$ & 71\% & 86\% & 100\% & 86\% \\
 & $\rho=0.5$ & 29\% & 0\% & 43\% & 24\% \\
 & $\rho=1$ & 14\% & 14\% & 14\% & 14\% \\
\cmidrule{2-6}
& Average & 50\% & 50\% & 64\% & 55\% \\
\midrule
Benchmarks & & 7 & 7 & 7 & 7 \\
\bottomrule
\end{tabular}
\end{table}

\textit{Finding 7: The advantage of joint learning on Do-PFN and UWYK holds across the tested range of dependence, while CausalPFN-C is about even with its baseline at matched resolution.} Averaged over the true $\rho$ and the $\rho$ used at inference, the JT models attain a lower interval score on $77\%$ of the benchmarks for Do-PFN, and on $85\%$ and $73\%$ for UWYK No-Anc and UWYK Anc (Table~\ref{tab:rho-ctrl-wins}). CausalPFN-C JT does so on $26\%$, and on $55\%$ against its matched-resolution baseline.

\subsection{CausalPFN two-stage comparison}
\label{app:twostagecomparison}

The published CausalPFN is trained in two stages: the first predicts observational outcomes on real datasets, and the second carries those weights over and predicts interventional outcomes on the synthetic prior. The first stage does not transfer to the joint setting: its weights do not match the bin structure of the joint output, and its real datasets contain no counterfactual outcomes to pair. We therefore apply joint learning to the second, causal stage and re-train the baseline from the same stage under the same protocol; the main results compare against this re-trained baseline, CausalPFN-C. Figure~\ref{fig:cmp-cpfn-summary} compares CausalPFN-C JT against the published two-stage checkpoint instead, in point estimation and in calibration, with both evaluated against the ITE (Appendix~\ref{app:metrics}).

\begin{figure}[tbp]
\centering
\includegraphics[width=\textwidth]{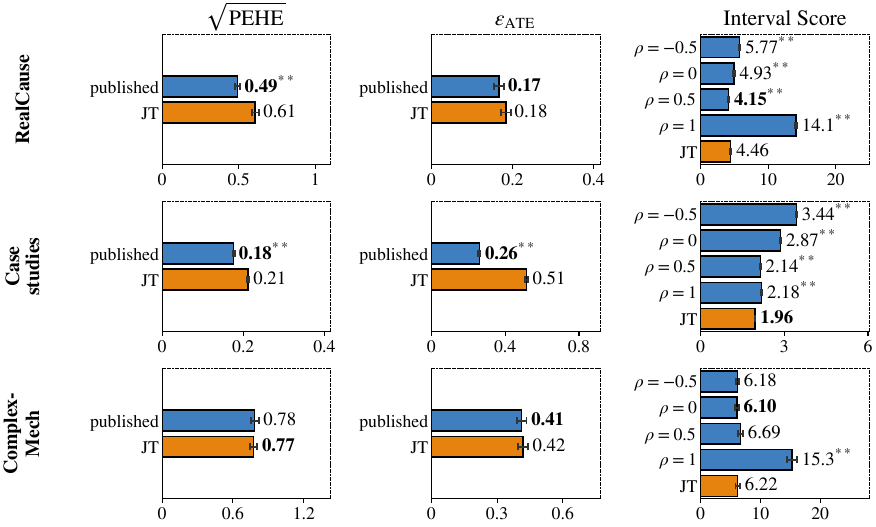}
\caption{\small Average$_g$ of $\sqrt{\mathrm{PEHE}}$, $\epsilon_{\mathrm{ATE}}$ and the interval score $\mathrm{IS}_{0.05}/\sigma_Y$ (lower is better) of CausalPFN-C JT (orange) against the published two-stage CausalPFN checkpoint (blue), instead of the re-trained CausalPFN-C of the main results, on \textbf{RealCause}, the \textbf{case studies} and \textbf{ComplexMech} (rows). The interval score shows the baseline under each assumed $\rho$ and the JT model with its learned dependence. Geometric mean over datasets $\pm$ its standard error, with $\sqrt{\mathrm{PEHE}}$ standardised by each dataset's outcome scale $\sigma_Y$. The lowest value of each panel is in bold. Significance against the JT model at the $0.1$ and $0.05$ level is indicated by $^{*}$ and $^{**}$, respectively.}
\label{fig:cmp-cpfn-summary}
\end{figure}

\textit{Finding 8: Against the published CausalPFN, joint learning trades point estimation for better intervals.} The published two-stage checkpoint has a clear advantage in point estimation. It is better in both metrics on the case studies and in the error of the estimated treatment effects on RealCause, while the two are comparable on ComplexMech. The posterior predictive intervals, however, favour joint learning. On the case studies, CausalPFN-C JT attains a lower interval score than the published checkpoint under every assumed $\rho$ ($1.96$ against $2.14$--$3.44$). On RealCause it attains a lower interval score under three of the four assumed values of $\rho$, and on ComplexMech the two are comparable ($6.22$ against $6.10$--$15.26$).

\subsection{UWYK binary re-training}
\label{app:binarytraining}
UWYK provides two training options, binary and continuous treatment prediction. The published checkpoint covers only the continuous version, so for fairness we also re-train the binary version. Figure~\ref{fig:cmp-uwykb-summary} compares the JT model against the re-trained binary version, UWYK-B, for both the ancestral and no-ancestral variants.

\begin{figure}[tbp]
\centering
\includegraphics[width=\textwidth]{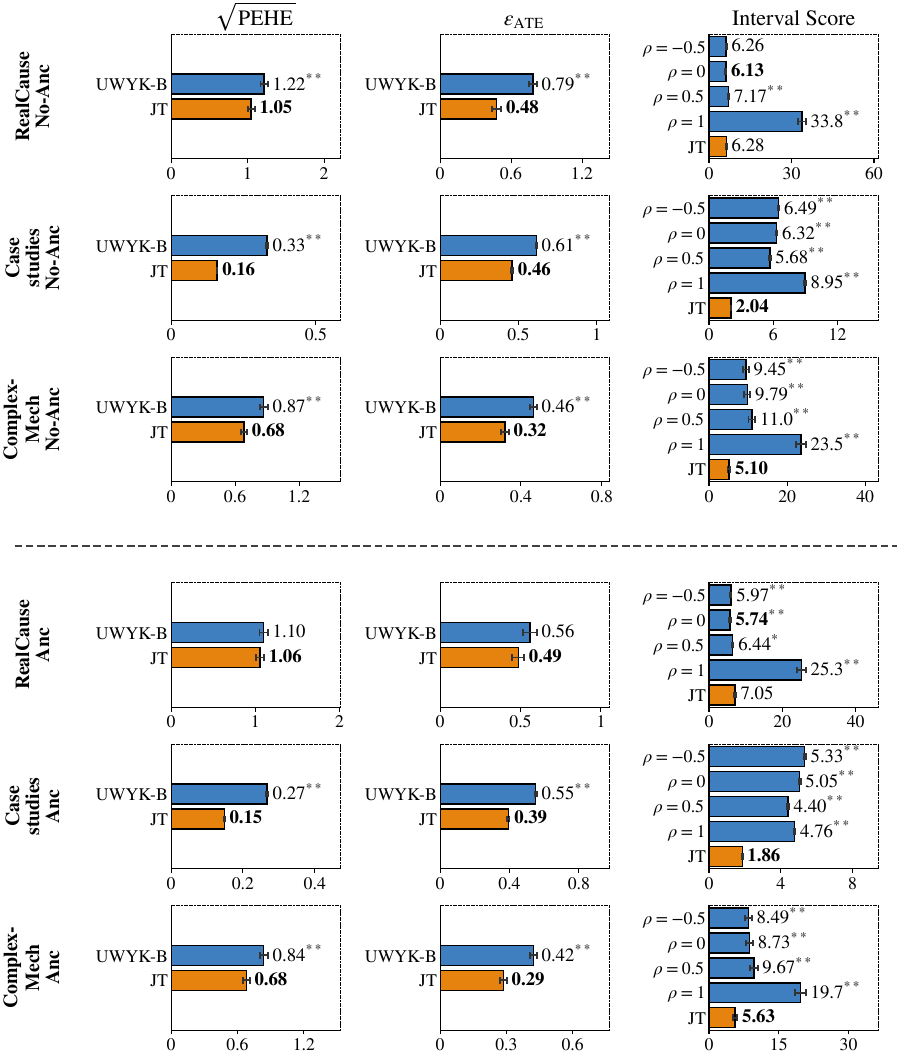}
\caption{\small Average$_g$ of $\sqrt{\mathrm{PEHE}}$, $\epsilon_{\mathrm{ATE}}$ and the interval score $\mathrm{IS}_{0.05}/\sigma_Y$ (lower is better) of UWYK JT (orange) against the re-trained binary-treatment baseline UWYK-B (blue), on \textbf{RealCause}, the \textbf{case studies} and \textbf{ComplexMech}, for the no-ancestral variant (No-Anc, above the dashed line) and the ancestral variant (Anc, below it). The interval score shows the baseline under each assumed $\rho$ and the JT model with its learned dependence. Geometric mean over datasets $\pm$ its standard error, with $\sqrt{\mathrm{PEHE}}$ standardised by each dataset's outcome scale $\sigma_Y$. The lowest value of each panel is in bold. Significance against the JT model at the $0.1$ and $0.05$ level is indicated by $^{*}$ and $^{**}$, respectively.}
\label{fig:cmp-uwykb-summary}
\end{figure}

\textit{Finding 9: Re-training UWYK on binary treatments does not change the main results.} In point estimation, UWYK JT improves upon UWYK-B on the synthetic benchmarks, for both variants: by $25$--$52\%$ and $29$--$45\%$ on the case studies, and $21$--$30\%$ and $19$--$32\%$ on ComplexMech, for the no-ancestral and ancestral variants respectively. On RealCause, only the no-ancestral variant improves, by $14$--$39\%$. The posterior predictive intervals follow the same pattern as against the published checkpoint. On the synthetic benchmarks, where the outcomes are dependent, UWYK JT attains a lower interval score than UWYK-B under every assumed $\rho$, for both variants: $2.8$--$4.4$ and $2.4$--$2.9$ times lower on the case studies, and $1.9$--$4.6$ and $1.5$--$3.5$ times lower on ComplexMech. On RealCause, where the outcomes are independent, UWYK-B attains a lower interval score under most assumed values of $\rho$ for the ancestral variant (three of four), consistent with Finding~4.

\section{Computational Details}
\label{app:compute}

Table~\ref{tab:train_cost} reports the inference time per dataset. Each JT model needs roughly half the forward time of its single-outcome baseline, as one pass yields the joint while the baselines require one pass per treatment arm. The ITE and ATE densities are post-processing on the CPU; the ATE barycenter adds about $0.01$\,s.

Each model was trained on a single GPU of one of three types: NVIDIA L40S (48\,GB) for the Do-PFN models, NVIDIA H100 (80\,GB, SXM5) for the CausalPFN models and the UWYK JT and binary models, and NVIDIA RTX PRO 6000 Blackwell Server Edition (96\,GB) for the UWYK matched-resolution and matched-label models. Table~\ref{tab:train_cost} reports the hardware, throughput, and cost of each run.

For evaluation, inference was done using a mix of single NVIDIA L40S GPUs (48 GB) per task, or Intel Granite Rapids Xeon CPUs (8 cores, 64GB at a time). 

The implementation uses Python 3.11, PyTorch 2.11.0, NumPy
1.26.4, SciPy 1.15.2,
scikit-learn 1.2.2, and XGBoost 3.2.0, the last of which
enters results through the synthetic-prior mechanism sampler.

\begin{sidewaystable}
\centering
\caption{\small Training cost per baseline, ablation, and jointly trained versions. Forward and ATE density: inference time per dataset on \textbf{RealCause} \textbf{IHDP} (realization 0; $672$ context units, $75$ queries), median over $20$ timed forward runs after $3$ warm-up runs on a single NVIDIA L40S, and over $5$ runs of the ATE-density step. Output dimension counts the outputs per query: the grid logits plus the models' auxiliary outputs, such as tail and region parameters ($+13$ for the JT models, Appendix~\ref{app:head-spec}; $+0$, $+4$, and $+10$ for the Do-PFN, UWYK, and CausalPFN single-outcome models). Parameter counts are deduplicated tensor counts of the checkpoints. Published models are not trained by us, so their training columns are not applicable.}
\label{tab:train_cost}
\small\setlength{\tabcolsep}{4pt}
\begin{tabular}{llcccccccc}
\toprule
Model & Variant & Grid & \makecell{Output\\dim} & \makecell{Total\\params} & \makecell{Throughput\\(s/step)} & Steps & \makecell{GPU-\\hours} & \makecell{Forward\\(s)} & \makecell{ATE\\density (s)} \\
\midrule

Do-PFN & Published & $100$ & $100$ & $7{,}336{,}721$ & --- & $262{,}144$ & --- & $0.275$ & $0.010$ \\
 & JT & $10^2$ & $10^2{+}13=113$ & $7{,}346{,}718$ & $0.223$ (L40S) & $262{,}144$ & $16.2$ & $0.136$ & $0.010$ \\
 & Matched resolution & $10$ & $10$ & $7{,}267{,}241$ & $0.216$ (L40S) & $262{,}144$ & $15.7$ & $0.274$ & $0.010$ \\
 & Matched labels & $100$ & $100$ & $7{,}336{,}721$ & $0.284$ (L40S) & $262{,}144$ & $20.6$ & $0.273$ & $0.009$ \\
\midrule
UWYK & Continuous (published) & $1000$ & $1000{+}4=1004$ & $11{,}186{,}463$ & --- & $50{,}000$ & --- & $0.126$ & $0.010$ \\
 & Binary (retrained) & $1000$ & $1000{+}4=1004$ & $11{,}186{,}463$ & $1.63$ (H100) & $50{,}000$ & $22.7$ & $0.127$ & $0.010$ \\
 & JT & $32^2$ & $32^2{+}13=1037$ & $14{,}010{,}945$ & $1.90$ (H100) & $50{,}000$ & $26.4$ & $0.073$ & $0.010$ \\
 & Matched resolution & $32$ & $32{+}4=36$ & $10{,}937{,}687$ & $1.71$ (Blackwell) & $50{,}000$ & $23.8$ & $0.126$ & $0.010$ \\
 & Matched labels & $1000$ & $1000{+}4=1004$ & $11{,}186{,}463$ & $1.69$ (Blackwell) & $50{,}000$ & $23.5$ & $0.127$ & $0.010$ \\
\midrule
CausalPFN & Two-stage (published) & $1024$ & $1024{+}10=1034$ & $18{,}844{,}034$ & --- & --- & --- & $0.032$ & $0.009$ \\
 & Stage-2 (reproduction) & $1024$ & $1024{+}10=1034$ & $18{,}844{,}034$ & $2.57$ (H100) & $50{,}000$ & $35.7$ & $0.032$ & $0.009$ \\
 & JT & $32^2$ & $32^2{+}13{+}10=1047$ & $18{,}852{,}002$ & $1.96$ (H100) & $50{,}000$ & $27.2$ & $0.016$ & $0.009$ \\
 & Matched resolution & $32$ & $32{+}10=42$ & $18{,}080{,}194$ & $1.90$ (H100) & $50{,}000$ & $26.4$ & $0.031$ & $0.009$ \\
 & Matched labels & $1024$ & $1024{+}10=1034$ & $18{,}844{,}034$ & $2.60$ (H100) & $50{,}000$ & $36.1$ & $0.032$ & $0.009$ \\
\bottomrule
\end{tabular}%

\end{sidewaystable}

\end{document}

%% file: math_commands.tex
\usepackage{amsmath,amsfonts,bm}
\usepackage{amsmath, amssymb}   %
\usepackage{mathtools}
\usepackage{hyperref}
\usepackage{xcolor}
\usepackage{enumitem}
\usepackage{microtype}
\usepackage{graphicx}
\graphicspath{{Figures/}{../paper/theory/figures/}{../benchmarks/plots/TE_demonstration/}{../benchmarks/plots/TE_demonstration/panels/}}
\usepackage{booktabs}
\usepackage{subcaption}
\usepackage{natbib}

\def\eqref#1{equation~\ref{#1}}

\def\1{\bm{1}}

\DeclareMathAlphabet{\mathsfit}{\encodingdefault}{\sfdefault}{m}{sl}
\SetMathAlphabet{\mathsfit}{bold}{\encodingdefault}{\sfdefault}{bx}{n}